\documentclass{article}

    \usepackage[final,nonatbib]{neurips_2020}

\usepackage{amsmath, amssymb, amsthm}
\usepackage{mathtools}
\usepackage{bbm}
\usepackage{dsfont}

\usepackage[usenames,dvipsnames]{xcolor}
\definecolor{shadecolor}{gray}{0.9}

\usepackage{graphicx}
\usepackage[labelfont=bf]{caption}
\usepackage[format=hang]{subcaption}
\usepackage{stfloats}

\usepackage{booktabs, array}
\usepackage{multirow}
\usepackage{rotating}

\usepackage[algoruled]{algorithm2e}
\usepackage{listings}
\usepackage{fancyvrb}
\fvset{fontsize=\small}

\usepackage[square,numbers]{natbib}
\usepackage[colorlinks,linktoc=all]{hyperref}
\usepackage[all]{hypcap}
\hypersetup{citecolor=MidnightBlue}
\hypersetup{linkcolor=MidnightBlue}
\hypersetup{urlcolor=MidnightBlue}
\usepackage[nameinlink,capitalise]{cleveref}

\usepackage[acronym,smallcaps,nowarn,section,nogroupskip,nonumberlist]{glossaries}
\glsdisablehyper{}

\lstdefinestyle{mystyle}{
    commentstyle=\color{OliveGreen},
    numberstyle=\tiny\color{black!60},
    stringstyle=\color{BrickRed},
    basicstyle=\ttfamily\scriptsize,
    breakatwhitespace=false,
    breaklines=true,
    captionpos=b,
    keepspaces=true,
    numbers=none,
    numbersep=5pt,
    showspaces=false,
    showstringspaces=false,
    showtabs=false,
    tabsize=2
}
\newacronym{mcmc}{mcmc}{Markov Chain Monte Carlo}
\newacronym{elbo}{elbo}{Evidence Lower BOund}
\newacronym{gp}{gp}{Gaussian Process}
\newacronym{np}{np}{Neural Process}
\newacronym{cnp}{cnp}{Conditional Neural Process}
\newacronym{anp}{anp}{Attentive Neural Process}
\newacronym{bnp}{bnp}{Bootstrapping Neural Process}
\newacronym{canp}{canp}{Conditional Attentive Neural Process}
\newacronym{banp}{banp}{Bootstrapping Attentive Neural Process}
\newacronym{de}{de}{Deep Ensemble}

\newcommand{\calE}{{\mathcal{E}}}

\newcommand{\calN}{{\mathcal{N}}}

\newcommand{\calT}{{\mathcal{T}}}

\newcommand{\bbE}{\mathbb{E}}

\theoremstyle{plain}%

\theoremstyle{definition}

\theoremstyle{remark}

\DeclareMathOperator{\unifdist}{Unif}

\newcommand{\msd}[2]{{#1}{\tiny $\pm${#2}}}

\newcommand{\tr}{^\top}

\newcommand{\iidsim}{\overset{\mathrm{i.i.d.}}{\sim}}
\newcommand{\1}[1]{\mathds{1}_{\{#1\}}}

\def\[#1\]{\begin{align}#1\end{align}}

\newcommand{\ssc}[1]{^{\scriptscriptstyle{(#1)}}}
\newcommand{\swrsim}{\overset{\mathrm{s.w.r.}}{\sim}}

\newcommand{\tx}{\tilde x}
\newcommand{\tX}{\tilde X}
\newcommand{\ty}{\tilde y}
\newcommand{\tY}{\tilde Y}
\usepackage{xr}
\makeatletter
\newcommand*{\addFileDependency}[1]{%
  \typeout{(#1)}
  \@addtofilelist{#1}
  \IfFileExists{#1}{}{\typeout{No file #1.}}
}
\makeatother

\usepackage[utf8]{inputenc} %
\usepackage[T1]{fontenc}    %
\usepackage{url}            %
\usepackage{nicefrac}       %
\usepackage{microtype}      %
\usepackage{pdfpages}

\usepackage[toc,page]{appendix}

\title{Bootstrapping Neural Processes}

\author{%
Juho Lee$^{1, 2 *}$,
Yoonho Lee$^{2 *}$,
Jungtaek Kim$^{3}$, \\
\textbf{
  Eunho Yang$^{1, 2}$,
  Sung Ju Hwang$^{1, 2}$,
  Yee Whye Teh$^{4}$ } \\
KAIST$^1$, Daejeon, South Korea,
AITRICS$^2$, Seoul, South Korea, \\
POSTECH$^3$, Pohang, South Korea, 
University of Oxford$^4$, Oxford, England \\
\texttt{juholee@kaist.ac.kr}
}

\newcommand\blfootnote[1]{%
  \begingroup
  \renewcommand\thefootnote{}\footnote{#1}%
  \addtocounter{footnote}{-1}%
  \endgroup
}
\begin{document}
\blfootnote{$^*$ Equal contribution}

\maketitle

\begin{abstract}
  Unlike in the traditional statistical modeling for which a user typically hand-specify a prior, \glspl{np} implicitly define a broad class of stochastic processes with neural networks. Given a data stream, \gls{np} learns a stochastic process that best describes the data. While this ``data-driven'' way of learning stochastic processes has proven to handle various types of data, 
  \glspl{np} still rely on an assumption that uncertainty in stochastic processes is modeled by a single latent variable, which potentially limits the flexibility. To this end, we propose the \gls{bnp}, a novel extension of the \gls{np} family using the bootstrap. The bootstrap is a classical data-driven technique for estimating uncertainty, which allows \gls{bnp} to learn the stochasticity in \glspl{np} without assuming a particular form. We demonstrate the efficacy of \gls{bnp} on various types of data and its robustness in the presence of model-data mismatch.
\end{abstract}

\section{Introduction}
\glsreset{np}
\label{sec:introduction}
\gls{np}~\citep{garnelo2018neural} is a class of stochastic processes defined by parametric neural networks. 
Traditional stochastic processes such as \gls{gp}~\citep{RasmussenCE2006book} are usually derived from mathematical objects based on certain prior beliefs on data (e.g., smoothness of functions quantified by Gaussian distributions).
On the other hand, given a stream of data, \gls{np} \emph{learns} to construct a stochastic process that might describe the data well. In that sense, \gls{np} may be considered as a data-driven way of defining stochastic processes. When appropriately trained, \gls{np} can define a flexible class of stochastic processes well suited for highly non-trivial functions that are not easily represented by existing stochastic processes.

Like other stochastic processes, \gls{np} induces stochasticity in function realizations. More specifically, \gls{np} defines a function value $y$ for a point $x$ as a conditional distribution $p(y|x, \dots)$ to model \emph{point-wise uncertainty}. Additionally, \gls{np} further introduces a \emph{global latent variable} capturing \emph{functional uncertainty} - a global uncertainty in the overal structure of the function. The global latent variable modeling functional uncertainty is empirically demonstrated to improve the predictive performance and diversity in function realizations~\citep{le2018empirical}.

Although it is clear both intuitively and empirically that adding functional uncertainty helps, it remains unclear whether modeling it with a single Gaussian latent variable is optimal. 
For instance, \cite{louizos2019functional} pointed out that the global latent variable acts as a bottleneck. One could introduce more complex architectures to better capture the functional uncertainty, but that would typically come with an architectural overhead. Moreover, it contradicts the philosophy behind \gls{np} to use minimal modeling assumptions and let the model learn from data.

This paper introduces a novel way of introducing functional uncertainty to the family of \gls{np} models.  We revisit the bootstrap~\citep{efron1979bootstrap}, a classic frequentist technique to model uncertainty in parameter estimation by simulating population distribution via resampling. The bootstrap is a simple yet effective way of modelling uncertainty in a data-driven way, making it well-suited for our purpose of giving uncertainty to \gls{np} with minimal modeling assumptions. To this end, we propose \gls{bnp}, an extension of \gls{np} using bootstrap to induce functional uncertainty. \gls{bnp} utilizes bootstrap to construct multiple resampled datasets and combines the predictions computed from them. The functional uncertainty is then naturally induced by the uncertainty in the bootstrap procedure.

 \gls{bnp} can be defined for any existing \gls{np} variants with minimal additional parameters and provides several benefits over existing models. One important aspect is its robustness under the presence of \emph{model-data mismatch}, where test data come from distributions different from the one used to train the model. An ensemble of bootstrap is well known to enhance the stability and accuracy~\citep{breiman1996bagging}. Recently, \cite{huggins2019using} showed that ensembling Bayesian posteriors from multiple bootstrap samples dramatically improves the robustness under model-data mismatch. We show that our extension of \gls{np} with bootstrap also enjoys this property. Using various data ranging from simple synthetic data to challenging real-world data, we demonstrate that \gls{bnp} is much more robust than the existing \gls{np} with global latent variables. This tendency was particularly strong under model-data mismatch, where the test data is significantly different from the datasets used to train the model.
\section{Background}
\label{sec:backgrounds}
\subsection{(Attentive) Neural Processes}

\label{subsec:np_review}
Consider a regression task $\calT = (X, Y, c)$ defined by an observation set $X = \{x_i\}_{i=1}^n$, a label set $Y = \{y_i\}_{i=1}^n$, and an index set $c \subsetneq \{1, \dots, n\}$ defining \emph{context} $(X_c, Y_c) := \{(x_i, y_i)\}_{i\in c}$. The goal is to learn a stochastic process (random function) mapping $x$ to $y$ given the context $(X_c, Y_c)$ as training data (a realization from the stochastic process),
i.e., learning
\[
\log p(Y|X, Y_c) = \sum_{i=1}^n \log p(y_i|x_i, X_c, Y_c).
\]
\gls{cnp}~\cite{garnelo2018conditional} models $p(y_i|x_i, X_c, Y_c)$ with a deterministic neural network taking $(X_c, Y_c)$ and $x_i$ to output the parameters of $p(y_i|x_i, X_c, Y_c)$. 
\gls{cnp} consists of an encoder and a decoder; the encoder summarizes $(X_c, Y_c)$ into a representation $\phi$ via permutation-invariant neural network~\cite{edwards2016towards,zaheer2017deep}, and the decoder transforms $\phi$ and $x_i$ into the target distribution (e.g., Gaussian),
\[
\label{eq:cnp_architecture}
&\phi = f_\text{enc}(X_c, Y_c) = f_{\text{enc}}\ssc{2}\bigg(\frac{1}{|c|}\sum_{i\in c} f_{\text{enc}}\ssc{1}(x_i, y_i)\bigg), \\
&(\mu_i, \sigma_i) = f_\text{dec}(\phi, x_i), \quad p(y_i|x_i, X_c, Y_c) = \calN(y_i | \mu_i, \sigma_i^2),
\]
where $f_{\text{enc}}\ssc{1}, f_{\text{enc}}\ssc{2}$ and $f_\text{dec}$ are feed-forward neural networks. \gls{cnp} is then trained to maximize the expected likelihood $\bbE_{p(\calT)}[\log p(Y|X, Y_c)]$. The variance $\sigma_i^2$ models the \emph{point-wise} uncertainty for $y_i$ given the context. \gls{np}~\cite{garnelo2018neural} further models \emph{functional uncertainty} using a \emph{global latent variable}. Unlike \gls{cnp}, which maps a context into a deterministic representation $\phi$, \gls{np} encodes the context into a Gaussian latent variable $z$, giving additional stochasticity in function construction.
Following \cite{kim2018attentive}, we consider a \gls{np} with both deterministic path and latent path, where the deterministic path models the overall skeleton of the function $\phi$, and the latent path
models the functional uncertainty:
\[\label{eq:np_architecture}
&\phi = f_\text{denc}(X_c, Y_c), \quad (\eta, \rho) = f_\text{lenc}(X_c, Y_c), \quad q(z|X_c, Y_c) = \calN(z;\eta,\rho^2)\\
&(\mu_i, \sigma_i) = f_\text{dec}(\phi, z, x_i), \quad p(y_i|x_i, z, \phi) = \calN(y_i|\mu_i, \sigma_i^2),
\]
with $f_\text{denc}$ and $f_\text{lenc}$ having the same structure as $f_\text{enc}$ in \eqref{eq:cnp_architecture}. 
The conditional probability is lower-bounded as
\[
\log p(Y|X, Y_c) \geq \sum_{i=1}^n \bbE_{q(z|X, Y)}\bigg[ \log \frac{p(y_i|x_i, z, \phi) p(z|X_c, Y_c)}{q(z|X, Y)} \bigg].
\]
We further approximate $p(z|X_c, Y_c) \approx q(z|X_c, Y_c)$
and train the model by maximizing this expected lower-bound over tasks.

\gls{anp}~\cite{kim2018attentive} and its conditional version without a global latent variable, \gls{canp}, both employ an attention mechanism~\cite{vaswani2017attention} to resolve the issue of underfitting in the vanila \gls{np} model. 
The encoder in \gls{anp} utilizes self-attention and cross-attention operation to better summarize the context into a representation $\phi$.
Please refer to \cref{sup:sec:model_architectures} for a detailed description about the architectures.

\subsection{Bootstrap, Bagging, and BayesBag}
Let $X = \{x_i\}_{i=1}^n$ be a dataset and $\theta = F(X)$ a parameter to estimate. Bootstrap~\cite{efron1979bootstrap} is a method that estimates the sampling distribution of $\theta$ from multiple datasets resampled from $X$,
\[
\tX\ssc{j} \swrsim X, \quad \tilde \theta\ssc{j} = F(\tX\ssc{j}) \,\, \text{ for } j=1,\dots, k,
\]
where $\swrsim$ denotes sampling with replacement%
~\footnote{Unless specified otherwise, we sample the same number of elements as the original set: $|X|=|\tX\ssc{j}|$.}. 
We call each $\tX\ssc{j}$ a \emph{bootstrap dataset} and $\tilde\theta\ssc{j}$ a \emph{bootstrap estimate}. The bootstrap estimates are used for assessing uncertainty, computing credible intervals, or statistical testing. One can interpret the bootstrap estimates as samples from an (approximate) nonparametric and noninformative posterior of $\theta$~\cite[page 272]{hastie01statisticallearning}. 
Contrary to standard Bayesian methods that specify an explicit prior $p(\theta)$, bootstrapping is a more ``data-driven'' way of computing the uncertainty of $\theta$.

\textbf{B}ootstrap \textbf{agg}regat\textbf{ing} (bagging)~\cite{breiman1996bagging} is a procedure that ensembles multiple predictors given by bootstrap estimates. 
Let $T(\theta)$ be a predictor based on a parameter $\theta$, and $\{\tilde\theta\ssc{j}\}_{j=1}^k$ be bootstrap estimates. 
The bagging predictor is computed as $\frac{1}{k}\sum_{j=1}^k T(\tilde\theta\ssc{j})$. 
Bagging is known to improve accuracy and stability on classification and regression problems~\cite{breiman1996bagging}. 

Instead of point estimates $T(\theta)$, one can also apply bagging to \emph{Bayesian posteriors} $p(T(\theta)|X)$. 
BayesBag~\cite{Douady2003comparison,huggins2019using} ensembles posteriors $\{p(T(\theta)|\tX\ssc{j})\}_{j=1}^k$ computed from bootstrapped datasets to get an aggregated posterior $\frac{1}{k}\sum_{j=1}^k p(T(\theta)|\tX\ssc{k})$.
Compared to bagging, BayesBag provides similar or often better results even with fewer bootstrap datasets and is more robust under model-data mistmatch~\cite{huggins2019using}.

\subsection{Residual Bootstrap}
\label{subsec:residual_bootstrap}
Consider the bootstrap for regression, where a dataset is $(X, Y) = \{(x_i, y_i)\}_{i=1}^n$ and we want to estimate the distribution of the regression parameters $\theta$ or the predictive distribution $p(y|x,\theta)$. The most straightforward way is the paired bootstrap (empirical bootstrap) where we resample pairs of $(x, y)$ with replacement: $\{(\tilde x_i, \tilde y_i)\}_{i=1}^n \swrsim \{(x_i, y_i)\}_{i=1}^n$. Unfortunately, since the probability of a pair $(x_i, y_i)$ being excluded in $(\tX, \tY)$ is approximately $(1-n^{-1})^n \overset{n\to\infty}{\to} 0.368$, influential observations are often discarded, degrading the predictive accuracy.

Another option is the \emph{residual bootstrap} which fixes $X$ and only resamples the residuals of predictions. Consider a nonparametric regression setting with prediction $\mu_i$, variance $\sigma^2_i$, and additive residual $\varepsilon_i$ ($\mu_i$ and $\sigma_i$ are functions of $x_i$), i.e., $y_i = \mu_i + \sigma_i \varepsilon_i$.
Then, the bootstrap datasets are resampled as
\begin{enumerate}
    \item \label{item:compute_res} Fit a model with $(X, Y)$ to obtain $\{(\mu_i, \sigma_i)\}_{i=1}^n$ and compute the residual $\varepsilon_i = \frac{y_i- \mu_i}{\sigma_i}$.
    \item \label{item:resboot_fit} Let $\calE = \{\varepsilon_i\}_{i=1}^n$, For $j=1,\dots, k$,
    \begin{enumerate}
        \item Resample the residuals: $\tilde\varepsilon\ssc{j}_1,\dots,\tilde\varepsilon\ssc{j}_n \swrsim \calE$.
        \item Construct a bootstrap dataset: for $i=1,\dots, n$, $\tx\ssc{j}_i = x_i,\,\, \ty\ssc{j}_i = \mu_i + \sigma_i \tilde\varepsilon\ssc{j}$.
    \end{enumerate}
\end{enumerate}

The residual bootstrap resolves the issue of missing $x$ in bootstrap datasets, which is why they are often recommended for regression problems. 
We focus on using the residual bootstrap for our purpose, but one may also consider alternative bootstrap variants (e.g., wild bootstrap, parametric bootstrap) to resample datasets.
\section{Bootstrapping Neural Processes}
\label{sec:bnp}
\begin{figure}
    \centering
    {\includegraphics[width=0.40\linewidth]{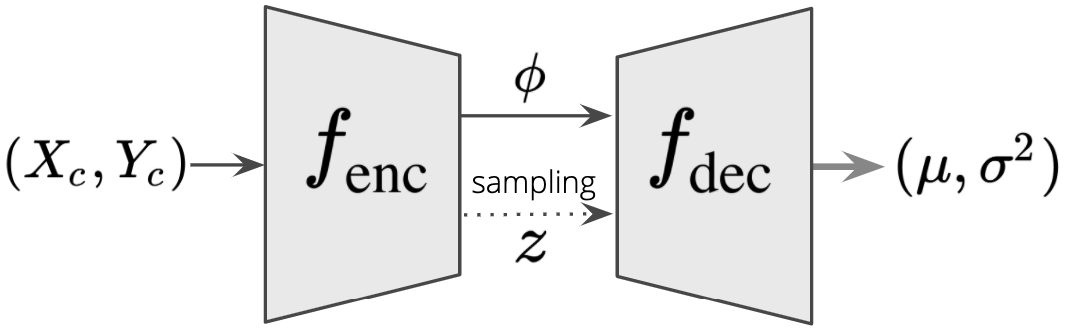}}
    \hspace{3pt}
    \includegraphics[width=0.50\linewidth]{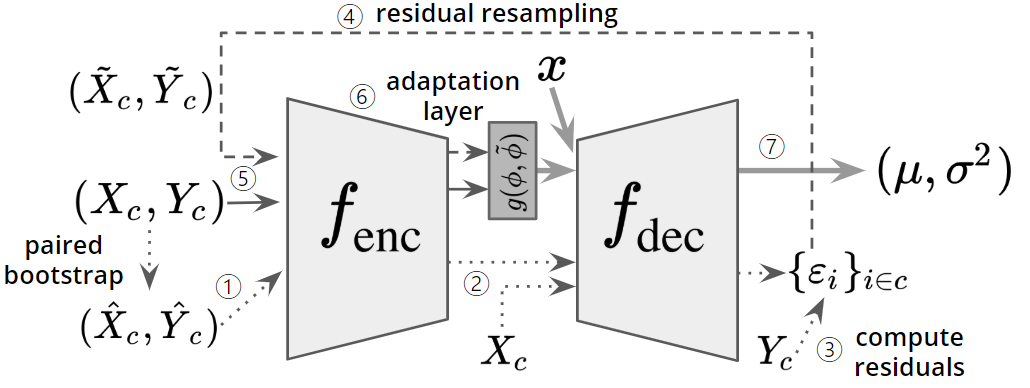}
    \caption{Diagrams for \gls{np} (left) and \gls{bnp} (right).}
    \label{fig:bnp_diagram}
\end{figure}

\subsection{Na\"ive application of residual bootstrap to \texorpdfstring{\gls{np}}{np} does not work}
\label{subsec:naive}
One may consider directly applying residual bootstrap to existing \gls{np} models. That is, given a task $\calT = (X, Y, c)$ and a \gls{np} model trained ordinarily, we can directly apply the residual bootstrap procedure described in \cref{subsec:residual_bootstrap} to get bootstrap contexts, and then compute bagged predictions by forwarding the bootstrap contexts through the \gls{np} model. \gls{np} is especially well-suited to this procedure because of its \emph{amortization} in the inference step -- it computes conditional probability $p(y|x, X_c, Y_c)$ efficiently as forward passes through neural networks.
However, unfortunately, we found this works poorly in terms of predictive accuracy~(\cref{sup:tab:1d_regression_ablation}). 
This may be because 1) the amortization is suboptimal, making the errors from fitting multiple bootstrap datasets accumulate, and 2) the \gls{np} model does not see bootstrap datasets during training, so feeding bootstrap datasets through the network acts like a model-data mismatch scenario that can fool the model.

\subsection{Bootstrapping Neural Processes}
\glsreset{bnp}
Beyond na\"ively applying bootstrap to \gls{np}, we propose a novel class of \gls{np} called \gls{bnp} which explicitly uses bootstrap datasets as additional inputs to induce functional uncertainty. 
\gls{bnp} uses the \gls{np} as its ``base'' model, and the extension to \gls{anp} which we name \gls{banp} is defined similarly. Let $f_\text{enc}$ and $f_\text{dec}$ be encoder and decoder of a base \gls{np} (defined as in \eqref{eq:cnp_architecture}), and $\calT = (X, Y, c)$ be a task. \gls{bnp} computes predictions through the following steps.

\paragraph{Resampling contexts via paired bootstrap} Before proceeding to residual bootstrap, we first resample the contexts from $(X_c, Y_c)$ via paired bootstrap, that is, for $j=1,\dots, k$,
\[\label{eq:bnp_step1}
(\hat X\ssc{j}, \hat Y\ssc{j}) := \{(\hat x_i\ssc{j}, \hat y_i\ssc{j})\}_{i=1}^{|c|} \swrsim \{(x_i, y_i)\}_{i\in c}.
\]
As noted in \cref{subsec:residual_bootstrap}, some resampled context $(\hat X\ssc{j}_c, \hat Y\ssc{j}_c)$ may miss several pairs from the original context. When passed to the model, such context would produce bad predictors, and thus large residuals. We empirically found that instead of computing single residuals computed from the full context $(X_c, Y_c)$ as in ordinary residual bootstrap, computing residuals from the multiple resampled contexts enhances robustness by exposing the model to residuals with diverse patterns during training. We present an ablation study comparing \gls{bnp} with and without this step in \cref{sup:tab:1d_regression_ablation}.

\paragraph{Residual bootstrap} Now we do the inference for the full context $(X_c, Y_c)$ using the resampled contexts $(\hat X_c\ssc{j}, \hat Y_c\ssc{j})$. As noted above, this can be done efficiently by forwarding $(\hat X_c\ssc{j}, \hat Y_c\ssc{j})$ through $f_\text{enc}$, $f_\text{dec}$ to get $\{(\hat\mu_i, \hat\sigma_i)\}_{i\in c}$. 
\[\label{eq:bnp_step2}
\hat\phi\ssc{j} = f_\text{enc}(\hat X\ssc{j}, \hat Y\ssc{j}),\quad
(\hat\mu\ssc{j}_i, \hat\sigma\ssc{j}_i) = f_\text{dec}(x_i, \hat\phi\ssc{j}) \text{ for }i \in c.
\]
Following the residual bootstrap procedure, we first compute residual, resample them,
\[\label{eq:bnp_step3}
\varepsilon\ssc{j}_i = \frac{y_i-\hat\mu\ssc{j}_i}{\hat\sigma\ssc{j}_i} \text{ for }i \in c,
\,\,\, \calE\ssc{j} = \{\varepsilon\ssc{j}_i\}_{i=1}^c, 
\,\,\, \tilde\varepsilon\ssc{j}_1,\dots, \tilde\varepsilon\ssc{j}_{|c|} \swrsim \calE\ssc{j}.
\]
and construct bootstrap contexts to be used for the final prediction.
\[\label{eq:bnp_step4}
&\tx\ssc{j}_i = x_i, \,\,\, \ty\ssc{j}_i = \hat\mu\ssc{j}_i + \hat\sigma\ssc{j}_i \tilde\varepsilon\ssc{j}_i \text{ for } i \in c, \nonumber\\
&(\tX_c\ssc{j}, \tY_c\ssc{j}) := \{(\tx_i\ssc{j}, \ty_i\ssc{j})\}_{i\in c} \text{ for } j=1,\dots, k.
\]

\paragraph{Encoding with adaptation layer}
We pass the bootstrap contexts into the encoder to get the representations of the contexts, $\tilde\phi\ssc{j} = f_\text{enc}(\tX\ssc{j}_c, \tY\ssc{j}_c) \text{ for } j=1,\dots, k$. The ordinary residual bootstrap would put each $\tilde\phi\ssc{j}$ into the decoder and ensemble the decoded conditional probabilities. Instead, like \gls{np} using both deterministic representation $\phi$ and global latent variable $z$, we put both the representation of the original context $\phi = f_\text{enc}(X_c,Y_c)$ and the bootstrapped representation $\tilde\phi\ssc{j}$ into the decoder. Since the decoder $f_\text{dec}$ is built to take only $\phi$, we add an \emph{adaptation layer} $g(\phi, \tilde\phi\ssc{j})$ to let $f_\text{dec}$ process a combined representation. The adaptation layer is the only part that we add to the base model, and can be implemented with a single linear layer. We empirically demonstrated that the adaptation layer is crucial for accurate prediction~(\cref{sup:tab:1d_regression_ablation}).
 
\paragraph{Prediction}
Finally, we construct predictions by ensembling the predictions decoded from the representations of bootstrap contexts. For a target point $x_i$,
\[
(\mu_i\ssc{j}, \sigma_i\ssc{j}) = f_\text{dec}(g(\phi, \tilde\phi\ssc{j}), x_i), \quad p(y_i|x_i, \phi, \tilde \phi\ssc{j}) = \calN(y_i | \mu_i\ssc{j}, (\sigma_i\ssc{j})^2).
\]
We compute this for $j=1,\dots, k$ to get an ensembled distribution,
\[\label{eq:bnp_bagged_conditional}
p(y_i|x_i, X_c, Y_c) \approx \frac{1}{k}\sum_{j=1}^k \calN(y_i|\mu_i\ssc{j}, (\sigma_i\ssc{j})^2).
\]
\cref{fig:bnp_diagram} shows diagrams comparing \gls{np} and \gls{bnp}. \gls{bnp} uses almost the same architecture except for the adaptation layer, but goes through the encoding-decoding process twice (first to compute residuals  only using the base model, and second to compute prediction with the adaptation layer added).

\subsection{Training}
\gls{bnp} requires special care for training because we need to balance the training of the base model (without bootstrap) and the full model (with bootstrap). 
If we only train the full model, the decoder of the base model computing the residuals would produce inaccurate predictions yielding large residuals, making the full model likely to ignore the residual path during the early training stages. To resolve this, we train the model with a combined objective to simultaneously train two paths as follows,
\[\label{eq:bnp_training}
\bbE_{p(\calT)}\bigg[ \sum_{i=1}^n \Big( \log p_\text{base}(y_i|x_i, X_c, Y_c) + \log \frac{1}{k} \sum_{j=1}^k \calN(y_i | \mu_i\ssc{j}, (\sigma_i\ssc{j})^2\Big) \bigg],
\]
where $p_\text{base}(y_i|x_i, X_c,Y_c)$ denotes the conditional probability computed from the base model (see \cref{sup:tab:1d_regression_ablation} for the ablation study). We also found that training with multiple bootstrap contexts \eqref{eq:bnp_bagged_conditional} ($k > 1$) is crucial for robustness. We fixed $k=4$ for all of our experiments.

\subsection{Discussion}
\paragraph{Parallel computation}
An advantage of bootstrap and bagging is the ease in parallelization of fitting multiple bootstrap datasets. Our model also enjoys such benefits: we compute all steps \eqref{eq:bnp_step1}-\eqref{eq:bnp_step4} in parallel by packing multiple bootstrap contexts into a tensor and feeding it through networks.

\paragraph{Our model and BayesBag}
Note that we are computing the aggregated conditional probability \eqref{eq:bnp_bagged_conditional}, which is similar to how BayesBag computes the aggregated posterior. The difference is that we aggregate the approximate distributions computed with a shared neural network ($f_\text{enc}$ and $f_\text{dec}$) while BayesBag independently computes posteriors. Although the theory in \cite{huggins2019using} does not directly apply to \gls{bnp}, the underlying intuition may still be valid for our model: the predictions computed from BayesBag is more conservative (and thus robust) because it combines the model's uncertainty with the data-driven uncertainty coming from bootstrap.

\paragraph{Why should \gls{np} be robust?}
Although we do not have theoretical claims that explain our model's robustness, we have intuitive explanations for such properties. When a \gls{bnp} model encounters a substantial shift in data distribution, the base model will fail, resulting in larger residuals than usual. These larger residuals will be reflected in bootstrap contexts and thus into the representations $\tilde\phi\ssc{j}$. This encourages the model to produce more conservative (larger $\sigma_i^2$) results~(e.g, \cref{fig:gp_vis}).

\section{Related Works}
\label{sec:related_works}
Since the first model \gls{cnp}~\citep{garnelo2018conditional}, there have been several follow-up works to improve \gls{np} classes in various aspects. \gls{np}~\citep{garnelo2018neural}
suggested to use a global latent variable to model functional uncertainty. \gls{anp}~\citep{kim2018attentive} further improved the reconstruction quality by employing attention mechanism,
and \cite{le2018empirical} conducted comprehensive comparison and empirically concluded that having global latent variable helps. \cite{singh2019sequential,willi2019recurrent} extended \gls{np} to work for sequential data. \cite{louizos2019functional} proposed a consistent \gls{np} model mainly using graph neural networks to build conditional probabilities. \cite{gordon2020convolutional} proposed a translation-equivariant version of \gls{np} model using convolution operation in context encoding. 

Bootstrap and bagging have been used ubiquitously over many areas in statistical modeling and machine learning. 
We list a few recent works (especially in the deep learning era) that have benefited from bootstrap and related ideas. Deep ensemble~\citep{lakshminarayanan2017simple} is a special case of bagging (but resampling with replacement) and has been shown to improve predictive accuracy and robustness on various tasks. \cite{reed2015training} demonstrated that bootstrapping can improve classification performance on noisy or incomplete labels. \cite{osband2016deep} showed that bootstrapping can improve exploration in deep reinforcement learning. \cite{nalisnick2017amortized}, which proposed the amortized bootstrap, is probably the most similar work to ours. They learn an implicit distribution that generates bootstrap estimates of a parameter of interest, and they show that bagging the bootstrap estimates generated from learned distribution outperforms ordinary bagging. The difference is that the amortized bootstrap targets a single task, meaning that they only learn an implicit bootstrap distribution for a single dataset. On the other hand, \gls{bnp} meta-learns a network that performs bootstrapping and bagging for any dataset from a particular task distribution.

\section{Experiments}
\label{sec:experiments}
In this section, we compare the baseline \gls{np} classes (\gls{cnp}, \gls{np}, \gls{canp}, and \gls{anp}) to our models (\gls{bnp}, \gls{banp}) on both synthetic and real-world datasets. 
We also compare ours against \gls{de} of \gls{cnp} and \gls{canp}~\citep{lakshminarayanan2017simple}, in which five identical models are trained with different random initializations and data streams, and
averaged for prediction.~\footnote{One could also consider \gls{de} of \gls{np} or \gls{bnp}, but here we want to compare the net effect of \gls{de} without any other source of uncertainty.} Following \cite{kim2018attentive}, we measured the \emph{context likelihood} $\frac{1}{|c|}\sum_{i\in c} \log p(y_i|x_i,X_c, Y_c)$ measuring the reconstruction quality of the contexts and \emph{target likelihood} $\frac{1}{n-|c|}\sum_{i \notin c} \log p(y_i|x_i, X_c, Y_c)$ measuring the prediction accuracy.
\gls{np}, \gls{anp}, \gls{bnp}, and \gls{banp} were trained with $k=4$ samples ($z$ for \gls{np} and \gls{anp}, and bootstrap contexts for \gls{bnp} and \gls{banp}) and tested with $k=50$ samples. 
Please refer to \cref{sup:sec:experimental_details} for further details.

\begin{table} \centering
\caption{1D regression results. ``context'' refers to context log-likelihoods, and ``target'' refers to target log-likelihoods. Means and standard deviations of five runs are reported.}
\vspace{5pt}
\resizebox{\textwidth}{!}{
\setlength{\tabcolsep}{3pt} 
\begin{tabular}{@{}c cc cc cc cc@{}}
\toprule
    &  \multicolumn{2}{c}{RBF} & \multicolumn{2}{c}{Mat\'ern 5/2} & \multicolumn{2}{c}{Periodic} & \multicolumn{2}{c}{$t$-noise}\\
\cmidrule[0.2pt]{2-9}  
    & context & target & context & target & context & target & context & target \\
\midrule
    CNP & \msd{0.972}{0.008} & \msd{0.448}{0.006} & \msd{0.846}{0.009} & \msd{0.206}{0.006} & \msd{-0.163}{0.008} & \msd{-1.747}{0.023} & \msd{0.363}{0.147} & \msd{-1.528}{0.068} \\
    NP & \msd{0.902}{0.009} & \msd{0.420}{0.008} & \msd{0.774}{0.012} & \msd{0.204}{0.010} & \msd{-0.181}{0.010} & \msd{-1.338}{0.025} & \msd{0.442}{0.016} & \msd{-0.792}{0.048} \\
    CNP+DE & 0.995 & 0.521 & 0.878 & \bf 0.313 & \bf -0.098 & -1.384 & 0.534 & -1.129\\
\cmidrule[0.2pt]{1-9} 
    BNP & \msd{\bf 1.013}{0.007} & \msd{\bf 0.526}{0.005} & \msd{\bf 0.890}{0.009} & \msd{\bf 0.317}{0.006} & \msd{-0.112}{0.007} & \msd{\bf -1.082}{0.011} & \msd{\bf 0.553}{0.009} & \msd{\bf -0.630}{0.014} \\
\cmidrule[0.5pt]{1-9}
    CANP & \msd{\bf 1.379}{0.000} & \msd{0.838}{0.001} & \msd{1.376}{0.000} & \msd{0.652}{0.001} & \msd{0.476}{0.043} & \msd{-5.896}{0.134} & \msd{1.104}{0.009} & \msd{-2.243}{0.031} \\
    ANP & \msd{\bf 1.379}{0.000} & \msd{0.842}{0.002} & \msd{1.376}{0.000} & \msd{0.660}{0.001} & \msd{0.600}{0.034} & \msd{-4.357}{0.182} & \msd{1.125}{0.003} & \msd{\bf -1.776}{0.021} \\
    CANP+DE & 1.378 & 0.847 & 1.376 & 0.670 & \bf 0.771 & -4.598 & \bf 1.161 & -1.991 \\
\cmidrule[0.2pt]{1-9}
    BANP & \msd{\bf 1.379}{0.000} & \msd{\bf 0.851}{0.002} & \msd{1.376}{0.000} & \msd{\bf 0.672}{0.001} & \msd{0.705}{0.016} & \msd{\bf -3.275}{0.114} & \msd{1.142}{0.007} & \msd{\bf -1.718}{0.055} \\
\bottomrule 
\end{tabular}
}
\label{tab:1d_regression}
\end{table}
\begin{figure}[t]
    \centering
    \includegraphics[width=0.27\linewidth]{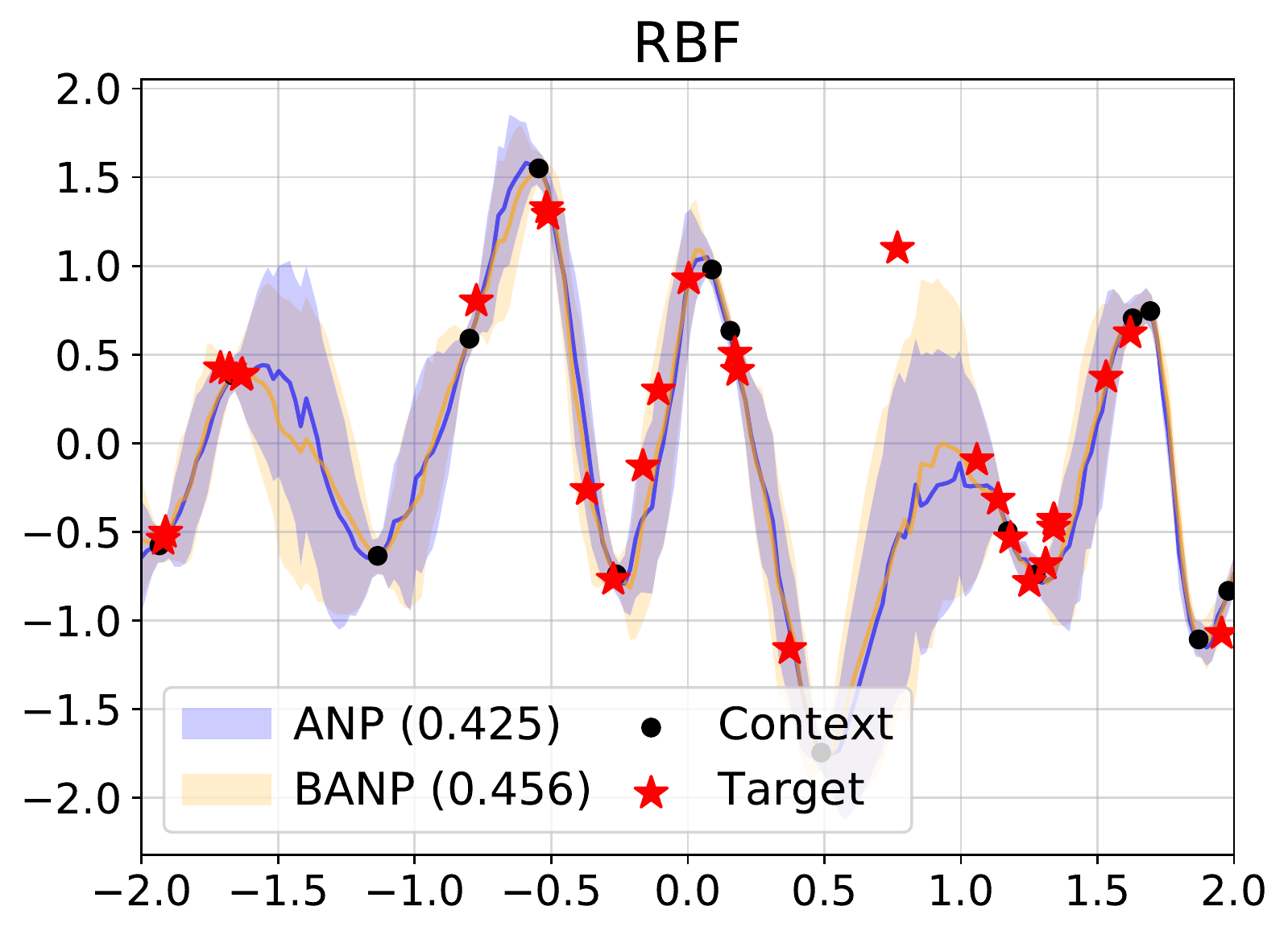}
    \includegraphics[width=0.27\linewidth]{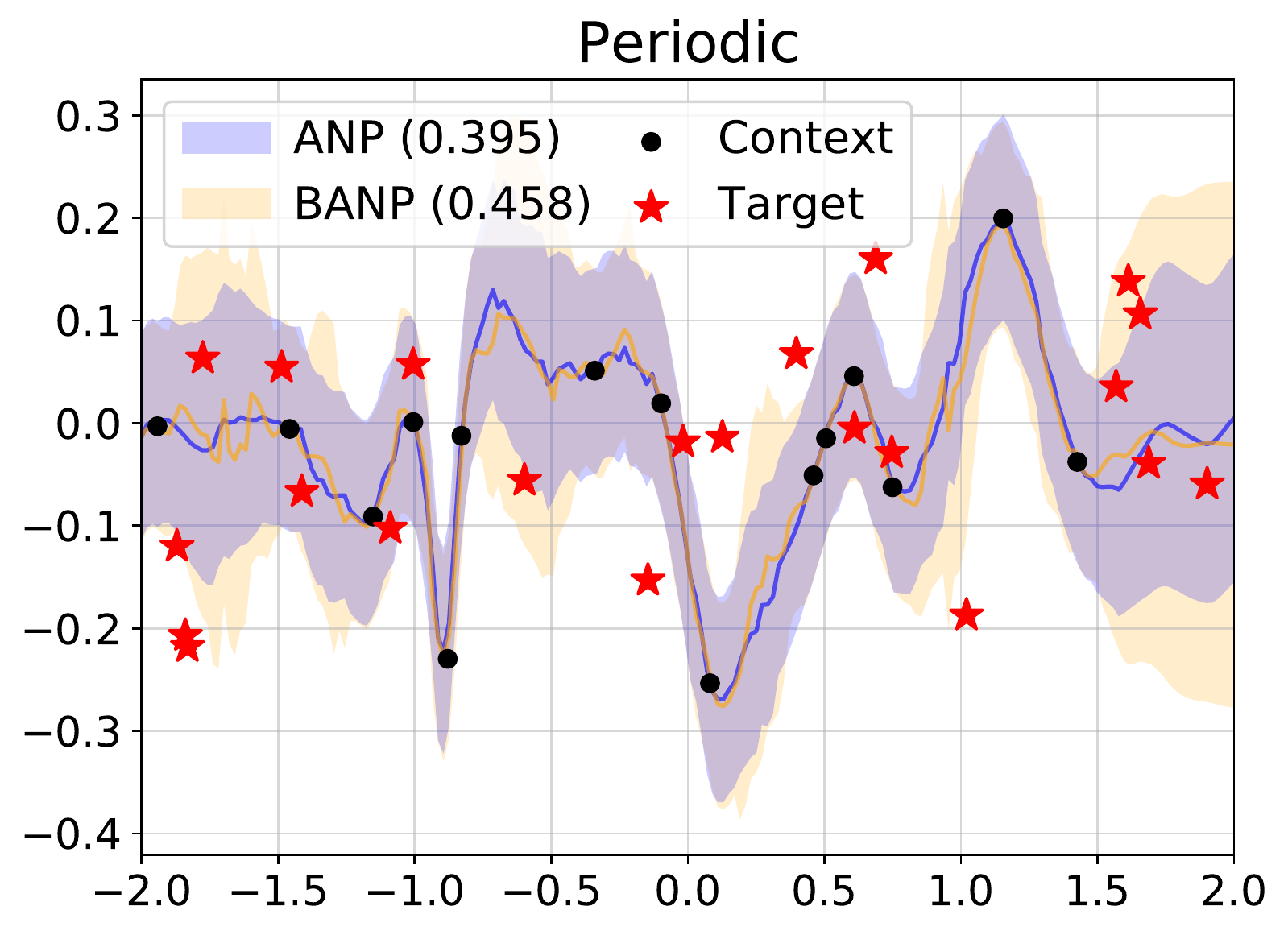}
    \includegraphics[width=0.27\linewidth]{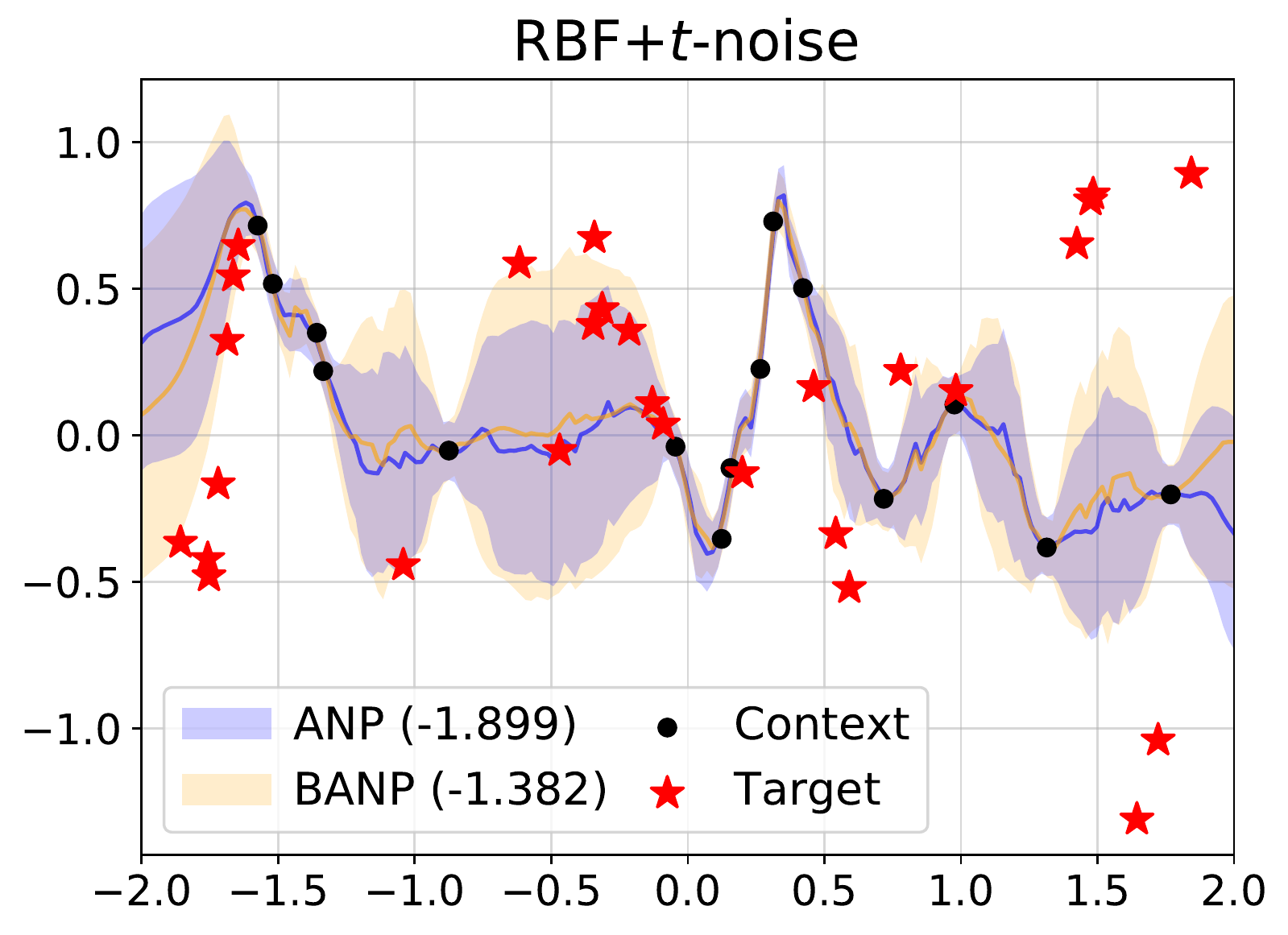}
    \caption{Visualization of \gls{anp} and \gls{banp} for 1D regression data. Ensembled means and $\pm$ standard deviations of 50 samples are displayed. The numbers in the legend denotes target log-likelihoods.}
    \label{fig:gp_vis}
\end{figure}

\begin{figure}[t]
    \centering
    \includegraphics[width=0.21\linewidth]{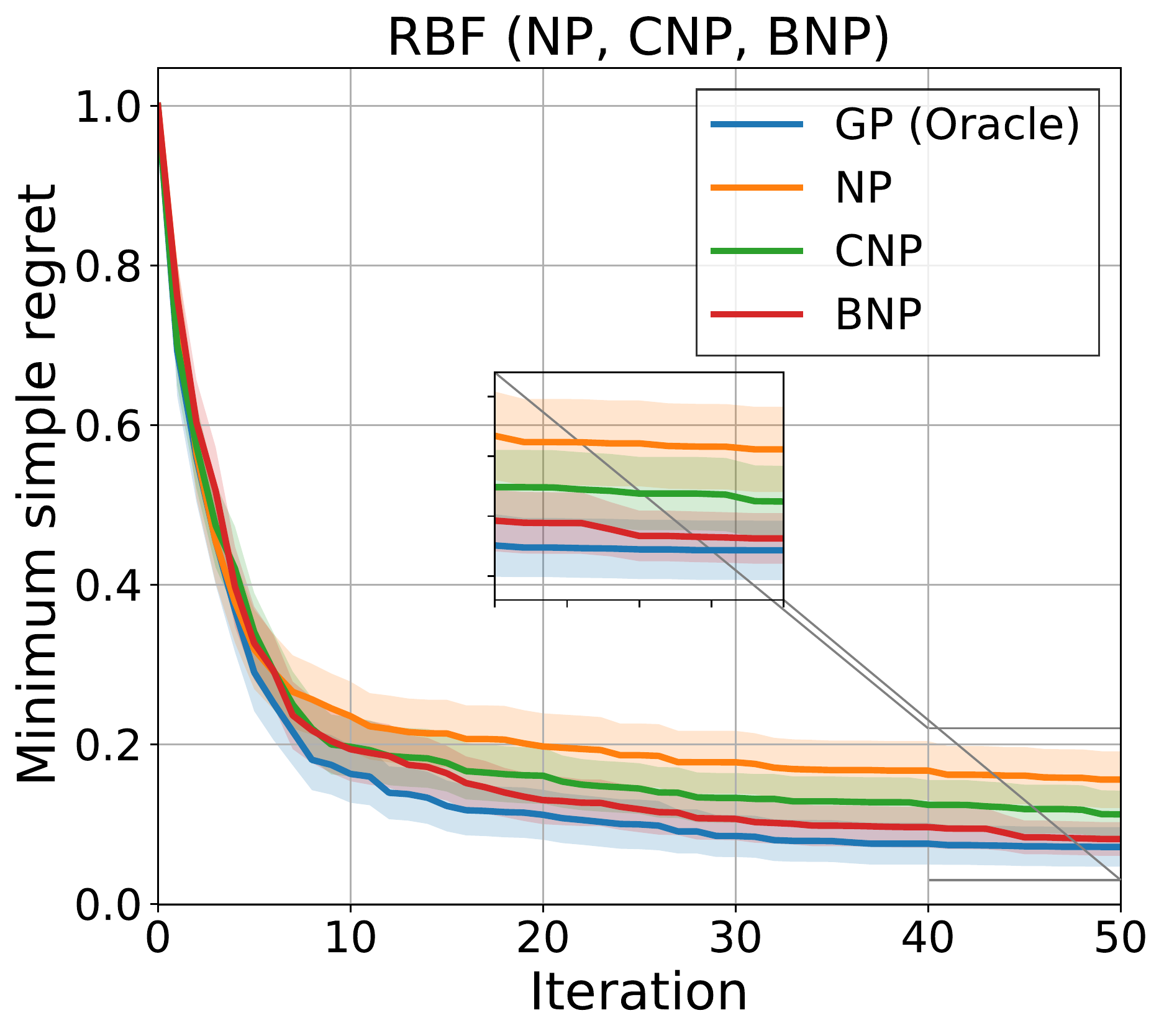}
    \includegraphics[width=0.21\linewidth]{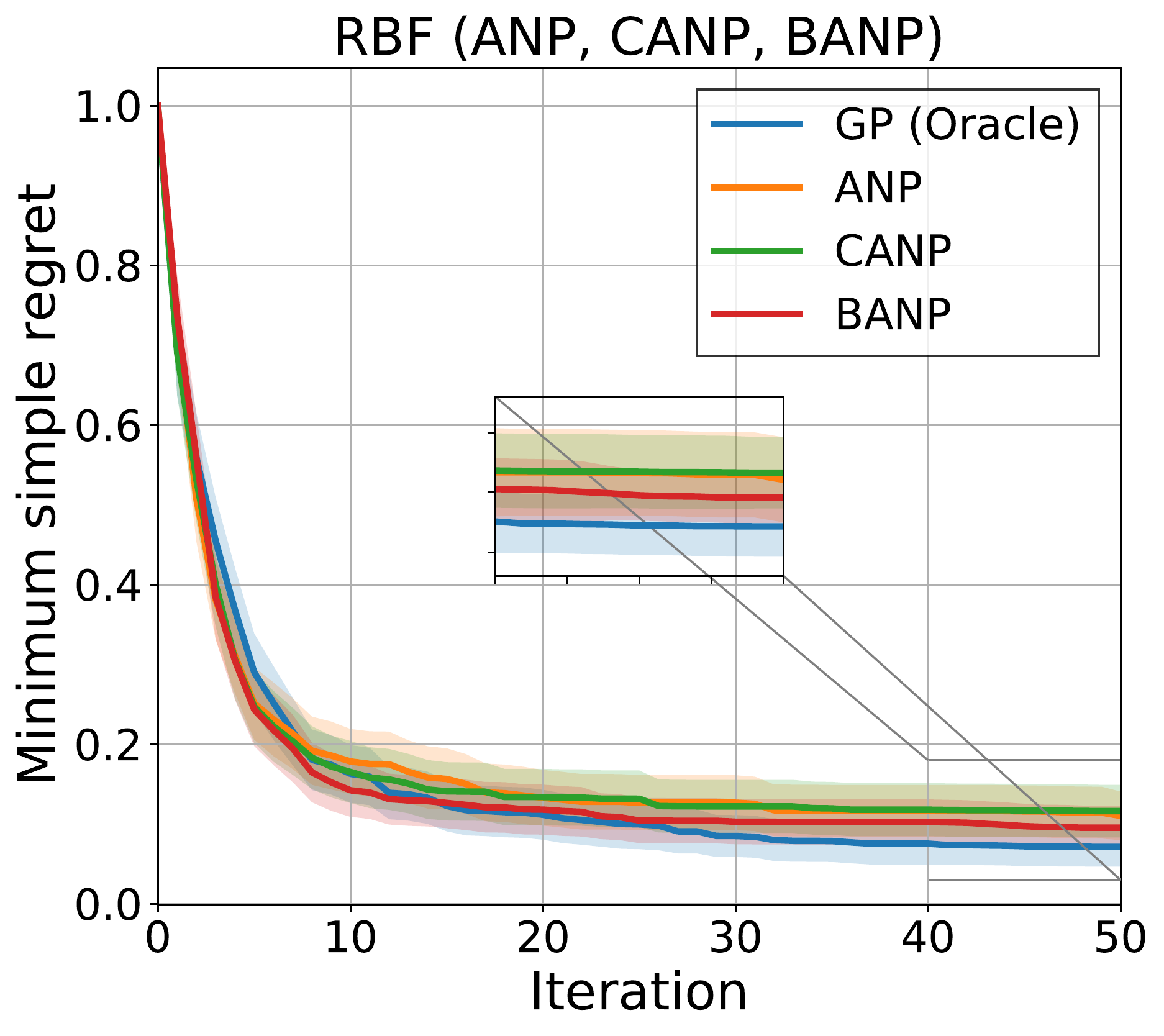}
    \includegraphics[width=0.21\linewidth]{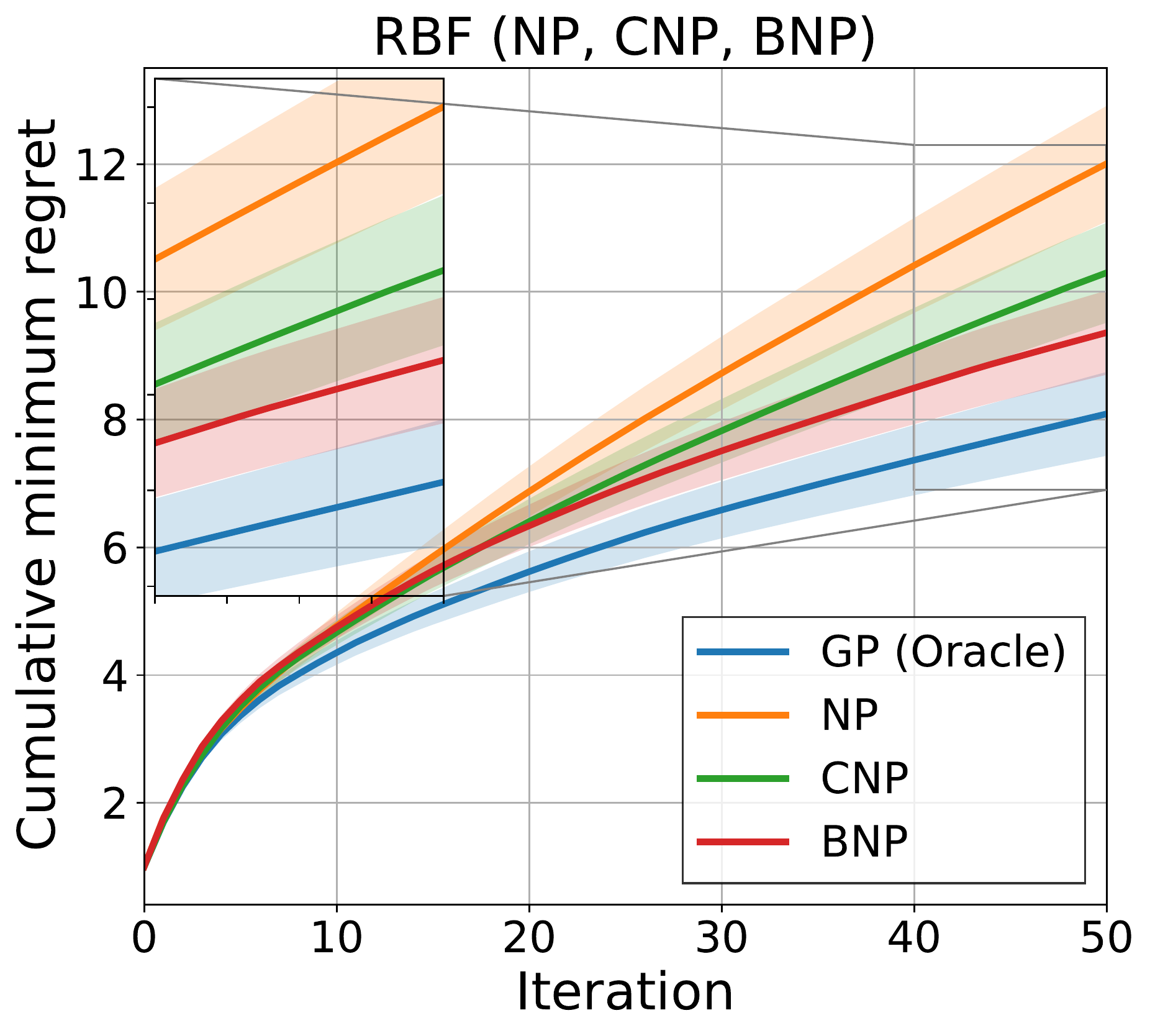}
    \includegraphics[width=0.21\linewidth]{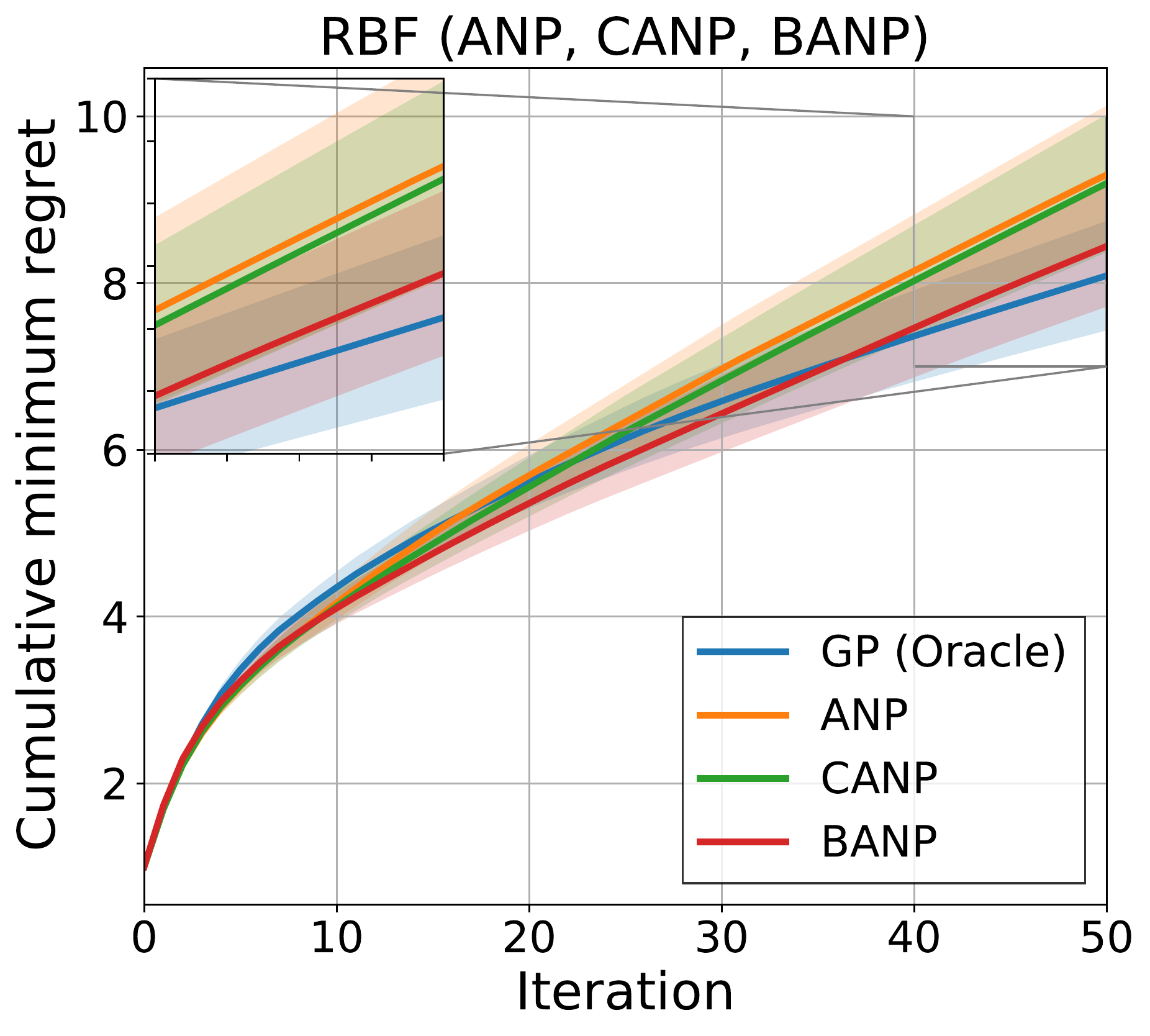}
    \caption{Bayesian optimization results for GP prior functions with RBF.}
    \label{fig:bo_rbf}
\end{figure}

\subsection{1D Regression}
\label{subsec:1d_regression}

We first conducted 1D regression experiments as in \cite{kim2018attentive}. We trained the models with curves generated from \gls{gp} with RBF kernels and tested in various settings, including model-data mismatch. More specifically, we tested the models trained with RBF kernel on the data generated from \gls{gp} with other types of kernels (Mat\'ern 5/2, Periodic), and
\gls{gp} with Student's $t$ noise added ($t$-noise). Please refer to \cref{sup:subsec:1d_regression} for a detailed description of network architectures, data generation, training, and testing. 
For a fair comparison, we set the models to use almost the same number of parameters. \cref{tab:1d_regression} summarizes the results. \gls{bnp} and \gls{banp} outperformed baselines and even \gls{de}, which has $5\times$ the number of parameters.
As expected, all models are less accurate in the model-data mismatch setting, but \gls{bnp} and \gls{banp} were affected less, demonstrating the robustness of our approach. \cref{fig:gp_vis} illustrates the behaviour of \gls{banp}: 
\gls{anp} and \gls{banp} show similar variances for ordinary test data (RBF), but for model-data mismatch data (periodic and $t$-noise), \gls{banp} produces wider variances than \gls{anp}. We further analyze this aspect by looking at calibrations and sharpness of the predictions in \cref{sup:sec:calib}.

\subsection{Bayesian Optimization}
\label{subsec:bayesian_optimization}
We evaluated the models trained in \cref{subsec:1d_regression} on Bayesian optimization~\citep{BrochuE2010arxiv} for functions generated from a \gls{gp} prior.
We reported the best simple regret, which represents the difference 
between the current best observation and the global optimum, 
and the cumulative best simple regret for $100$ sampled functions.
For consistent comparison, we fixed initializations and normalized the results.
Results in \cref{fig:bo_rbf} show that \gls{bnp} and \gls{banp} consistently 
achieve lower regret than other \gls{np} variants. See \cref{sup:subsec:bo_more} for more results including model-data mismatch settings.

\subsection{Image Completion}
\begin{table} \centering
\caption{EMNIST results. Means and standard deviations of 5 runs are reported.}
\vspace{2pt}
\resizebox{0.85\textwidth}{!}{
\setlength{\tabcolsep}{3pt}
\begin{tabular}{@{}c cc cc cc@{}}
\toprule
    &  \multicolumn{2}{c}{Seen classes (0-9)} & \multicolumn{2}{c}{Unseen classes (10-46)} & \multicolumn{2}{c}{$t$-noise}\\
\cmidrule[0.2pt]{2-7} 
    & context & target & context & target & context & target\\
\midrule
    CNP & \msd{0.926}{0.007} & \msd{0.751}{0.005} & \msd{0.766}{0.009} & \msd{0.498}{0.012} & \msd{-0.288}{0.140} & \msd{-0.478}{0.129} \\
    NP & \msd{0.948}{0.006} & \msd{0.806}{0.005} & \msd{0.808}{0.005} & \msd{0.600}{0.009} & \msd{0.071}{0.042} & \msd{-0.146}{0.034} \\
    CNP+DE & 0.954 & 0.813 & 0.818 & 0.616 & \bf 0.107 & \bf -0.020 \\
    \cmidrule[0.2pt]{1-7}
    BNP & \msd{\bf 1.004}{0.008} & \msd{\bf 0.880}{0.005} & \msd{\bf 0.883}{0.010} & \msd{\bf 0.722}{0.006} & \msd{-0.027}{0.069} & \msd{\bf 0.003}{0.037} \\
\cmidrule[0.5pt]{1-7}
    CANP & \msd{1.383}{0.000} & \msd{0.950}{0.004} & \msd{1.382}{0.000} & \msd{0.834}{0.002} & \msd{0.133}{0.196} & \msd{-0.492}{0.108} \\
    ANP & \msd{1.383}{0.000} & \msd{0.993}{0.005} & \msd{\bf 1.383}{0.000} & \msd{0.894}{0.004} & \msd{0.249}{0.084} & \msd{-0.132}{0.029} \\
    CANP+DE &  1.383 & 0.976 & \bf 1.383 & 0.881 & 0.307 & -0.240 \\
\cmidrule[0.2pt]{1-7}
    BANP & \msd{1.383}{0.000} & \msd{\bf 1.010}{0.006} & \msd{1.382}{0.000} & \msd{\bf 0.942}{0.005} & \msd{\bf 0.524}{0.102} & \msd{\bf 0.124}{0.060} \\
\bottomrule          
\end{tabular}
}
\label{tab:emnist}
\end{table}

\begin{figure}
    \centering
    \includegraphics[width=0.42\linewidth]{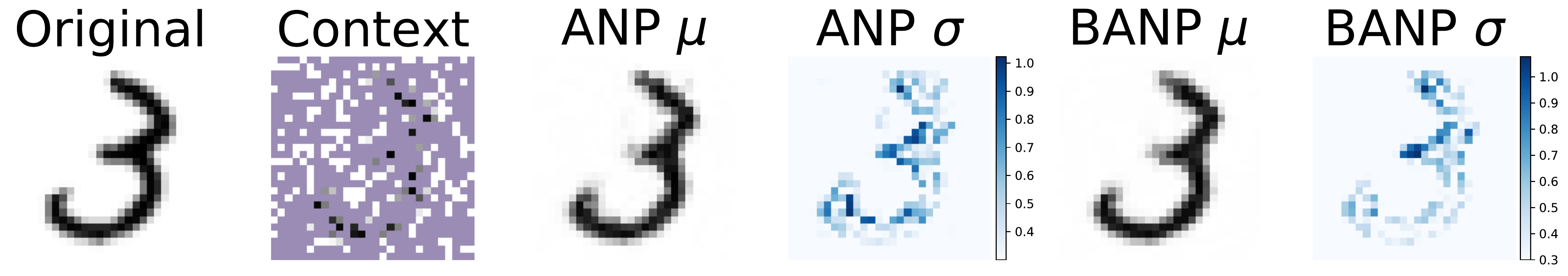} \hspace{1pt}
    \includegraphics[width=0.42\linewidth]{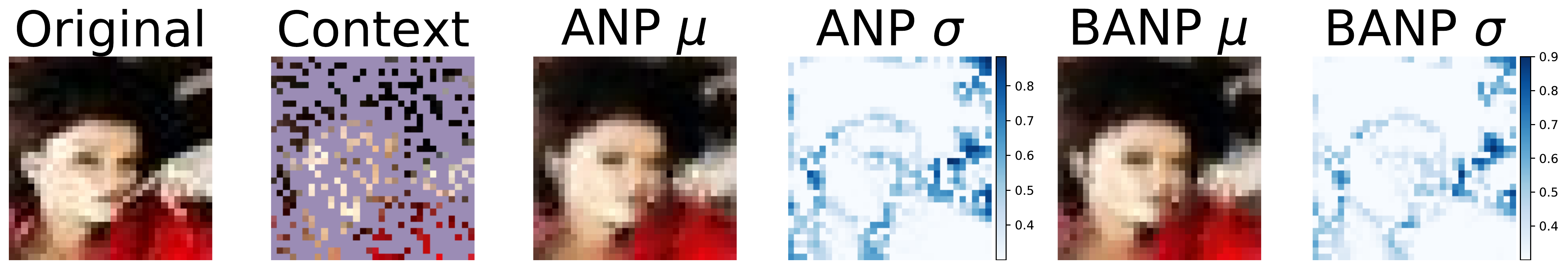}\\
    \includegraphics[width=0.42\linewidth]{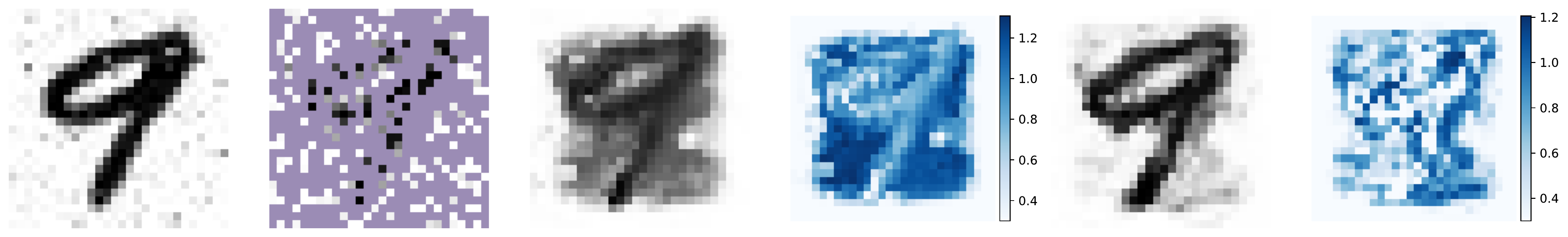} \hspace{1pt}
    \includegraphics[width=0.42\linewidth]{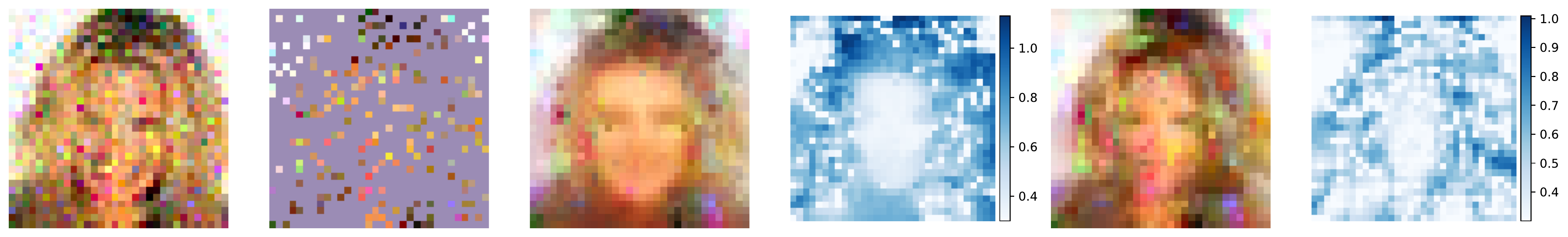}\\
    \includegraphics[width=0.42\linewidth]{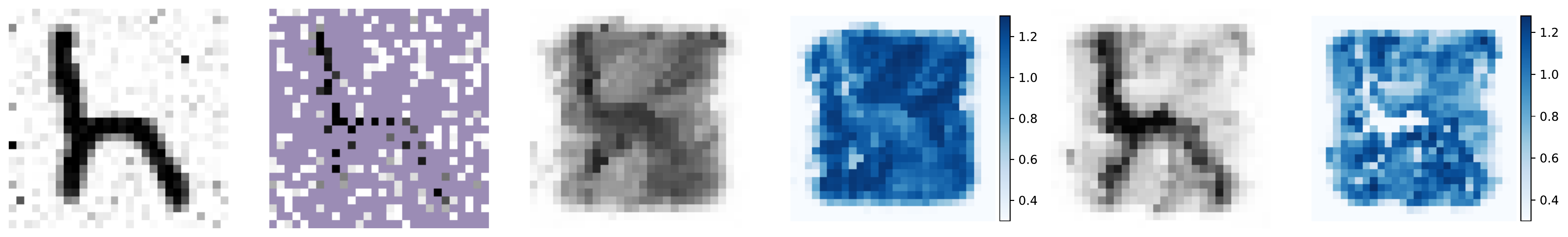} \hspace{1pt}
    \includegraphics[width=0.42\linewidth]{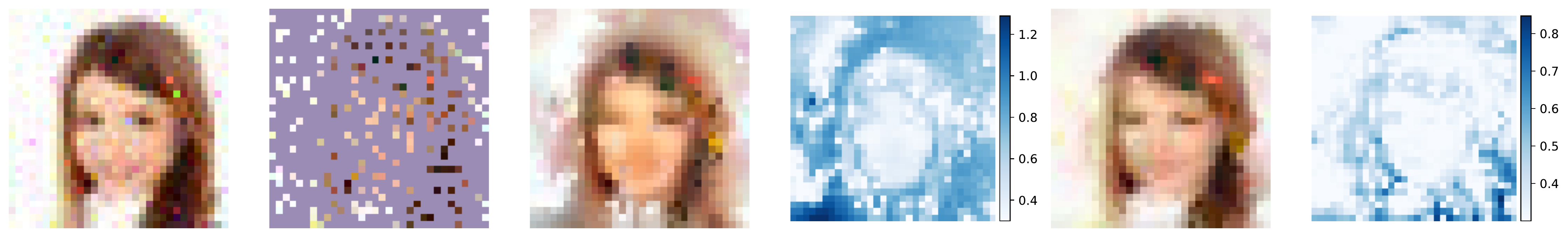}
    \caption{\gls{anp} vs \gls{banp} on EMNIST and CelebA32. The second and third row contains $t$-noises in the image. Ensembled means and standard deviations of 50 samples are displayed.}
    \label{fig:image_completion_vis}
\end{figure}

We compared the models on image completion tasks on EMNIST~\citep{cohen2017emnist} and CelebA~\citep{liu2015faceattributes} (resized to 32$\times$32). 
We followed the setting in \cite{garnelo2018neural,kim2018attentive}; see \cref{sup:subsec:image_completion} for details. 
As a model-data mismatch setting, we trained the models for EMNIST using the first $10$ classes and tested on the remaining $37$ classes. 
We also tested the setting for which Student's $t$-noise were added to the pixel values.  We summarize results in \cref{tab:emnist} and \cref{tab:celeba}. Except for \gls{bnp} for EMNIST with $t$-noise setting, ours outperformed the baselines. 
\cref{fig:image_completion_vis} compares the completion results of \gls{anp} and \gls{banp}.  \gls{anp} often breaks down with noise, while \gls{banp} successfully recovers the shapes of objects in images with less blur.

\subsection{Predator-Prey Model}
\label{subsec:predator-prey}

\begin{table}[t]
\centering
    \begin{minipage}{0.495\linewidth} \centering
    \caption{CelebA32 results.}
    \resizebox{\textwidth}{!}{ \setlength{\tabcolsep}{2pt}
        \begin{tabular}{@{}c cc cc@{}}
    \toprule
      &  \multicolumn{2}{c}{Without noise} & \multicolumn{2}{c}{$t$-noise} \\
    \cmidrule[0.2pt]{2-5} &  context & target & context & target \\
    \midrule
CNP & \msd{2.975}{0.013} & \msd{2.199}{0.003} & \msd{0.350}{0.384} & \msd{-1.468}{0.329} \\
NP & \msd{3.066}{0.011} & \msd{2.492}{0.014} & \msd{0.005}{0.195} & \msd{-0.217}{0.104} \\
    CNP+DE & 3.082 & 2.426 & \bf 1.361 & -0.451 \\
    \cmidrule[0.2pt]{1-5}
BNP & \msd{\bf 3.269}{0.008} & \msd{\bf 2.788}{0.005} & \msd{1.224}{0.422} & \msd{\bf 0.454}{0.094} \\
    \cmidrule[0.5pt]{1-5}
CANP & \msd{\bf 4.150}{0.000} & \msd{2.731}{0.006} & \msd{2.985}{0.149} & \msd{-0.730}{0.045} \\
ANP & \msd{\bf 4.150}{0.000} & \msd{2.947}{0.007} & \msd{3.037}{0.102} & \msd{\bf -0.099}{0.150} \\
    CANP+DE & \bf 4.150 & 2.814 & \bf 3.401 & \bf -0.0466 \\
    \cmidrule[0.2pt]{1-5}
BANP & \msd{4.149}{0.000} & \msd{\bf 3.129}{0.005} & \msd{3.395}{0.078} & \msd{\bf 0.083}{0.126} \\
    \bottomrule          
    \end{tabular}

    }
    \label{tab:celeba}
    \end{minipage}
    \hfill
    \begin{minipage}{0.495\linewidth} \centering
    \caption{Predator-prey model results.}
    \resizebox{\textwidth}{!}{ \setlength{\tabcolsep}{2pt}
        \begin{tabular}{@{}c cc cc@{}}
    \toprule
      &  \multicolumn{2}{c}{Simulated} & \multicolumn{2}{c}{Real} \\
    \cmidrule[0.2pt]{2-5} &  context & target & context & target \\
    \midrule
CNP & \msd{0.088}{0.031} & \msd{-0.142}{0.028}  & \msd{-2.702}{0.007} & \msd{-3.013}{0.025} \\
NP & \msd{-0.002}{0.039} & \msd{-0.252}{0.036}  & \msd{-2.747}{0.019} & \msd{-3.057}{0.020} \\
CNP+DE & 0.176 & \bf -0.026 & -2.670 & \bf -2.952 \\
    \cmidrule[0.2pt]{1-5}
BNP & \msd{\bf 0.213}{0.045} & \msd{\bf -0.011}{0.041}  & \msd{\bf -2.654}{0.005} & \msd{\bf -2.942}{0.010} \\
    \cmidrule[0.5pt]{1-5}
CANP & \msd{2.573}{0.014} & \msd{1.819}{0.021}  & \msd{1.767}{0.089} & \msd{-8.007}{0.538} \\
ANP & \msd{2.582}{0.007} & \msd{1.828}{0.007}  & \msd{1.720}{0.257} & \msd{-7.809}{0.642} \\
CANP+DE & \bf 2.591 & \bf 1.874 & \bf 2.021 &  \bf -5.440 \\
    \cmidrule[0.2pt]{1-5}
BANP & \msd{2.586}{0.009} & \msd{1.855}{0.009}  & \msd{1.783}{0.156} & \msd{\bf -5.465}{0.278} \\
\bottomrule          
    \end{tabular}

    }
    \label{tab:predator_prey}
    \end{minipage}
\end{table}

\begin{figure}[ht!]
    \centering
    \includegraphics[width=0.7\linewidth]{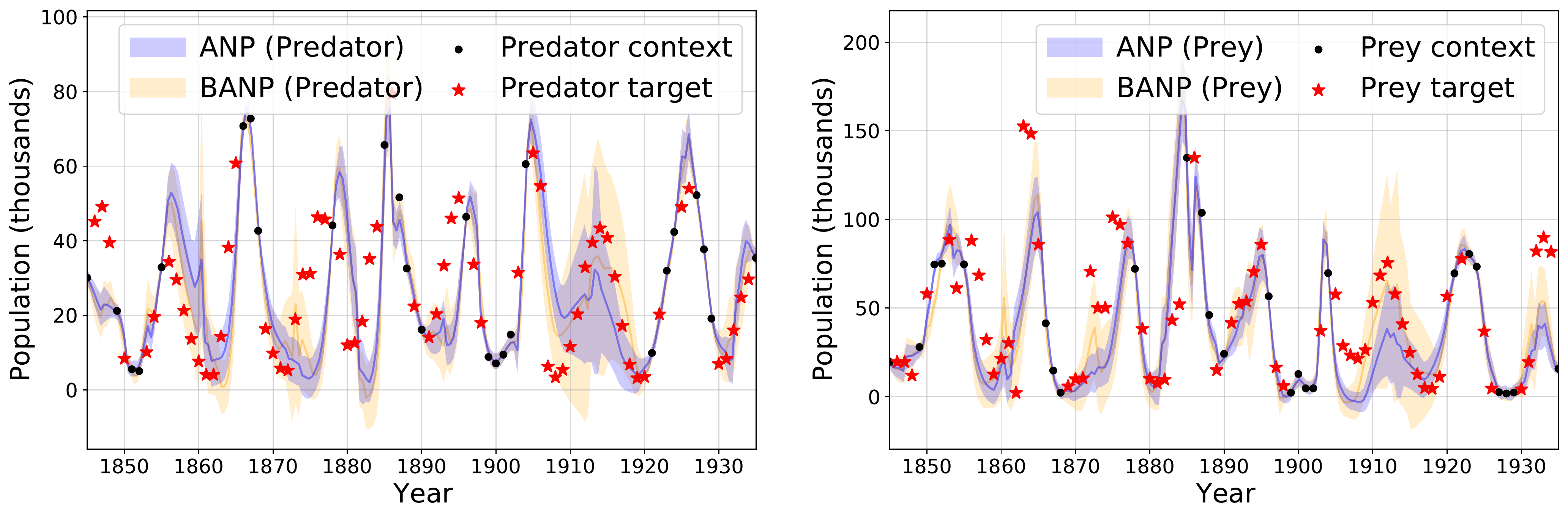}
    \caption{\gls{anp} vs \gls{banp} on Hudson's Bay hare (right)-lynx (left) data. Ensembled means and standard deviations of 50 samples are displayed.}
    \label{fig:predator_prey}
\end{figure}

Finally, following \cite{gordon2020convolutional}, we applied the models to predator-prey population data. We first trained the models using simulated data generated from a Lotka-Volterra model~\citep{wilkinson2011stochastic} and tested on real-world data (Hudson's Bay hare-lynx data).
As noted and empirically demonstrated in \cite{gordon2020convolutional}, the hare-lynx data is quite different from the simulated data, so it acts as a mismatch scenario.
The results are summarized in \cref{tab:predator_prey}.
We obtained the similar results as before; \gls{bnp} and \gls{banp} outperformed the baselines and were comparable to \gls{de} for both simulated and real-world data. 
\cref{fig:predator_prey} shows a similar trend as in \cref{fig:gp_vis}; \gls{banp} tends to be more conservative for mismatch data by producing larger variances.
\section{Conclusion}
\label{sec:conclusion}
In this paper, we proposed \gls{bnp}, a novel member of the \gls{np} family, which uses bootstrapping to induce functional uncertainty. We demonstrated that \gls{bnp} could successfully learn robust predictors, especially under model-data mismatch settings. Although not presented here, our model can be applied to any \gls{np} variants (or more) seamlessly. For instance, ours can readily be applied to recently proposed convolutional CNP~\citep{gordon2020convolutional}. As future work, one could consider developing a bootstrap resampling algorithm for more general settings. Here we presented an example of using residual bootstrap for regression, but this is not directly applicable for classification. Designing a framework that could ``learn'' to resample bootstrap datasets in a data-driven way would be an interesting and promising research direction. Finally, we want to stress that the idea of using bootstrap for inducing uncertainty may be useful for many other machine learning problems, especially the ones processing sets of data (e.g., \cite{zaheer2017deep}).

\section*{Broader Impact}
Uncertainty, robustness, interpretability in predictions have been important desiderata for machine learning algorithms, especially because we have seen actual incidents showing that the algorithms without those could lead to serious damage even threatening human life. The proposed approach suggests a way to enhance robustness by considering uncertainty in data distribution, and the idea of enhancing robustness via bootstrap can be applied to many algorithms over various fields. Therefore, we think that our paper potentially has a positive impact on many areas. Among the experiments we conducted, the predator-prey data experiment~(\cref{subsec:predator-prey}) shows this well, where data generated from a well-established model (Lotka-Volterra model) could be seriously different from real data (Hudson's Bay hare-lynx data), and our model could reduce the risk of failure in such case. However, we admit that the proposed approach may still be vulnerable to various scenario could happen in real life, so should not be treated as an absolute standard to follow. Our model just reduces the probability of failure in a more natural way (i.e., more ``data-driven'' way).

\begin{ack}
This work was supported by Engineering Research Center Program through the National Research Foundation of Korea (NRF) funded by the Korean Government MSIT (NRF-2018R1A5A1059921), Institute of Information \& communications Technology Planning \& Evaluation (IITP) grant funded by the Korea government (MSIT) (No.2019-0-00075), IITP grant funded by the Korea government(MSIT) (No.2017-0-01779, XAI) and the grant  funded  by  2019  IT Promotion  fund
(Development  of  AI  based  Precision Medicine  Emergency  System) of  the  Korea  government (Ministry of Science and ICT).
EY is also supported by Samsung Advanced Institute of Technology (SAIT). YWT's research leading to these results has received funding from the
European Research Council under the European Union's Seventh Framework
Programme (FP7/2007-2013) ERC grant agreement no. 617071.
\end{ack}

\bibliography{references}

\begin{thebibliography}{31}
\providecommand{\natexlab}[1]{#1}
\providecommand{\url}[1]{\texttt{#1}}
\expandafter\ifx\csname urlstyle\endcsname\relax
  \providecommand{\doi}[1]{doi: #1}\else
  \providecommand{\doi}{doi: \begingroup \urlstyle{rm}\Url}\fi

\bibitem[Ba et~al.(2016)Ba, Kiros, and Hinton]{ba2016layer}
J.~L. Ba, J.~R. Kiros, and G.~E. Hinton.
\newblock Layer normalization.
\newblock \emph{arXiv preprint arXiv:1607.06450}, 2016.

\bibitem[Breiman(1996)]{breiman1996bagging}
L.~Breiman.
\newblock Bagging predictors.
\newblock \emph{Machine Learning}, 24\penalty0 (2):\penalty0 123--140, 1996.

\bibitem[Brochu et~al.(2010)Brochu, Cora, and {de Freitas}]{BrochuE2010arxiv}
E.~Brochu, V.~M. Cora, and N.~{de Freitas}.
\newblock A tutorial on {Bayesian} optimization of expensive cost functions,
  with application to active user modeling and hierarchical reinforcement
  learning.
\newblock \emph{{arXiv} preprint {arXiv}:1012.2599}, 2010.

\bibitem[Burda et~al.(2015)Burda, Grosse, and
  Salakhutdinov]{burda2015importance}
Y.~Burda, R.~Grosse, and R.~Salakhutdinov.
\newblock Importance weighted autoencoders.
\newblock In \emph{International Conference on Learning Representations
  (ICLR)}, 2015.

\bibitem[Cohen et~al.(2017)Cohen, Afshar, and van Schaik]{cohen2017emnist}
G.~Cohen, J.~Afshar, S. adn~Tapson, and A.~van Schaik.
\newblock {EMNIST}: an extension of {MNIST} to handwritten letters.
\newblock \emph{arXiv preprint arXiv:1702.05373}, 2017.

\bibitem[Douady et~al.(2003)Douady, Delsuc, Boucher, Doolittle, and
  P.]{Douady2003comparison}
C.~J. Douady, F.~Delsuc, Y.~Boucher, W.~F. Doolittle, and D.~E.~J. P.
\newblock Comparison of {Bayesian} and maximum likelihood bootstrap measures of
  phylogenetic reliability.
\newblock \emph{Molecular Biology and Evolution}, 20\penalty0 (2):\penalty0
  248--254, 2003.

\bibitem[Edwards and Storkey(2016)]{edwards2016towards}
H.~Edwards and A.~Storkey.
\newblock Towards a neural statistician.
\newblock In \emph{International Conference on Learning Representations
  (ICLR)}, 2016.

\bibitem[Efron(1979)]{efron1979bootstrap}
B.~Efron.
\newblock Bootstrap methods: another look at the jackknife.
\newblock \emph{Annals of Statistics}, 7\penalty0 (1):\penalty0 1--26, 1979.

\bibitem[Garnelo et~al.(2018{\natexlab{a}})Garnelo, Rosenbaum, Maddison,
  Ramalho, Saxton, Shanahan, Teh, Rezende, and Eslami]{garnelo2018conditional}
M.~Garnelo, D.~Rosenbaum, C.~J. Maddison, T.~Ramalho, D.~Saxton, M.~Shanahan,
  Y.~W. Teh, D.~J. Rezende, and S.~M.~A. Eslami.
\newblock Conditional neural processes.
\newblock In \emph{International Conference on Machine Learning (ICML)},
  2018{\natexlab{a}}.

\bibitem[Garnelo et~al.(2018{\natexlab{b}})Garnelo, Schwarz, Rosenbaum, Viola,
  Rezende, Eslami, and Teh]{garnelo2018neural}
M.~Garnelo, J.~Schwarz, D.~Rosenbaum, F.~Viola, D.~J. Rezende, S.~M.~A. Eslami,
  and Y.~W. Teh.
\newblock Neural processes.
\newblock \emph{ICML Workshop on Theoretical Foundations and Applications of
  Deep Generative Models}, 2018{\natexlab{b}}.

\bibitem[Gordon et~al.(2020)Gordon, Bruinsma, Foong, Requeima, Dubois, and
  Turner]{gordon2020convolutional}
J.~Gordon, W.~P. Bruinsma, A.~Y.~K. Foong, J.~Requeima, Y.~Dubois, and R.~E.
  Turner.
\newblock Convolutional conditional neural processes.
\newblock In \emph{International Conference on Learning Representations
  (ICLR)}, 2020.

\bibitem[Hastie et~al.(2001)Hastie, Tibshirani, and
  Friedman]{hastie01statisticallearning}
T.~Hastie, R.~Tibshirani, and J.~Friedman.
\newblock \emph{The Elements of Statistical Learning}.
\newblock Springer Series in Statistics. Springer New York Inc., 2001.

\bibitem[Huggins and Miller(2019)]{huggins2019using}
J.~H. Huggins and J.~W. Miller.
\newblock Using bagged posteriors for robust inference and model criticism.
\newblock \emph{arXiv preprint arXiv:1912.07104}, 2019.

\bibitem[Jones et~al.(1998)Jones, Schonlau, and Welch]{JonesDR1998jgo}
D.~R. Jones, M.~Schonlau, and W.~J. Welch.
\newblock Efficient global optimization of expensive black-box functions.
\newblock \emph{Journal of Global Optimization}, 13:\penalty0 455--492, 1998.

\bibitem[Kim et~al.(2018)Kim, Mnih, Schwarz, Garnelo, Eslami, Rosenbaum, and
  Oriol]{kim2018attentive}
H.~Kim, A.~Mnih, J.~Schwarz, M.~Garnelo, S.~M.~A. Eslami, D.~Rosenbaum, and
  V.~Oriol.
\newblock Attentive neural processes.
\newblock In \emph{International Conference on Learning Representations
  (ICLR)}, 2018.

\bibitem[Kim and Choi(2017)]{KimJ2017bayeso}
J.~Kim and S.~Choi.
\newblock {bayeso}: A {Bayesian} optimization framework in {Python}.
\newblock \url{http://bayeso.org}, 2017.

\bibitem[Kingma and Ba(2015)]{kingma2015adam}
D.~P. Kingma and J.~Ba.
\newblock Adam: a method for stochastic optimization.
\newblock In \emph{International Conference on Learning Representations
  (ICLR)}, 2015.

\bibitem[Kuleshov et~al.(2018)Kuleshov, Fenner, and
  Ermon]{kuleshov2018accurate}
V.~Kuleshov, N.~Fenner, and S.~Ermon.
\newblock Accurate uncertainties for deep learning using calibrated regression.
\newblock In \emph{International Conference on Machine Learning (ICML)}, 2018.

\bibitem[Lakshminarayanan et~al.(2017)Lakshminarayanan, Pritzel, and
  Blundell]{lakshminarayanan2017simple}
B.~Lakshminarayanan, A.~Pritzel, and C.~Blundell.
\newblock Simple and scalable predictive uncertainty estimation using deep
  ensembles.
\newblock In \emph{Neural Information Processing Systems (NeurIPS)}, 2017.

\bibitem[Le et~al.(2018)Le, Kim, Garnelo, Rosenbaum, Schwarz, and
  Teh]{le2018empirical}
T.~A. Le, H.~Kim, M.~Garnelo, D.~Rosenbaum, J.~Schwarz, and Y.~W. Teh.
\newblock Empirical evaluation of neural process objectives.
\newblock \emph{NeurIPS Workshop on Bayesian Deep Learning}, 2018.

\bibitem[Liu et~al.(2015)Liu, Luo, Wang, and Tang]{liu2015faceattributes}
Z.~Liu, P.~Luo, X.~Wang, and X.~Tang.
\newblock Deep learning face attributes in the wild.
\newblock In \emph{Proceedings of International Conference on Computer Vision
  (ICCV)}, 2015.

\bibitem[Louizos et~al.(2019)Louizos, Shi, Schutte, and
  Welling]{louizos2019functional}
C.~Louizos, X.~Shi, K.~Schutte, and M.~Welling.
\newblock The functional neural process.
\newblock In \emph{Neural Information Processing Systems (NeurIPS)}, 2019.

\bibitem[Nalisnick and Smyth(2017)]{nalisnick2017amortized}
E.~Nalisnick and P.~Smyth.
\newblock The amortized bootstrap.
\newblock \emph{ICML 2017 Workshop on Implicit Models}, 2017.

\bibitem[Osband et~al.(2016)Osband, Blundell, Pritzel, and Roy]{osband2016deep}
I.~Osband, C.~Blundell, A.~Pritzel, and B.~V. Roy.
\newblock Deep exploration via bootstrapped {DQN}.
\newblock In \emph{Neural Information Processing Systems (NeurIPS)}, 2016.

\bibitem[Rasmussen and Williams(2006)]{RasmussenCE2006book}
C.~E. Rasmussen and C.~K.~I. Williams.
\newblock \emph{Gaussian Processes for Machine Learning}.
\newblock MIT Press, 2006.

\bibitem[Reed et~al.(2015)Reed, Lee, Anguelov, Szegedy, Erhan, and
  Rabinovich]{reed2015training}
S.~E. Reed, H.~Lee, D.~Anguelov, C.~Szegedy, D.~Erhan, and A.~Rabinovich.
\newblock Training deep neural networks on noisy labels with bootstrapping.
\newblock In \emph{International Conference on Learning Representations
  (ICLR)}, 2015.

\bibitem[Singh et~al.(2019)Singh, Yoon, Son, and Ahn]{singh2019sequential}
G.~Singh, J.~Yoon, Y.~Son, and S.~Ahn.
\newblock Sequential neural processes.
\newblock In \emph{Neural Information Processing Systems (NeurIPS)}, 2019.

\bibitem[Vaswani et~al.(2017)Vaswani, Shazeer, Parmar, Uszkoreit, Jones, Gomez,
  Kaiser, and Polosukhin]{vaswani2017attention}
A.~Vaswani, N.~Shazeer, N.~Parmar, J.~Uszkoreit, L.~Jones, A.~N. Gomez,
  L.~Kaiser, and I.~Polosukhin.
\newblock Attention is all you need.
\newblock In \emph{Neural Information Processing Systems (NeurIPS)}, 2017.

\bibitem[Wilkinson(2011)]{wilkinson2011stochastic}
D.~J. Wilkinson.
\newblock \emph{Stochastic modelling for systems biology}.
\newblock CRC Press, 2011.

\bibitem[Willi et~al.(2019)Willi, Masci, Schmidhuber, and
  Osendorfer]{willi2019recurrent}
T.~Willi, J.~Masci, J.~Schmidhuber, and C.~Osendorfer.
\newblock Recurrent neural processes.
\newblock \emph{arXiv preprint arXiv:1906.05915}, 2019.

\bibitem[Zaheer et~al.(2017)Zaheer, Kottur, Ravanbakhsh, Poczos, Salakhutdinov,
  and Smola]{zaheer2017deep}
M.~Zaheer, S.~Kottur, S.~Ravanbakhsh, B.~Poczos, R.~Salakhutdinov, and
  A.~Smola.
\newblock Deep sets.
\newblock In \emph{Neural Information Processing Systems (NeurIPS)}, 2017.

\end{thebibliography}

\clearpage
\appendix 
\begin{appendices}
\renewcommand{\thetable}{\Alph{section}.\arabic{table}}
\renewcommand{\thefigure}{\Alph{section}.\arabic{figure}}
\renewcommand{\theequation}{\Alph{section}.\arabic{equation}}
\setcounter{table}{0}
\setcounter{figure}{0}
\setcounter{equation}{0}

\section{Model Architectures}
\label{sup:sec:model_architectures}
\subsection{\texorpdfstring{\gls{cnp}}{cnp}, \texorpdfstring{\gls{np}}{np} and \texorpdfstring{\gls{bnp}}{bnp}}
\label{sup:subsec:np_architectures}
We borrowed most of the architectures from the paper~\citep{kim2018attentive} and their source code released~\footnote{\url{https://github.com/deepmind/neural-processes}}.
\paragraph{Encoder}
Let $\mathrm{MLP}(\ell, d_\text{in}, d_h, d_\text{out}), (\ell \geq 2)$  be a multilayer perceptron having the structure
\[
\mathrm{MLP}(\ell, d_\text{in}, d_h, d_\text{out}) &= \mathrm{Linear}(d_h, d_\text{out}) \nonumber\\
& \circ \underbrace{(\mathrm{ReLU}\circ\mathrm{Linear}(d_h, d_h) \circ \dots )}_{\times (\ell-2)} \nonumber\\
& \circ \mathrm{Linear}(d_h, d_\text{in}).
\]
An encoder of a \gls{np} consists of a deterministic path and a latent path using two identical structures (but with separate parameters),
\[
&h_1 = \frac{1}{|c|} \sum_{i \in c} \mathrm{MLP}(\ell_\text{pre}, d_x+d_y, d_h, d_h)([x_i, y_i]), \nonumber\\
&\phi = \mathrm{MLP}(\ell_\text{post}, d_h, d_h)(h_1), \quad f_\text{denc}(X_c, Y_c) = \phi \nonumber\\
&h_2 = \frac{1}{|c|} \sum_{i \in c} \mathrm{MLP}(\ell_\text{pre}, d_x+d_y, d_h, d_h)([x_i, y_i]), \nonumber\\
&(\eta, \rho') = \mathrm{MLP}(\ell_\text{post}, d_h, 2d_z)(h_2), \nonumber\\
&\rho' = 0.1 + 0.9 \cdot \mathrm{sigmoid}(\tilde\rho), \quad f_\text{lenc}(X_c, Y_c) = (\eta, \rho),
\]
where $d_x$ and $d_y$ are the dimensionalities of $x$ and $y$ respectively, and $d_h$ is fixed to 128 for all experiments.

An original \gls{cnp} uses only one deterministic encoder, but that would perform worse than \gls{np} because it uses twice less number of parameters. For a fair comparison,
we used two identical encoders for \gls{cnp} as well.
\[
&h_1 = \frac{1}{|c|} \sum_{i \in c} \mathrm{MLP}(\ell_\text{pre}, d_x+d_y, d_h, d_h)([x_i, y_i]), \nonumber\\
&\phi_1 = \mathrm{MLP}(\ell_\text{post}, d_h, d_h)(h_1)  \nonumber\\
&h_2 = \frac{1}{|c|} \sum_{i \in c} \mathrm{MLP}(\ell_\text{pre}, d_x+d_y, d_h, d_h)([x_i, y_i]), \nonumber\\
&\phi_2 = \mathrm{MLP}(\ell_\text{post}, d_h, d_h)(h_2)  \nonumber\\
&\phi = [\phi_1, \phi_2], \quad f_\text{enc}(X_c, Y_c) = \phi.
\]

\gls{bnp} uses exactly the same network encoder as \gls{cnp}. 

\paragraph{Decoder}
A decoder in \gls{cnp} and \gls{np} take a represerntation of a context and transform it to parameters of conditional probability. Let $x_*$ be a target data point. A decoder of \gls{cnp} is defined as
\[
&(\mu, \sigma') = \mathrm{MLP}(\ell_\text{dec}, 2d_h+d_x, d_h, 2d_y)([\phi, x_*]) \nonumber\\
&\sigma = 0.1 + 0.9 \cdot \mathrm{softplus}(\sigma'), \quad f_\text{dec}(\phi, x_*) = (\mu, \sigma).
\]
A decoder for \gls{np} uses excatly the same architecture except for that it takes $[\phi, z]$ instead.
\[
&(\mu, \sigma') = \mathrm{MLP}(\ell_\text{dec}, d_h+d_z+d_x, d_h, 2d_y)([\phi, z, x_*]) \nonumber\\
&\sigma = 0.1 + 0.9 \cdot \mathrm{softplus}(\sigma'), \quad f_\text{dec}(\phi, x_*) = (\mu, \sigma).
\]
\gls{bnp} uses the same decoder as \gls{cnp} when computing the deterministic representation without bootstrapping (base model). 
When decoding an aggregated representations from an original context $\phi$ and a bootstrapped context $\tilde\phi$, we add an adaptation layer to the first linear layer of the MLP.
\[
&h_1 = \mathrm{Linear}(2d_h + d_x, d_h)([\phi, x_*]) \nonumber\\
&h_2 = \mathrm{Linear}(2d_h, d_h)(\tilde \phi) \quad \text{(adaptation layer)}\nonumber\\
&(\mu, \sigma') = \mathrm{MLP}(\ell_\text{dec}-1, d_h, d_h, 2d_y)(\mathrm{ReLU}(h_1 + h_2)) \nonumber\\
&\sigma = 0.1 + 0.9 \cdot \mathrm{softplus}(\sigma'), \quad f_\text{dec}(\phi, \tilde\phi, x_*) = (\mu, \sigma).
\]

\subsection{\texorpdfstring{\gls{canp}}{canp}, \texorpdfstring{\gls{anp}}{anp} and \texorpdfstring{\gls{banp}}{banp}}
\paragraph{Encoder}
An encoder of \gls{anp} has a deterministic path and latent path. A deterministic path uses a self-attention and cross-attention to summarize contexts.
Let $\mathrm{MHA}(d_\text{out})$ be a multi-head attention~\citep{vaswani2017attention} comptued as follows:
\[
&Q' = \{\mathrm{Linear}(d_\text{q}, d_\text{out})(q)\}_{q \in Q}, \quad \{Q'_j\}_{j=1}^{n_\text{head}} = \mathrm{split}(Q', n_\text{head}) \nonumber\\
&K' = \{\mathrm{Linear}(d_\text{k}, d_\text{out})(k)\}_{k \in K}, \quad \{K'_j\}_{j=1}^{n_\text{head}} = \mathrm{split}(K', n_\text{head}) \nonumber\\
&V' = \{\mathrm{Linear}(d_\text{v}, d_\text{out})(v)\}_{v \in V}, \quad \{V'_j\}_{j=1}^{n_\text{head}} = \mathrm{split}(V', n_\text{head}) \nonumber\\
&H_j = \mathrm{softmax}( Q'_j (K_j')\tr / \sqrt{d_\text{out}} ) V'_j, \quad H = \mathrm{concat}( \{H_j\}_{j=1}^{n_\text{head}}) \nonumber\\
&H' = \mathrm{LN}(Q' + H) \nonumber\\
&\mathrm{MHA}(d_\text{out})(Q, K, V) = \mathrm{LN}(H' + \mathrm{ReLU}(\mathrm{Linear}(d_\text{out}, d_\text{out}))).
\]
Here, $(q_\text{k}, q_\text{k}, q_\text{v})$ denotes the dimensionalities of query $Q$, key $K$, and value $V$ respectively, $d_\text{out}$ is an output dimension,
$n_\text{head}$ is a number of heads, $\mathrm{split}$ and $\mathrm{concat}$ are splitting and concatenating operation in feature axis, and $\mathrm{LN}$ is the layer normalization~\citep{ba2016layer}.
A self-attention is defined as simply tying $Q=K=V$, $\mathrm{SA}(d_\text{out})(X) = \mathrm{MHA}(d_\text{out})(X, X, X)$.
A deterministic path of \gls{anp} is then defined as
\[
&f_\text{qk} = \mathrm{MLP}(\ell_\text{qk}, d_x, d_h, d_h) \nonumber\\
&q = f_\text{qk}(x_*), \quad K = \{ f_\text{qk}(x_i) \}_{i \in c}\nonumber \\
&V = \mathrm{SA}(d_h)(\{ \mathrm{MLP}(\ell_\text{v}, d_x+d_y, d_h)([x_i, y_i])\}_{i\in c})) \nonumber\\
&\phi = \mathrm{MHA}(d_h)(q, K, V), \quad f_\text{denc}(X_c, Y_c, x_*) = \phi.
\]
A latent path of \gls{anp} is
\[
& H = \mathrm{SA}(d_h)(\{ \mathrm{ReLU} \circ \mathrm{MLP}(\ell_\text{pre}, d_x+d_y, d_h, d_h)([x_i,y_i]) \}_{i\in c}) \nonumber\\
&(\eta, \rho') = \mathrm{MLP}(\ell_\text{post}, d_h, 2d_z) \bigg(\frac{1}{|c|}  \sum_{i\in c} h_{i}\bigg) \nonumber\\
&\rho = 0.1 + 0.9\cdot \mathrm{sigmoid}(\rho'), \quad (\eta, \rho) = f_\text{lenc}(X_c, Y_c).
\]
For \gls{canp} and \gls{banp}, we use the same architecture having two paths as follows:
\[
&f_\text{qk} = \mathrm{MLP}(\ell_\text{qk}, d_x, d_h, d_h) \nonumber\\
&q = f_\text{qk}(x_*), \quad K = \{ f_\text{qk}(x_i) \}_{i \in c}\nonumber \\
&V = \mathrm{SA}(d_h)(\{ \mathrm{MLP}(\ell_\text{v}, d_x+d_y, d_h)([x_i, y_i])\}_{i\in c}) \nonumber\\
&\phi_1 = \mathrm{MHA}(d_h)(q, K, V) \nonumber\\
&H = \mathrm{SA}(d_h)(\{ \mathrm{ReLU} \circ \mathrm{MLP}(\ell_\text{pre}, d_x+d_y, d_h, d_h)([x_i,y_i]) \}_{i\in c})\nonumber \\
&\phi_2 = \mathrm{MLP}(\ell_\text{post}, d_h, d_h) \bigg(\frac{1}{|c|}  \sum_{i\in c} h_{i}\bigg)\nonumber\\
&\phi = [\phi_1, \phi_2], \quad f_\text{enc}(X_c, Y_c, x_*) = \phi.
\]
\paragraph{Decoder}
Decoders are the same is in \cref{sup:subsec:np_architectures}.

\section{Experimental Details}
\label{sup:sec:experimental_details}
\subsection{1D Regression}
\label{sup:subsec:1d_regression}
\paragraph{Architectures} For models without attention (\gls{cnp}, \gls{np}, \gls{bnp}), we set $\ell_\text{pre}=4, \ell_\text{post}=2, \ell_\text{dec}=3, d_h=128$. For $\gls{np}$ we set $d_z=128$.
For models with attention (\gls{canp}, \gls{anp}, \gls{banp}), we set $\ell_\text{v}=2, \ell_\text{qk}=2, \ell_\text{pre}=2, \ell_\text{post}=2, \ell_\text{dec}=3, d_h=128, n_\text{head}=8$ and $d_z=128$ for \gls{anp}.

\paragraph{Data generation} We trained all the models using data generated from \glspl{gp} with RBF kernel. For each task $(X, Y, c)$, we first generated $x \iidsim \unifdist(-2, 2)$ and generated $Y$
from using RBF Kernel $k(x, x') = s^2\cdot \exp(-\Vert x-x' \Vert^2/2\ell^2)$ with $s \sim \unifdist(0.1, 1.0)$ and $\ell \sim \unifdist(0.1, 0.6)$, and output additive noise $\calN(0, 10^{-2})$. The size of the task and the size of the context $c$ was drawn as $|c| \sim \unifdist(3, 47)$ and $n-|c| \sim \unifdist(3, 50-|c|)$. For model-data mismatch scenario, we generated data from \gls{gp} with Matern52 kernels, periodic kernels, and \gls{gp} with RBF kernel plus Student's $t$ noise. For Matern52 kernel $k(x, x') = s^2(1+\sqrt{5}d/\ell + 5d^2/(3\ell^2)) \exp(-\sqrt{5}d/\ell),\quad  (d=\Vert x-x'\Vert)$, we sampled $s \sim \unifdist(0.1, 1.0)$ and $\ell \sim \unifdist(0.1, 0.6)$. For periodic kernel $k(x,x')= s^2\exp(-2\sin^2(\pi \Vert x-x' \Vert^2/p)/\ell^2)$, we sampled $s \sim \unifdist(0.1, 1.0)$ and $\ell \sim \unifdist(0.1, 0.6)$ and $p \sim \unifdist(0.1, 0.5)$. For Student-$t$ noise, we added $\varepsilon \sim \gamma\cdot \calT(2.1)$ to the curves generated from \gls{gp} with RBF kernel, where $\calT(2.1)$ is a Student's $t$ distribution with degree of freedom $2.1$ and $\gamma \sim \unifdist(0, 0.15)$.

\paragraph{Training and testing}
We trained all the model for 100,000 steps with each step computes updates with a batch containing 100 tasks. We used Adam optimizer~\citep{kingma2015adam} with initial learning rate $5\cdot 10^{-4}$ and decayed the learning rate using cosine annealing scheme. \gls{np} and \gls{anp} were trained using $k=4$ samples for $z$ (as in \cite{burda2015importance}), and tested with $k=50$ samples. \gls{bnp} and \gls{banp} were trained with $k=4$ bootstrap contexts and tested with $k=50$ samples. The size of the task and the size of the context $c$ was drawn as $|c| \sim \unifdist(3, 200)$ and $n-|c| \sim \unifdist(3, 200-|c|)$.
Testings were done for 3,000 batches with each batch containing 16 tasks (48,000 tasks in total).

\subsection{Bayesian Optimization}
\label{sup:subsec:bo_more}

\paragraph{Architectures / Training and testing}
For these experiments, we followed the settings described in \cref{subsec:1d_regression}.

\paragraph{Prior function generation}
We sampled 100 GP prior functions from zero mean and unit variance.
After realizing them, the prior functions are used to optimize via Bayesian optimization.
We normalized these functions in order to fairly compare simple regrets and cumulative regrets 
across distinct sampled functions (Basically, since they are sampled from same distributions, 
the scales of them are quite similar, but we used more precise evaluations).

\paragraph{Bayesian optimization setting}
As presented in the Bayesian optimization results, all the methods are started from same initializations.
We employed Gaussian process regression~\citep{RasmussenCE2006book} with squared exponential kernels as a surrogate model, 
and expected improvement~\citep{JonesDR1998jgo} as an acquisition function, 
which is optimized by the multi-started local optimization method, L-BFGS-B 
with 100 initial points.
All the experiments are implemented with \citep{KimJ2017bayeso}.

\subsection{Image Completion}
\label{sup:subsec:image_completion}
\paragraph{EMNIST architectures} For models without attention (\gls{cnp}, \gls{np}, \gls{bnp}), we set $\ell_\text{pre}=5, \ell_\text{post}=3, \ell_\text{dec}=4, d_h=128$. For $\gls{np}$ we set $d_z=128$.
For models with attention (\gls{canp}, \gls{anp}, \gls{banp}), we set $\ell_\text{v}=3, \ell_\text{qk}=3, \ell_\text{pre}=3, \ell_\text{post}=3, \ell_\text{dec}=4, d_h=128, n_\text{head}=8$ and $d_z=128$ for \gls{anp}.

\paragraph{CelebA32 architectures} For models without attention (\gls{cnp}, \gls{np}, \gls{bnp}), we set $\ell_\text{pre}=6, \ell_\text{post}=3, \ell_\text{dec}=5, d_h=128$. For $\gls{np}$ we set $d_z=128$.
For models with attention (\gls{canp}, \gls{anp}, \gls{banp}), we set $\ell_\text{v}=4, \ell_\text{qk}=3, \ell_\text{pre}=4, \ell_\text{post}=3, \ell_\text{dec}=5, d_h=128, n_\text{head}=8$ and $d_z=128$ for \gls{anp}.

\paragraph{Data generation}
Each task $(X, Y, c)$ was sampled from an image. Following \cite{garnelo2018neural,kim2018attentive}, we sampled 2D coordinates from an image and rescaled the values into $[-1, 1]$ to comprise $X$, and rescaled the corresponding pixel values into $[-0.5, 0,5]$ to comprise $Y$. The size of the task and the size of the context $c$ was drawn as $|c| \sim \unifdist(3, 200)$ and $n-|c| \sim \unifdist(3, 200-|c|)$.
For EMNIST we used the first 10 classes during training, and tested on remaining 37 classes as a model-data mismatch scenario. 

\paragraph{Training and testing}
Same as \cref{subsec:1d_regression}, except that all the models were trained for 200 epochs through the datasets. The models were tested on entire test set where each sample in a test set comprises a task.
For a model-data mismatch scenario with Student's $t$ noise, we added $\varepsilon \sim \gamma \cdot \calT(2.1)$ with $\gamma \sim \unifdist(0, 0.09)$ to $Y$. 

\subsection{Lotka-Volterra}
\paragraph{Architectures} For models without attention (\gls{cnp}, \gls{np}, \gls{bnp}), we set $\ell_\text{pre}=4, \ell_\text{post}=2, \ell_\text{dec}=3, d_h=128$. For $\gls{np}$ we set $d_z=128$.
For models with attention (\gls{canp}, \gls{anp}, \gls{banp}), we set $\ell_\text{pre}=2, \ell_\text{post}=2, \ell_\text{dec}=3, d_h=128, n_\text{head}=8$ and $d_z=128$ for \gls{anp}.

\paragraph{Dataset generation}
We followed the setting in \cite{gordon2020convolutional}, please refer to the description in the paper. A task $(X, Y, c)$ is then constructed by uniformly subsampling $X$ and corresponding $Y$ from the generated series. The size of the task and the size of the context $c$ was drawn as $|c| \sim \unifdist(15, 85)$ and $n-|c| \sim \unifdist(15, 100-|c|)$. 
Due to the scaling issue, $X$ and $Y$ values were standardized using the statistics computed from the context:
\[
x_i' = \frac{x_i - \mathrm{mean}(X_c)}{\mathrm{std}(X_c)+10^{-5}}, \quad y_i' = \frac{y_i - \mathrm{mean}(Y_c)}{\mathrm{std}(Y_c)+10^{-5}}.
\]

\paragraph{Training and testing}
We trained for 100,000 steps with each step is computed with a batch containing 50 tasks. The other details are the same as in \cref{subsec:1d_regression}. 
Testing was done on 3,000 batches with each batch containing 16 tasks. For real-data testing as a model-data mismatch scenario, following~\cite{gordon2020convolutional},
we generated 1,000 batches with each batch containing 16 tasks from Hudson's Bay hare-lynx data. Each task contained $|c| \sim \unifdist(15, 76)$ and $n \sim \unifdist(15, 91-|c|)$ points subsampled from the data,
and standardized as above.

\section{On calibration and sharpness of the models}
\label{sup:sec:calib}
\begin{table}
    \centering
    \scriptsize
\setlength{\tabcolsep}{3pt}
\caption{Calibration error and sharpness of the models for 1D regression experiments. Means and standard deviations of 5 runs are reported.}
\vspace{1pt}
    \begin{tabular}{@{}c cc cc cc cc@{}}
    \toprule
    &  \multicolumn{2}{c}{RBF} & \multicolumn{2}{c}{Mat\'ern 5/2} & \multicolumn{2}{c}{Periodic} & \multicolumn{2}{c}{$t$-noise}\\
    \cmidrule[0.2pt]{2-9}  &  CE & Sharpness & CE & Sharpness & CE & Sharpness & CE & Sharpness \\
    \midrule
CNP & \msd{0.059}{0.003} & \msd{0.072}{0.001} & \msd{0.012}{0.001} & \msd{0.079}{0.001} & \msd{0.171}{0.004} & \msd{0.226}{0.004} & \msd{0.029}{0.002} & \msd{0.093}{0.001} \\
NP & \msd{0.016}{0.001} & \msd{0.06}{0.001} & \msd{0.037}{0.005} & \msd{0.067}{0.001} & \msd{0.306}{0.016} & \msd{0.224}{0.001} & \msd{0.138}{0.012} & \msd{0.082}{0.001} \\
\cmidrule[0.2pt]{1-9}
BNP & \msd{0.049}{0.002} & \msd{0.069}{0.000} & \msd{0.011}{0.001} & \msd{0.077}{0.000} & \msd{0.145}{0.002} & \msd{0.243}{0.008} & \msd{0.032}{0.001} & \msd{0.098}{0.001} \\
\cmidrule[0.5pt]{1-9}
CANP & \msd{0.276}{0.005} & \msd{0.057}{0.001} & \msd{0.127}{0.003} & \msd{0.066}{0.000} & \msd{0.251}{0.022} & \msd{0.157}{0.006} & \msd{0.038}{0.003} & \msd{0.086}{0.002} \\
ANP & \msd{0.144}{0.009} & \msd{0.048}{0.001} & \msd{0.051}{0.003} & \msd{0.055}{0.002} & \msd{0.402}{0.031} & \msd{0.165}{0.007} & \msd{0.154}{0.014} & \msd{0.074}{0.003} \\
\cmidrule[0.2pt]{1-9}
BANP & \msd{0.264}{0.001} & \msd{0.057}{0.000} & \msd{0.121}{0.001} & \msd{0.067}{0.000} & \msd{0.0.226}{0.002} & \msd{0.176}{0.003} & \msd{0.035}{0.001} & \msd{0.095}{0.001} \\
\bottomrule          
    \end{tabular}
    \label{tab:1d_regression_calib}
\end{table}

\begin{table}
    \centering
    \setlength{\tabcolsep}{3pt}
    \scriptsize
\caption{Calibration error and sharpness of the models for EMNIST experiments. Means and standard deviations of 5 runs are reported.}
    \begin{tabular}{@{}c cc cc cc@{}}
    \toprule
      &  \multicolumn{2}{c}{Seen classes (0-9)} & \multicolumn{2}{c}{Unseen classes (10-46)} & \multicolumn{2}{c}{$t$-noise}\\
    \cmidrule[0.2pt]{2-7} &  CE & sharpness & CE & Sharpness & CE & Sharpness \\
    \midrule
CNP & \msd{0.448}{0.007} & \msd{0.035}{0.001} & \msd{0.355}{0.007} & \msd{0.043}{0.001} & \msd{0.066}{0.008} & \msd{0.066}{0.0.055} \\
NP & \msd{0.423}{0.007} & \msd{0.042}{0.001} & \msd{0.337}{0.004} & \msd{0.050}{0.001} & \msd{0.046}{0.008} & \msd{0.069}{0.001} \\
\cmidrule[0.2pt]{1-7}
BNP & \msd{0.435}{0.007} & \msd{0.037}{0.001} & \msd{0.342}{0.006} & \msd{0.046}{0.001} & \msd{0.044}{0.014} & \msd{0.070}{0.003} \\
\cmidrule[0.5pt]{1-7}
CANP & \msd{0.533}{0.006} & \msd{0.029}{0.000} & \msd{0.463}{0.003} & \msd{0.032}{0.000} & \msd{0.327}{0.065} & \msd{0.085}{0.006} \\
ANP & \msd{0.489}{0.010} & \msd{0.034}{0.001} & \msd{0.442}{0.008} & \msd{0.036}{0.001} & \msd{0.197}{0.041} & \msd{0.085}{0.006} \\
\cmidrule[0.2pt]{1-7}
BANP & \msd{0.511}{0.011} & \msd{0.032}{0.001} & \msd{0.449}{0.006} & \msd{0.035}{0.001} & \msd{0.117}{0.023} & \msd{0.076}{0.006} \\
    \bottomrule          
    \end{tabular}
    \label{tab:emnist_calib}
\end{table}

\begin{table}
\centering
    \setlength{\tabcolsep}{3pt}
    \scriptsize
    \caption{Calibration error and sharpness of the models on CelebA32 experiments. Means and standard deviations of 5 runs are reported.}
    \begin{tabular}{@{}c cc cc@{}}
    \toprule
      &  \multicolumn{2}{c}{Without noise} & \multicolumn{2}{c}{$t$-noise} \\
    \cmidrule[0.2pt]{2-5} &  CE & Sharpness & CE & Sharpness \\
    \midrule
    CNP & \msd{0.019}{0.000} & \msd{0.056}{0.000} & \msd{0.003}{0.000} & \msd{0.080}{0.002} \\
    NP & \msd{0.017}{0.000} & \msd{0.065}{0.000} & \msd{0.062}{0.002} & \msd{0.009}{0.003} \\
    \cmidrule[0.2pt]{1-5}
    BNP & \msd{0.008}{0.000} & \msd{0.065}{0.009} & \msd{0.035}{0.006} & \msd{0.101}{0.002} \\
    \cmidrule[0.5pt]{1-5}
    CANP & \msd{0.069}{0.000} & \msd{0.054}{0.000} & \msd{0.007}{0.002} & \msd{0.110}{0.010} \\
    ANP & \msd{0.018}{0.000} & \msd{0.062}{0.000} & \msd{0.082}{0.002} & \msd{0.096}{0.001} \\
    \cmidrule[0.2pt]{1-5}
    BANP & \msd{0.018}{0.000} & \msd{0.065}{0.000} & \msd{0.075}{0.012} & \msd{0.100}{0.002} \\
    \bottomrule          
    \end{tabular}
    \label{tab:celeba_calib}
\end{table}

\begin{table}
\centering
    \setlength{\tabcolsep}{3pt}
    \scriptsize
    \caption{Calibration error and sharpness of the models on Predator-prey experiments. Means and standard deviations of 5 runs are reported.}
    \begin{tabular}{@{}c cc cc@{}}
    \toprule
      &  \multicolumn{2}{c}{Simulated} & \multicolumn{2}{c}{Real} \\
    \cmidrule[0.2pt]{2-5} &  CE & Sharpness & CE & Sharpness \\
    \midrule
    CNP & \msd{0.001}{0.000} & \msd{0.578}{0.013} & \msd{0.072}{0.008} & \msd{1.866}{0.058} \\
    NP & \msd{0.002}{0.003} & \msd{0.567}{0.009} & \msd{0.087}{0.000} & \msd{1.877}{0.069} \\
    \cmidrule[0.2pt]{1-5}
    BNP & \msd{0.003}{0.000} & \msd{0.542}{0.016} & \msd{0.076}{0.011} & \msd{1.975}{0.004} \\
    \cmidrule[0.5pt]{1-5}
    CANP & \msd{0.146}{0.003} & \msd{0.076}{0.001} & \msd{0.565}{0.034} & \msd{0.350}{0.034} \\
    ANP & \msd{0.104}{0.004} & \msd{0.064}{0.001} & \msd{0.814}{0.036} & \msd{0.248}{0.015} \\
    \cmidrule[0.2pt]{1-5}
    BANP & \msd{0.140}{0.003} & \msd{0.074}{0.001} & \msd{0.539}{0.039} & \msd{0.352}{0.019} \\
    \bottomrule          
    \end{tabular}
    \label{tab:predator_prey_calib}
\end{table}
We further analyze the learned models using the framework introduced in \cite{kuleshov2018accurate}. Let $\calT = (X, Y, c)$ be a task. We see how the predictions for the targets $\{(x_i, y_i)\}_{i\notin c}$ is calibrated, and how large the variances are. Let $F_{x_i}(y_i)$ be the CDF of the prediction $p(y_i|x_i, X_c, Y_c)$. We say a model is perfectly calibrated~\citep{kuleshov2018accurate} if for any $p\in [0,1]$,
\[
\frac{1}{n-|c|}\sum_{i\notin c} \1{y_i\leq F^{-1}_{x_i}(p) \leq  p} \to p \text{ as } n \to \infty,
\]
The calibration error (CE) is then defined as
\[
0\leq p_1 \leq \dots p_m \leq 1, \quad 
\hat p_\ell = \frac{1}{n-|c|}\sum_{i\notin c}^n \1{y_i  \leq  F^{-1}_{x_i}(p_\ell)}, \quad \mathrm{CE}(\calT) = \sum_{\ell=1}^m (p_\ell - \hat p_\ell)^2.
\]
In our case, we set $p(y_i|x_i, X_c, Y_c) = \calN(y_i|\mu_i, \sigma_i^2)$, so
\[
F^{-1}_{x_i}(p_\ell) = \mu_i + \sigma_i\sqrt{2}\mathrm{erf}^{-1}(2 p_\ell-1).
\]
For the models using the ensemble of multiple predictions (\gls{np}, \gls{anp}, \gls{bnp}, \gls{banp}), we report the ensembled calibration error.
\[
(F_{x_i}\ssc{j})^{-1}(p_\ell) &= \mu_i\ssc{j} + \sigma_i\ssc{j}\sqrt{2}\mathrm{erf}^{-1}(2 p_\ell-1),\\
\hat p_\ell\ssc{j} &= \frac{1}{n-|c|}\sum_{i\notin c} \1{y_i \leq (F_{x_i}\ssc{j})^{-1}(p_\ell)},\\
\mathrm{CE}(\calT) &= \frac{1}{k}\sum_{j=1}^k \sum_{i=1}^n (p_\ell - \hat p_\ell\ssc{j})^2.
\]
We also measure the sharpness~\citep{kuleshov2018accurate} which essentially is a average prediction variance.
\[
\text{Sharpness}(\calT) = \frac{1}{n-|c|} \sum_{i\notin c} \sigma_i^2.
\]

We evaluated the CE and sharpness of  \gls{cnp},\gls{np},\gls{bnp},\gls{canp},\gls{anp}, and \gls{banp} trained in the experiments. The results are summarized in \cref{tab:1d_regression_calib,tab:emnist_calib,tab:celeba_calib,tab:predator_prey_calib}. In general, ours (\gls{bnp} and \gls{banp}) were better calibrated for model-data mismatch settings, but
worse calibrated than \gls{np} and \gls{anp} for normal test settings or model-data mismatch settings not very different from the normal test setting (e.g., Mat\'ern 5/2 kernels in 1D regression experiments and unseen classes for EMNIST). The reason is that, as we stated in the main text, \gls{bnp} and \gls{banp} tends to produce conservative credible intervals, so become under-confident in normal-test settings and less over-confident in mismatch settings. This corresponds to the observation and theory in \cite{huggins2019using}, where BayesBag is proven to yield credible intervals that are twice larger than the credible intervals produced by normal Bayesian models when the model is correctly specified. The sharpness values also support this claim, where \gls{bnp} and \gls{banp} generally shows higher values than others especially for the mismatch settings. Interestingly, \gls{cnp} and \gls{canp} exhibit similar trends to ours (larger sharpness values than \gls{np} or \gls{anp}), presumably because they output only one predictor without any functional uncertainty and thus are encouraged to be conservative than \gls{np} or \gls{anp} to cover wider range predictions. Still, \gls{bnp} and \gls{banp} produced the largest sharpness values in overall. 
Although this trend we discussed is apparent in 1D regression and predator-prey experiments, we fail to find any of such trend for image completion experiments. We conjecture that this is because for image completion experiments we are restricting the range of function values $y$ to lie in $[-0.5, 0.5]$. This suggests that at least for image completion experiments, the robustness of ours (which is clearly demonstrated both in terms of likelihood values and qualitative samples) comes from a different reason.

\section{Additional results}
\label{sup:sec:additional_figures}

\subsection{1D Regression}
\label{sup:subsec:1d_regression_additional}
\paragraph{Ablation study}
We present an ablation study to empirically validate our design choices for \gls{bnp} and \gls{banp} on 1D regression experiment. 
We compared our full model to the followings: 1) na\"ive residual bootstrap applied to \gls{cnp} and \gls{canp} as described in \cref{subsec:naive}, 2) \gls{bnp} and \gls{banp} without context resampling via paired bootstrap, and 3) \gls{bnp} and \gls{banp} without adaptation path so decoder just taking the representations of bootstrapped contexts, and 4) \gls{bnp} and \gls{banp} trained without additional $p_\text{base}$ loss in \cref{eq:bnp_training}. \cref{sup:tab:1d_regression_ablation} summarizes the results. Except for the case without adaptation layer which showed slightly better performance on mismatch settings, every ablation cases showed poor performance. Naive bootstrap didn't work well for both normal test and mismatch settings, the models without paired bootstrap worked poorly on mismatch settings, and the models without adaptation layer didn't perform well on normal test settings.

\begin{table}
    \centering
    \scriptsize
\setlength{\tabcolsep}{3pt}
\caption{Ablation study for 1D regression.}
\vspace{1pt}
    \begin{tabular}{@{}c cc cc cc cc@{}}
    \toprule
    &  \multicolumn{2}{c}{RBF} & \multicolumn{2}{c}{Mat\'ern 5/2} & \multicolumn{2}{c}{Periodic} & \multicolumn{2}{c}{$t$-noise}\\
    \cmidrule[0.2pt]{2-9}  &  context & target & context & target & context & target & context & target \\
    \midrule
BNP & \msd{ 1.012}{0.006} & \msd{ 0.523}{0.004} & \msd{ 0.891}{0.007} & \msd{ 0.316}{0.004} & \msd{-0.111}{0.002} & \msd{-1.089}{0.009} & \msd{ 0.554}{0.006} & \msd{-0.644}{0.010} \\
na\"ive bootstrap & \msd{0.774}{0.015} & \msd{0.304}{0.011} & \msd{0.642}{0.017} & \msd{0.088}{0.008} & \msd{-0.261}{0.004} & \msd{-1.368}{0.019} & \msd{0.329}{0.012} & \msd{-1.203}{0.030} \\
- paired bootstrap & \msd{0.990}{0.005} & \msd{0.491}{0.004} & \msd{0.865}{0.006} & \msd{0.269}{0.004} & \msd{-0.144}{0.004} & \msd{-1.342}{0.014} & \msd{0.455}{0.037} & \msd{-1.130}{0.025} \\
- adaptation layer &  \msd{0.900}{0.010} & \msd{0.455}{0.007} & \msd{0.803}{0.011} & \msd{0.294}{0.006} & \msd{0.009}{0.008} & \msd{-0.845}{0.006} & \msd{0.579}{0.010} & \msd{-0.337}{0.015} \\
- $p_\text{base}$ loss & \msd{0.992}{0.010} & \msd{0.496}{0.007} & \msd{0.868}{0.011} & \msd{0.273}{0.007} & \msd{-0.135}{0.010} & \msd{-1.315}{0.016} & \msd{0.468}{0.014} & \msd{-1.068}{0.032} \\
\cmidrule[0.5pt]{1-9}
BANP & \msd{1.379}{0.000} & \msd{ 0.849}{0.001} & \msd{ 1.376}{0.000} & \msd{0.671}{0.001} & \msd{0.688}{0.044} & \msd{ -3.429}{0.084} & \msd{ 1.137}{0.007} & \msd{ -1.750}{0.031} \\
na\"ive bootstrap & \msd{1.365}{0.008} & \msd{0.822}{0.014} & \msd{1.356}{0.011} & \msd{0.632}{0.014} & \msd{0.502}{0.068} & \msd{-3.729}{0.151} & \msd{1.041}{0.023} & \msd{-1.782}{0.020} \\
- paired bootstrap & \msd{1.379}{0.000} & \msd{0.841}{0.002} & \msd{1.377}{0.000} & \msd{0.655}{0.002} & \msd{0.830}{0.031} & \msd{-4.510}{0.138} & \msd{1.141}{0.014} & \msd{-2.179}{0.019} \\
- adaptation layer & \msd{1.370}{0.000} & \msd{0.830}{0.001} & \msd{1.361}{0.000} & \msd{0.639}{0.002} & \msd{0.523}{0.030} & \msd{-3.598}{0.099} & \msd{1.046}{0.003} & \msd{-1.765}{0.014} \\
- $p_\text{base}$ loss & \msd{1.378}{0.000} & \msd{0.836}{0.002} & \msd{1.375}{0.000} & \msd{0.661}{0.001} & \msd{0.647}{0.041} & \msd{-3.801}{0.294} & \msd{1.132}{0.004} & \msd{-1.697}{0.050} \\
\bottomrule          
    \end{tabular}
    \label{sup:tab:1d_regression_ablation}
\end{table}

\begin{figure}
\centering
\includegraphics[width=0.4\linewidth]{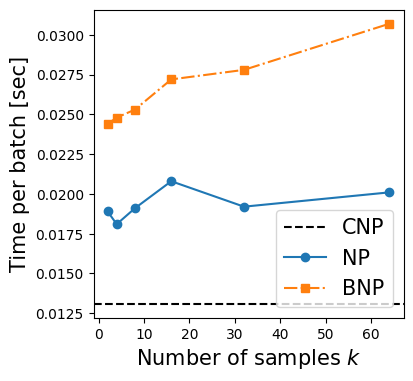}
\includegraphics[width=0.38\linewidth]{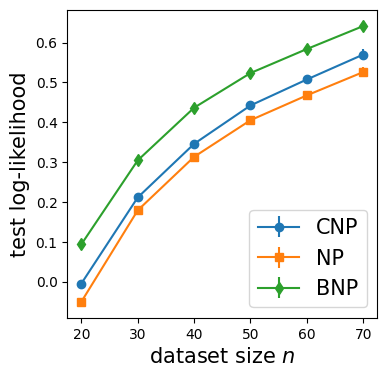}
\caption{(Left) processing time per batch. (Right) log-likelihood with different dataset sizes $n$.}
\label{sup:fig:1d_regression_graphs}
\end{figure}

\paragraph{Training time}
We measured averaging training time per batch for \gls{cnp}, \gls{np}, and \gls{bnp} on 1D regression task~(\cref{sup:fig:1d_regression_graphs}, left). \gls{bnp} forwards the data to the model twice, but the actual computation time for \gls{bnp} is less than the twice of the computation time of \gls{np}, because the first pass to compute residuals uses only the context set $(X_c, Y_c)$ which is a subset of the entire batch $(X, Y)$. Thanks to the parallelization by packing every dataset into a tensor, the computation times for all models does not scale linearly with the number of samples $k$.

\paragraph{Performance for various dataset size $n$}
We measured the average target log-likelihood with varying dataset size $n$ on 1D regression task~(\cref{sup:fig:1d_regression_graphs}, right). \gls{bnp} uniformly performed better than \gls{cnp} and \gls{np} by a significant margin.

\begin{figure}
    \centering
    \includegraphics[width=0.4\linewidth]{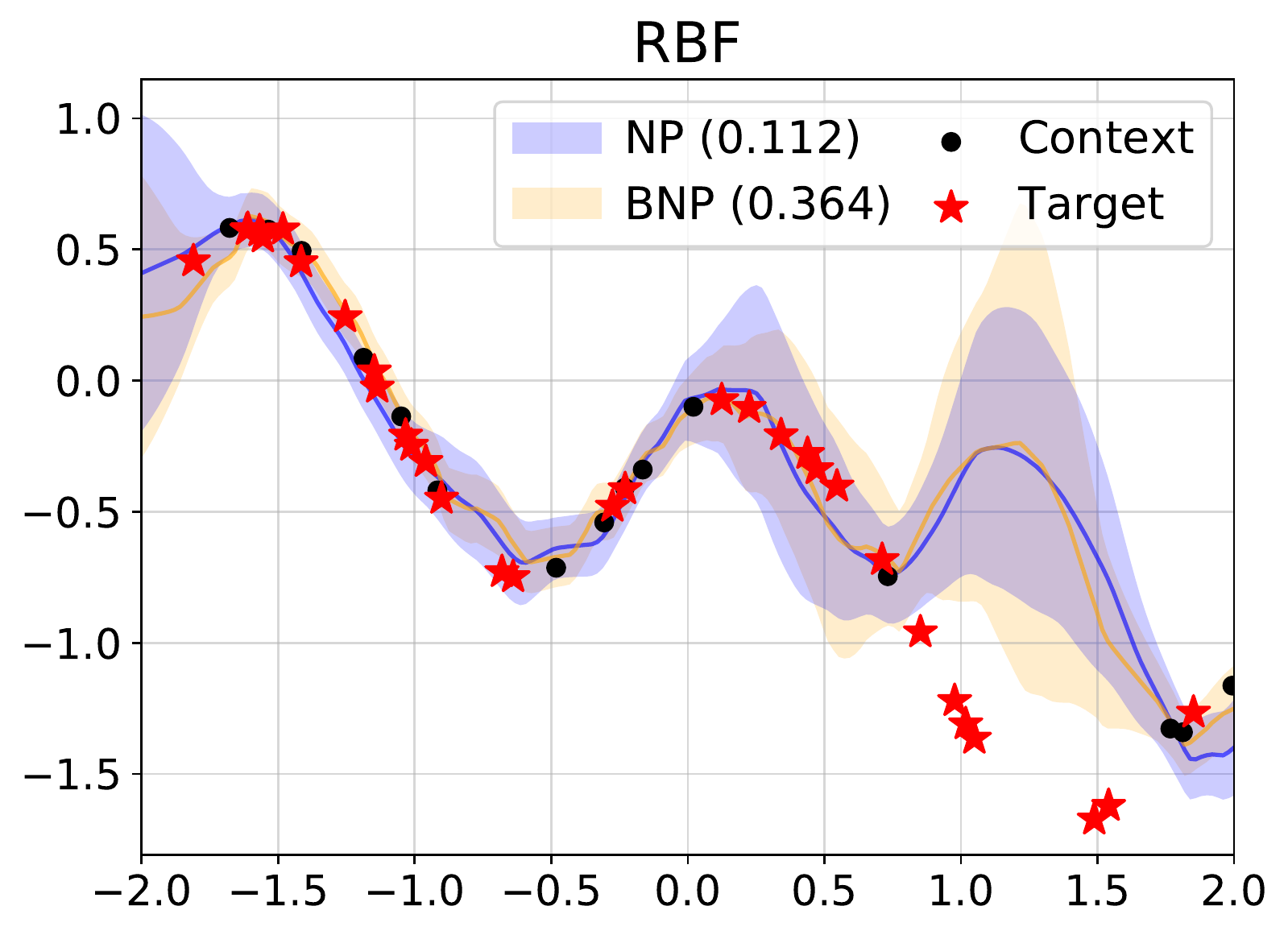}
    \includegraphics[width=0.4\linewidth]{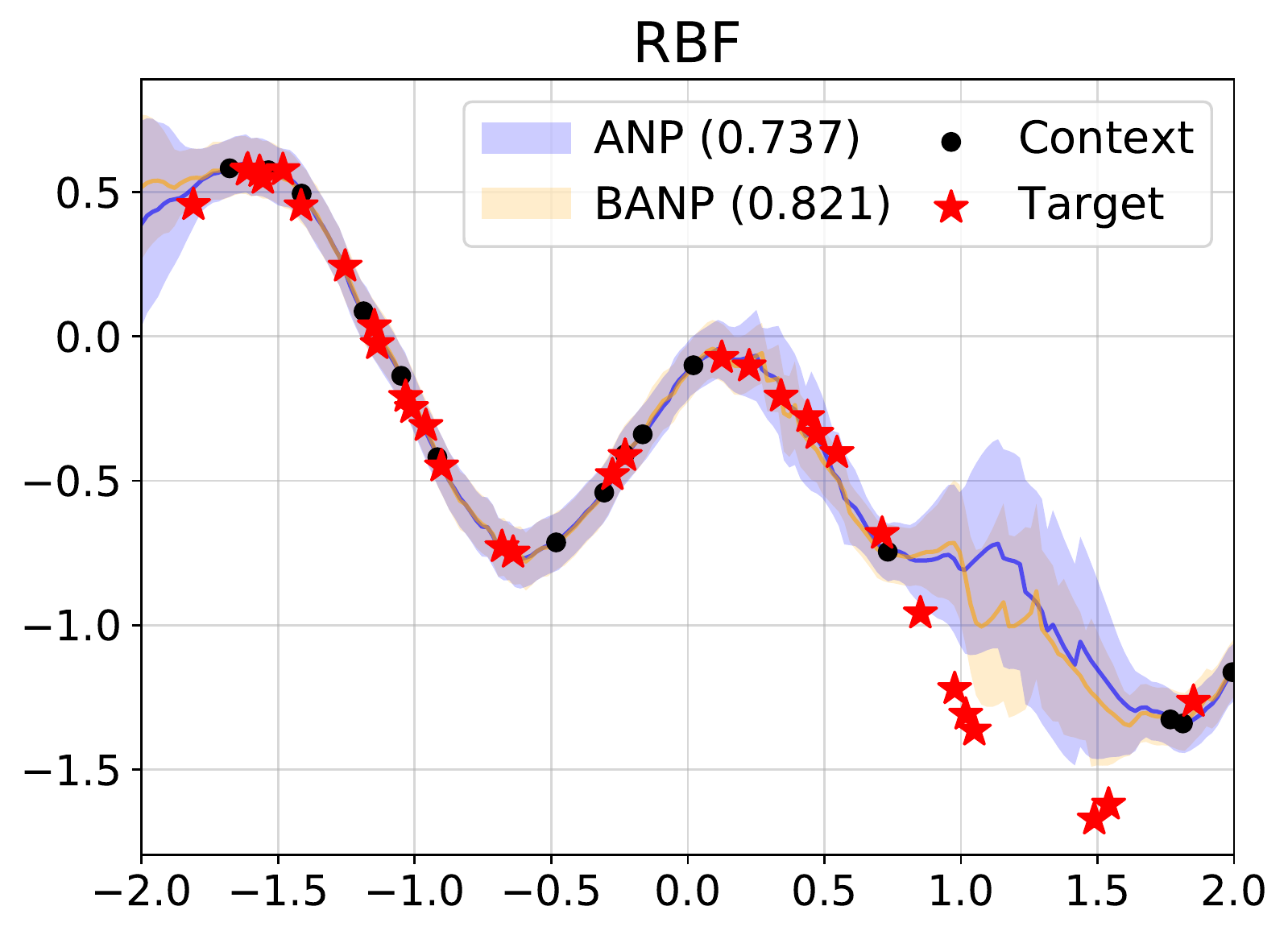}
    \includegraphics[width=0.4\linewidth]{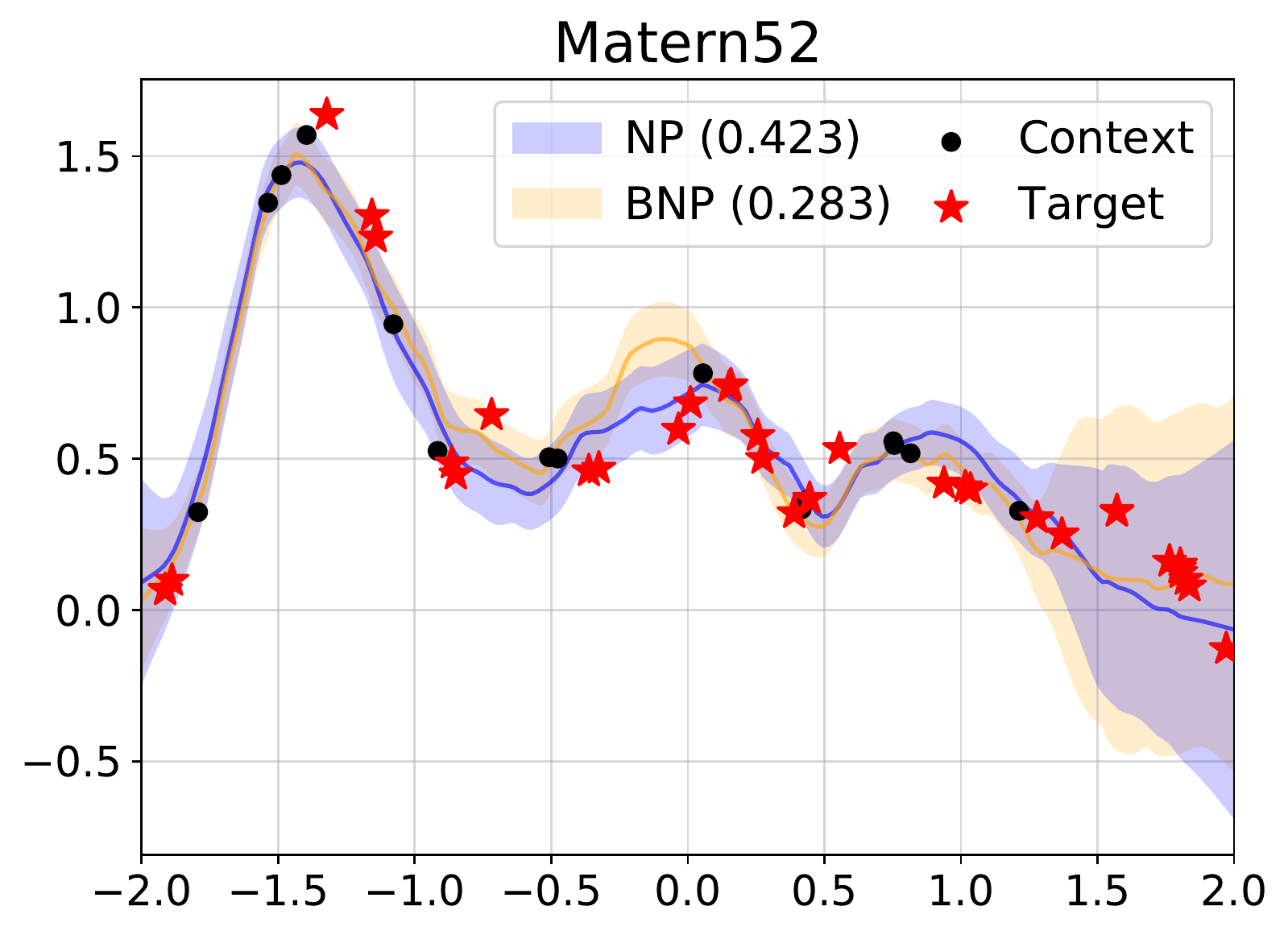}
    \includegraphics[width=0.4\linewidth]{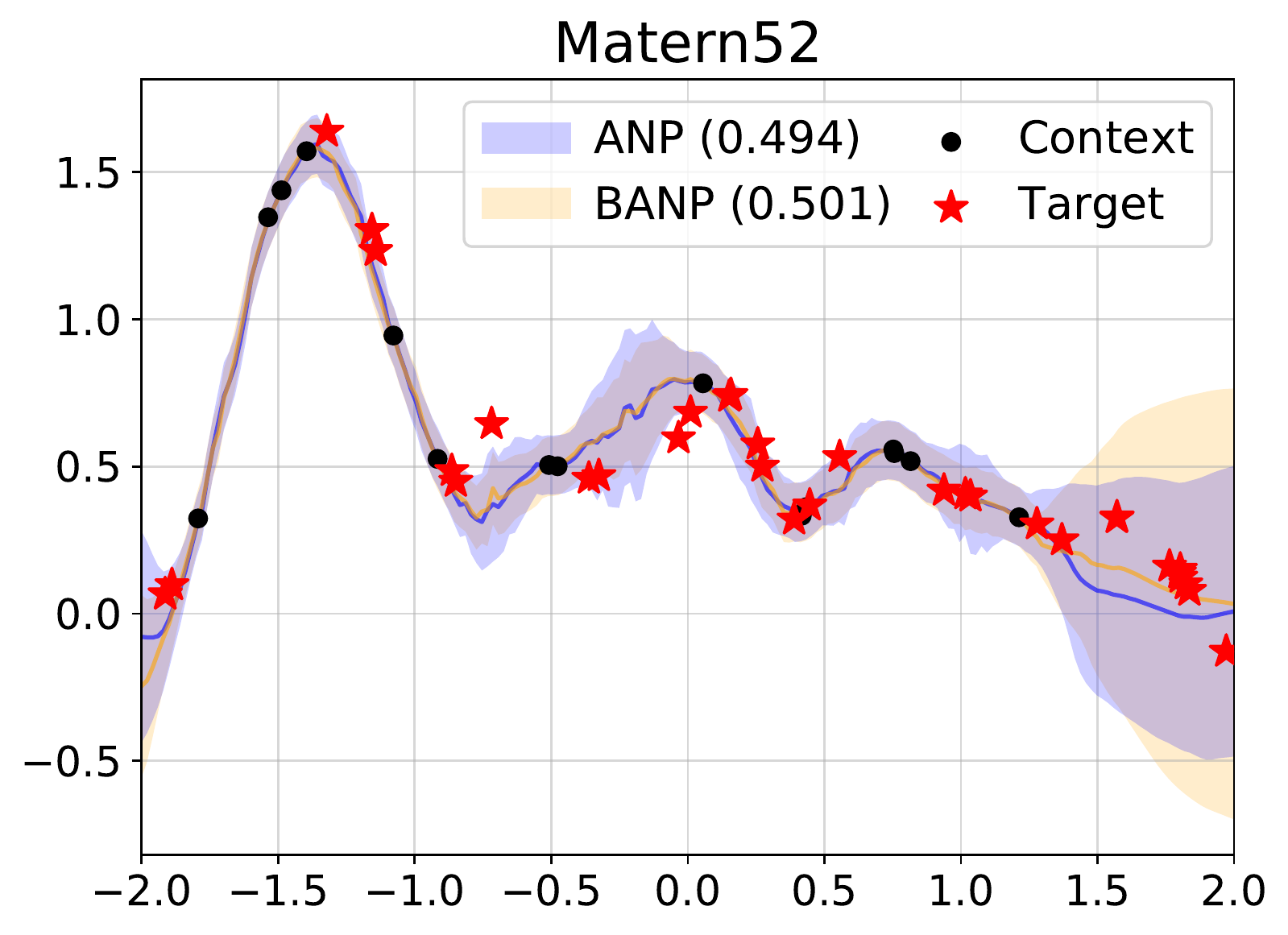}
    \includegraphics[width=0.4\linewidth]{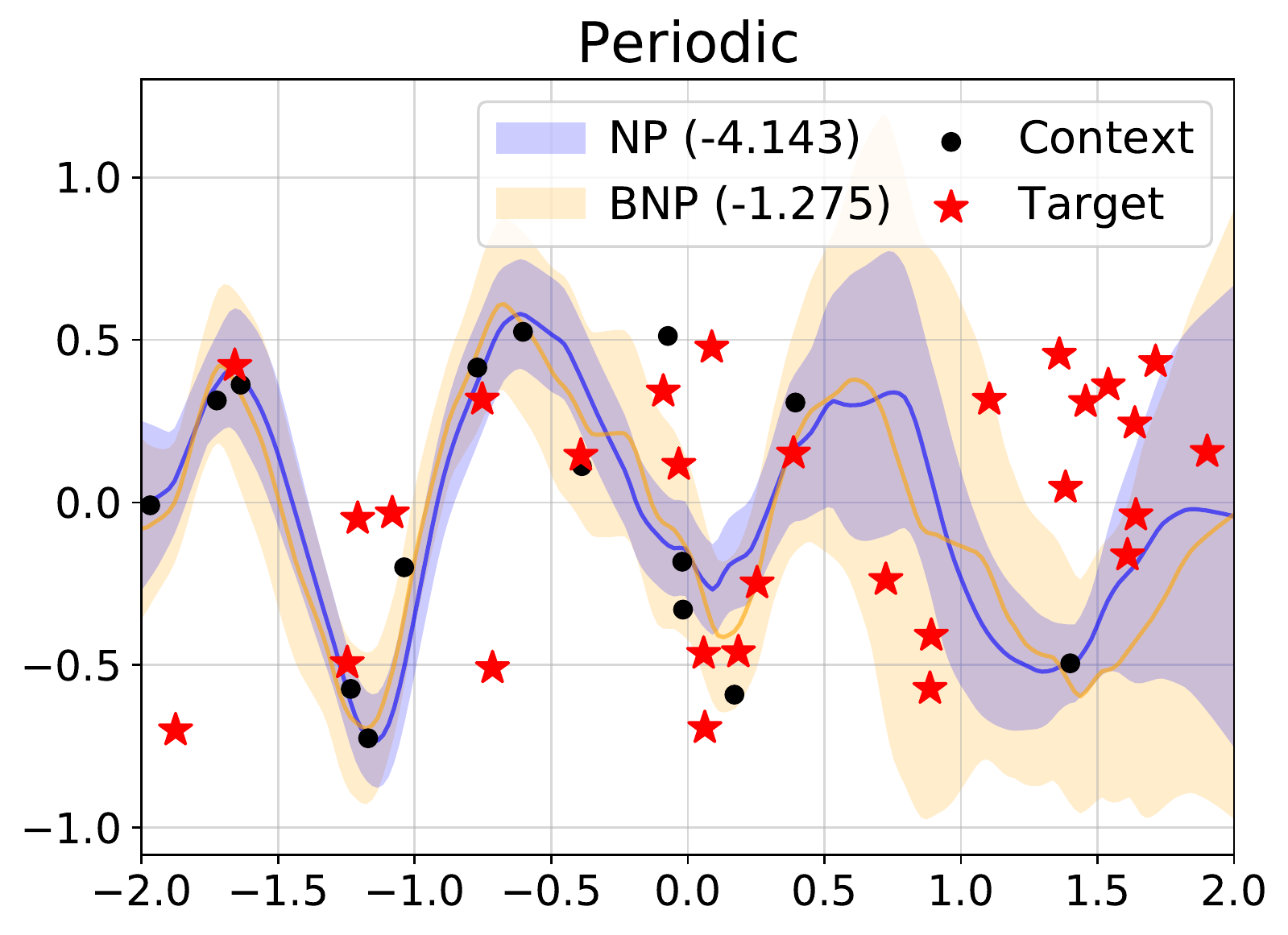}
    \includegraphics[width=0.4\linewidth]{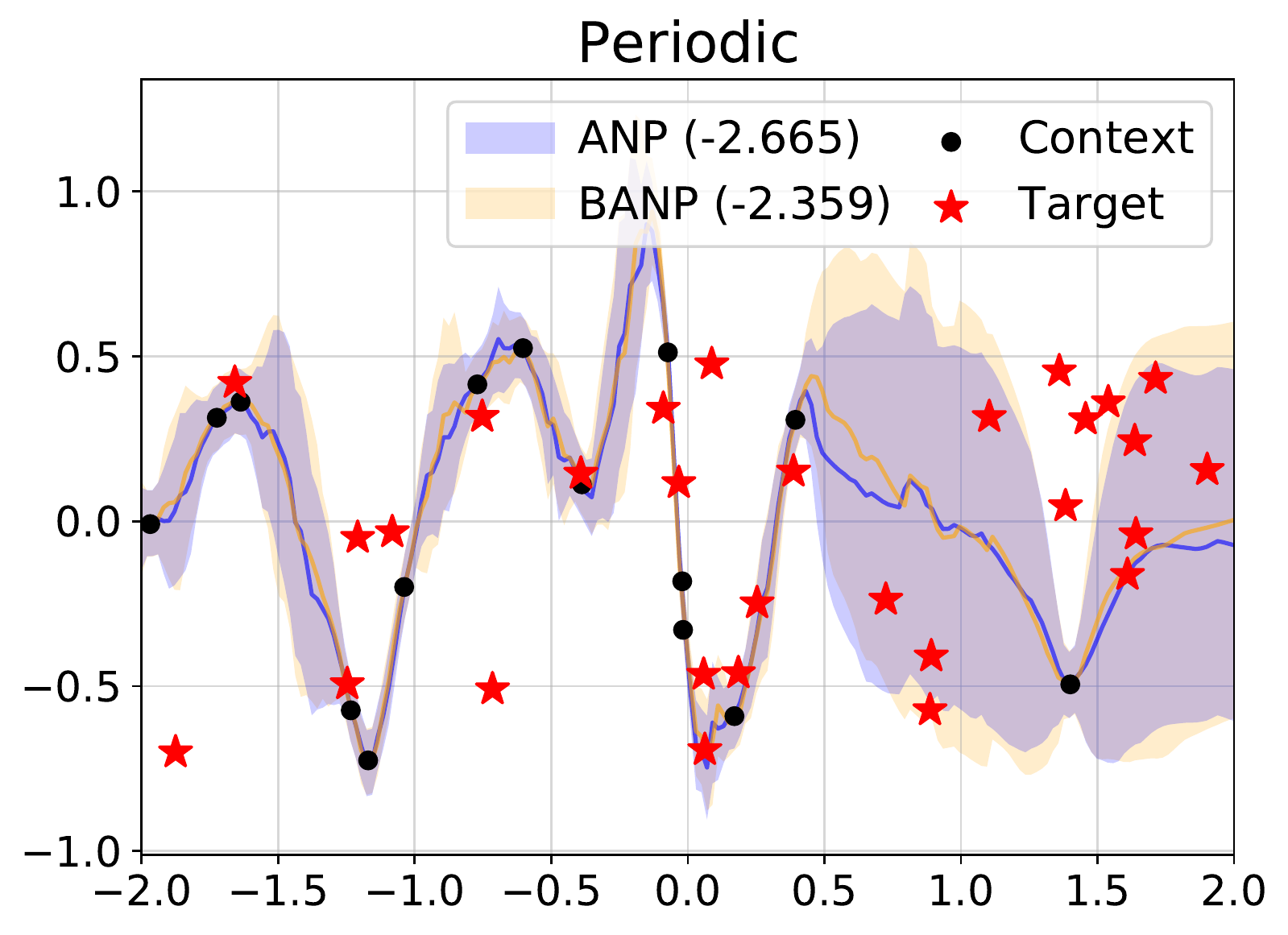}
    \includegraphics[width=0.4\linewidth]{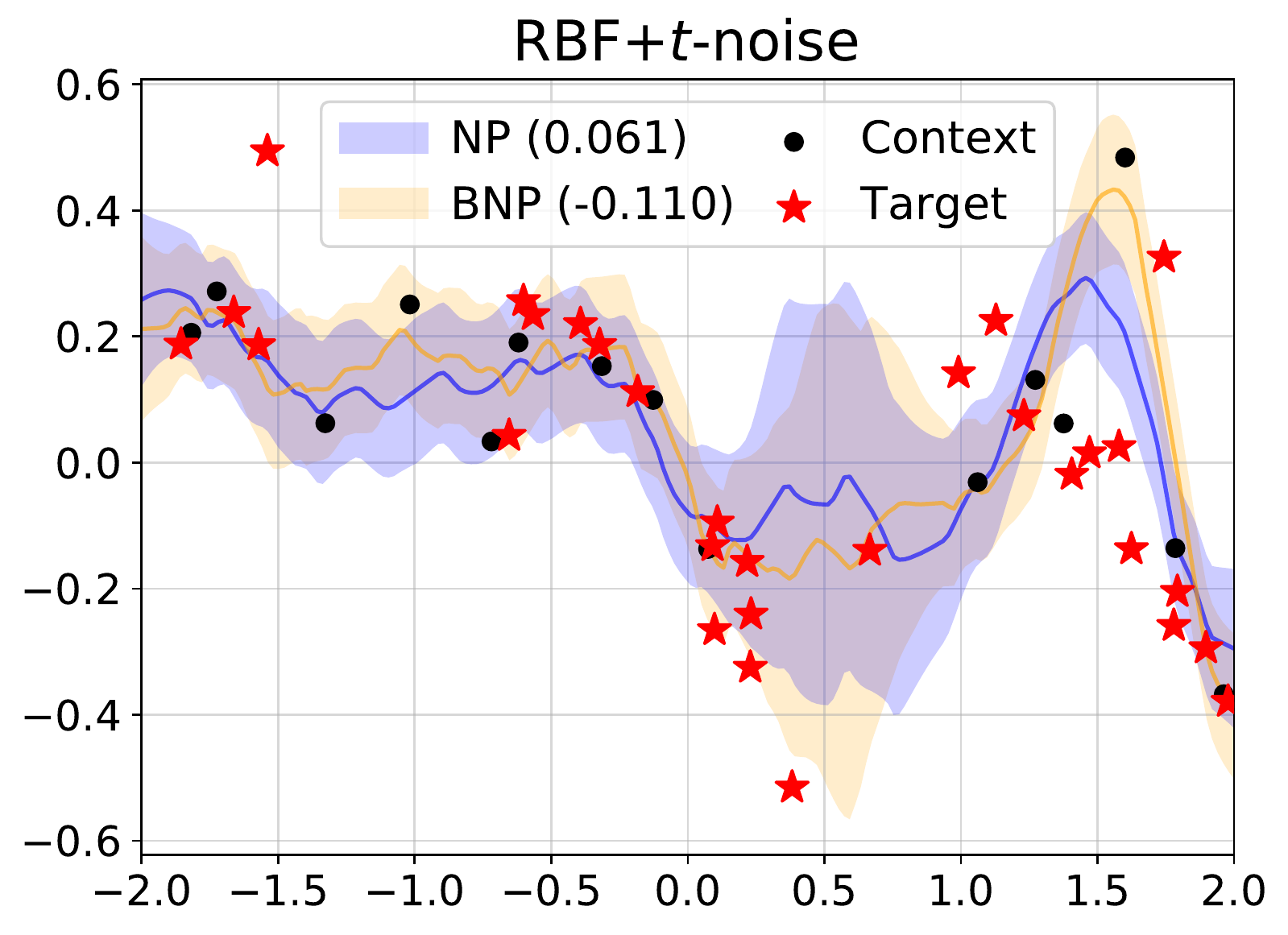}
    \includegraphics[width=0.4\linewidth]{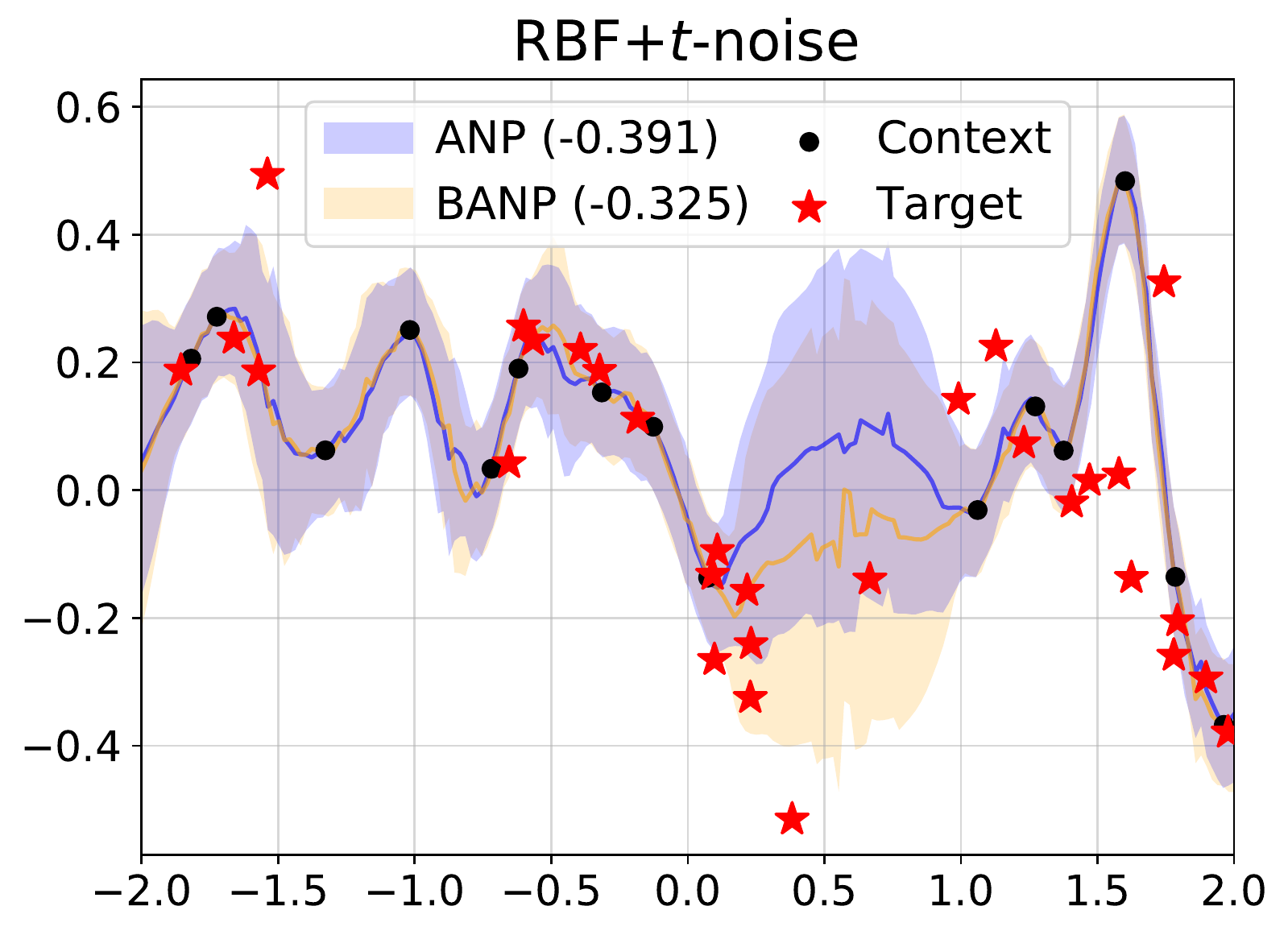}
    \caption{More visualizations for 1D regression experiment.}
    \label{sup:fig:1d_regression_more_vis}
\end{figure}

\paragraph{Additional figures} 
Here we present additional samples in \cref{sup:fig:1d_regression_more_vis}.

\subsection{Bayesian optimization}
\label{subsec:bo_additional}

\begin{figure}[t]
    \centering
    \includegraphics[width=0.21\linewidth]{figures/bo/bo_rbf_instantaneous_wo_attention.pdf}
    \includegraphics[width=0.21\linewidth]{figures/bo/bo_rbf_instantaneous_w_attention.pdf}
    \includegraphics[width=0.21\linewidth]{figures/bo/bo_rbf_cumulative_wo_attention.pdf}
    \includegraphics[width=0.21\linewidth]{figures/bo/bo_rbf_cumulative_w_attention.pdf}
    \includegraphics[width=0.21\linewidth]{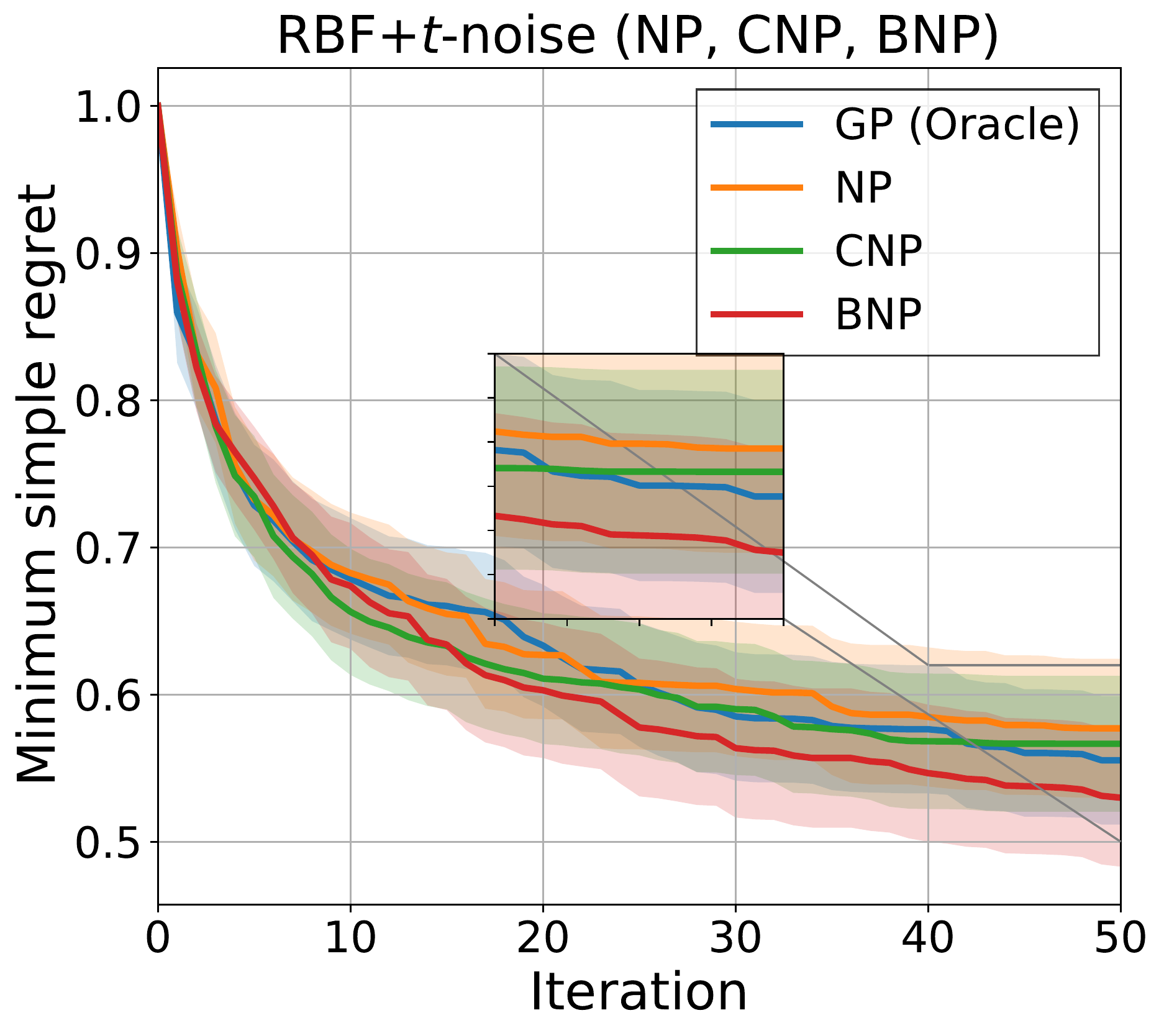}
    \includegraphics[width=0.21\linewidth]{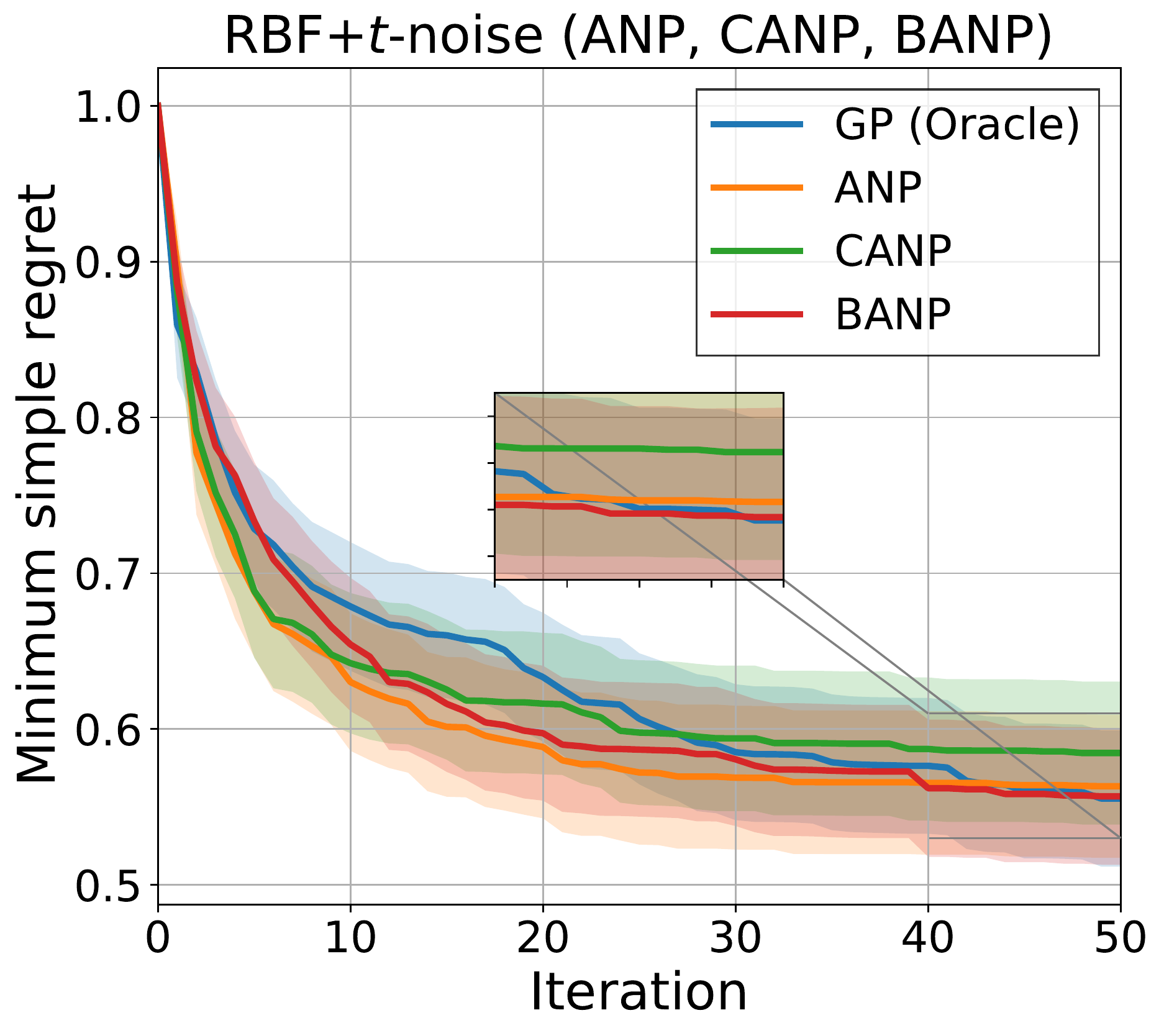}
    \includegraphics[width=0.21\linewidth]{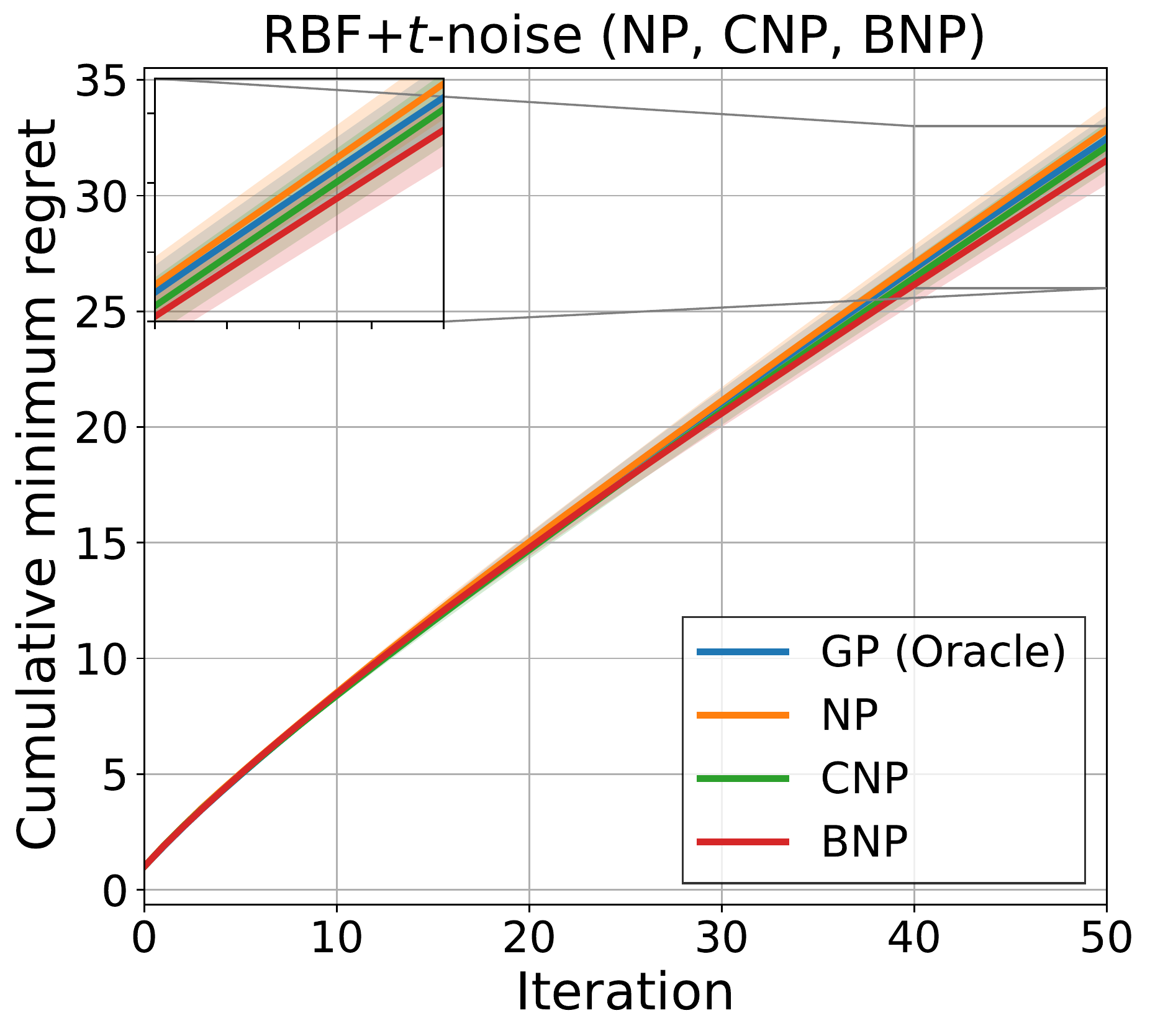}
    \includegraphics[width=0.21\linewidth]{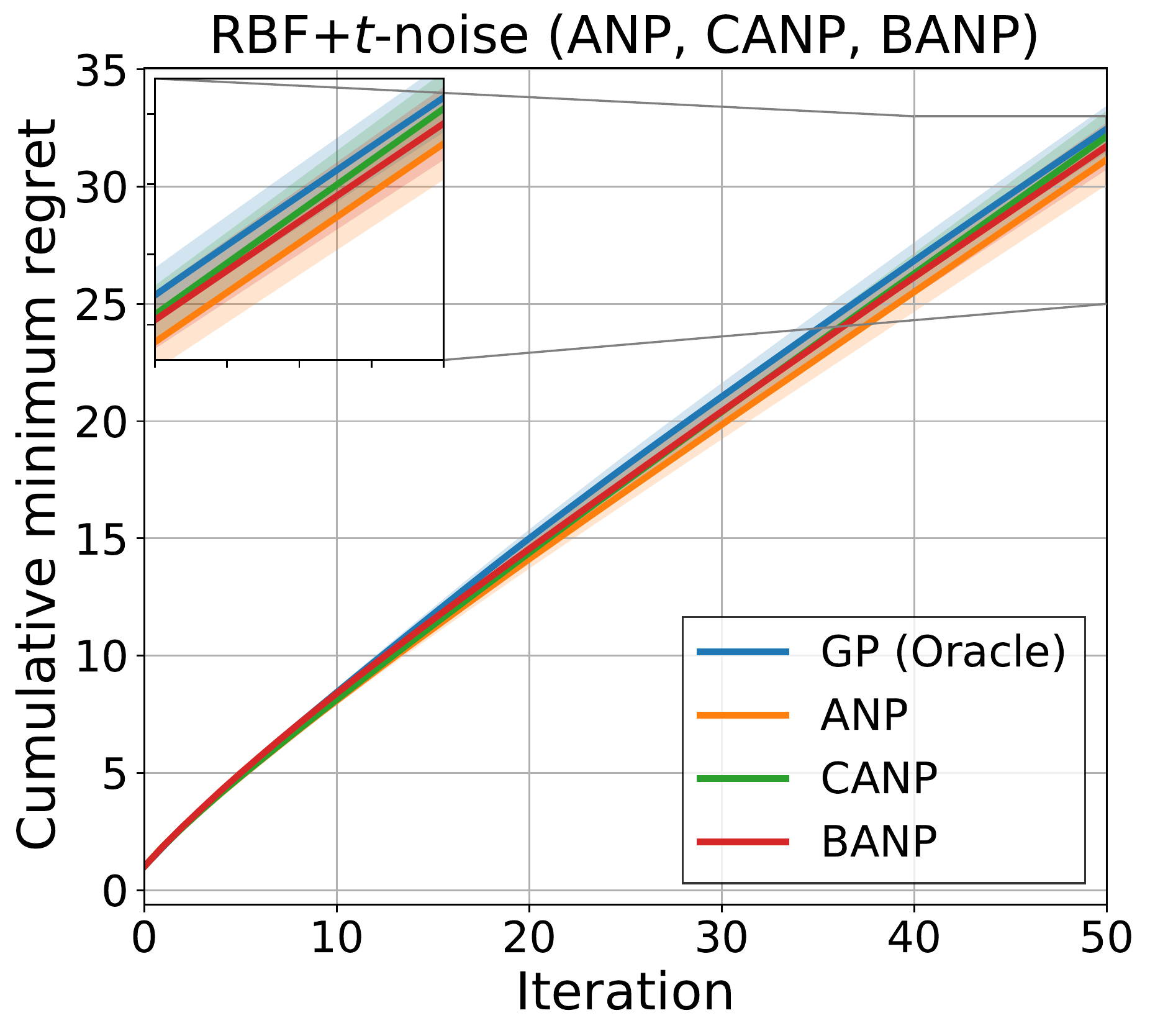}
    \includegraphics[width=0.21\linewidth]{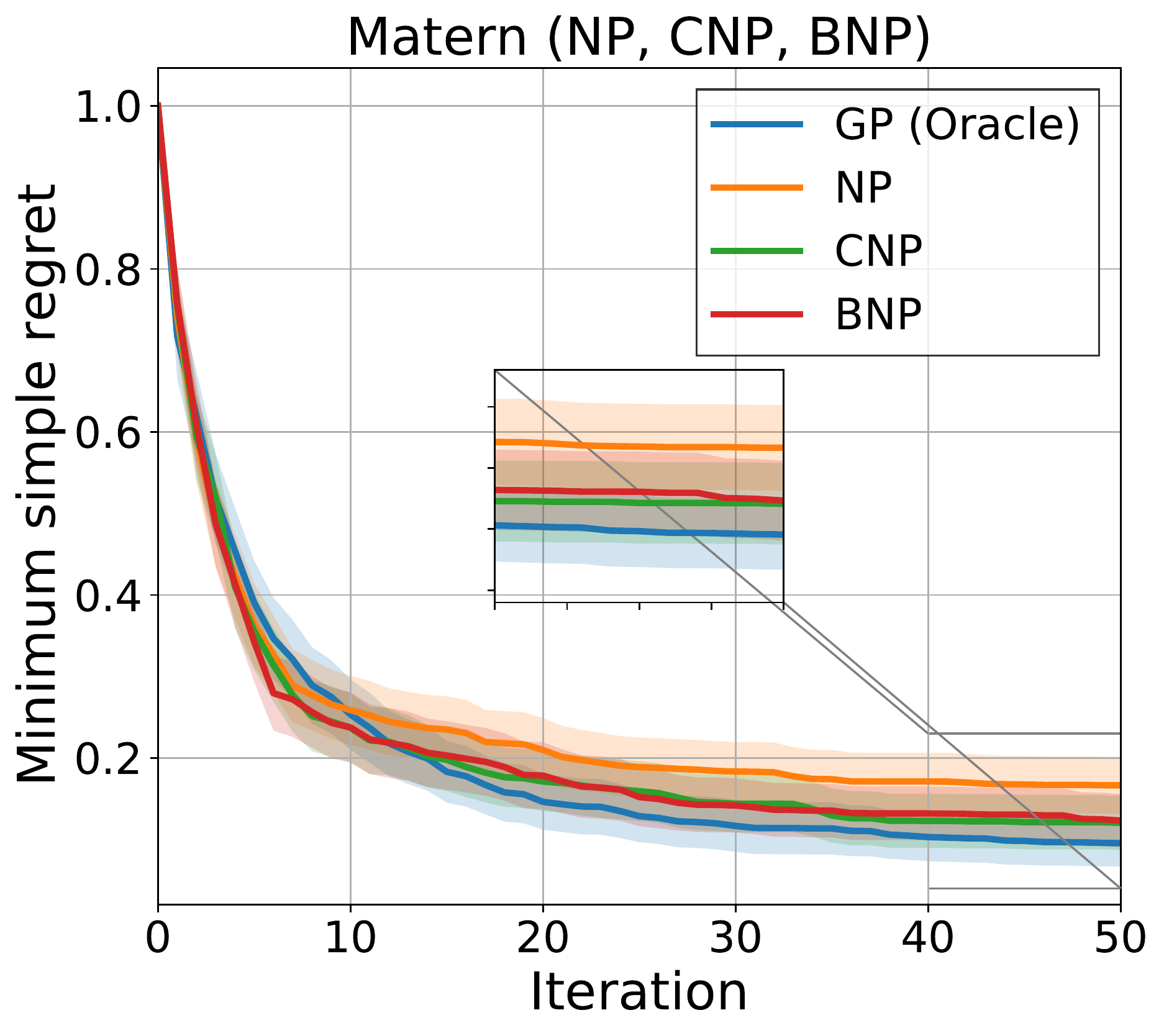}
    \includegraphics[width=0.21\linewidth]{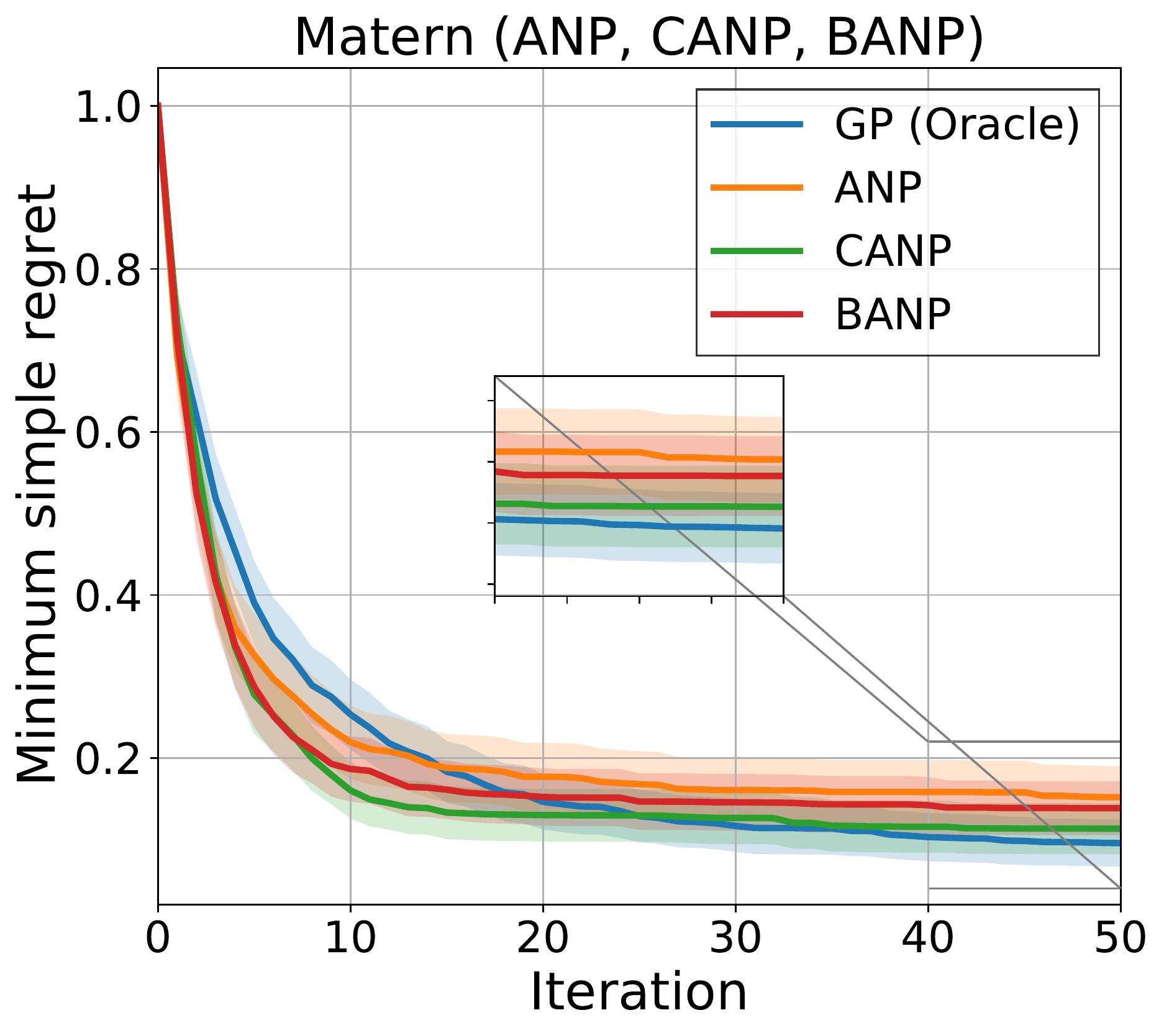}
    \includegraphics[width=0.21\linewidth]{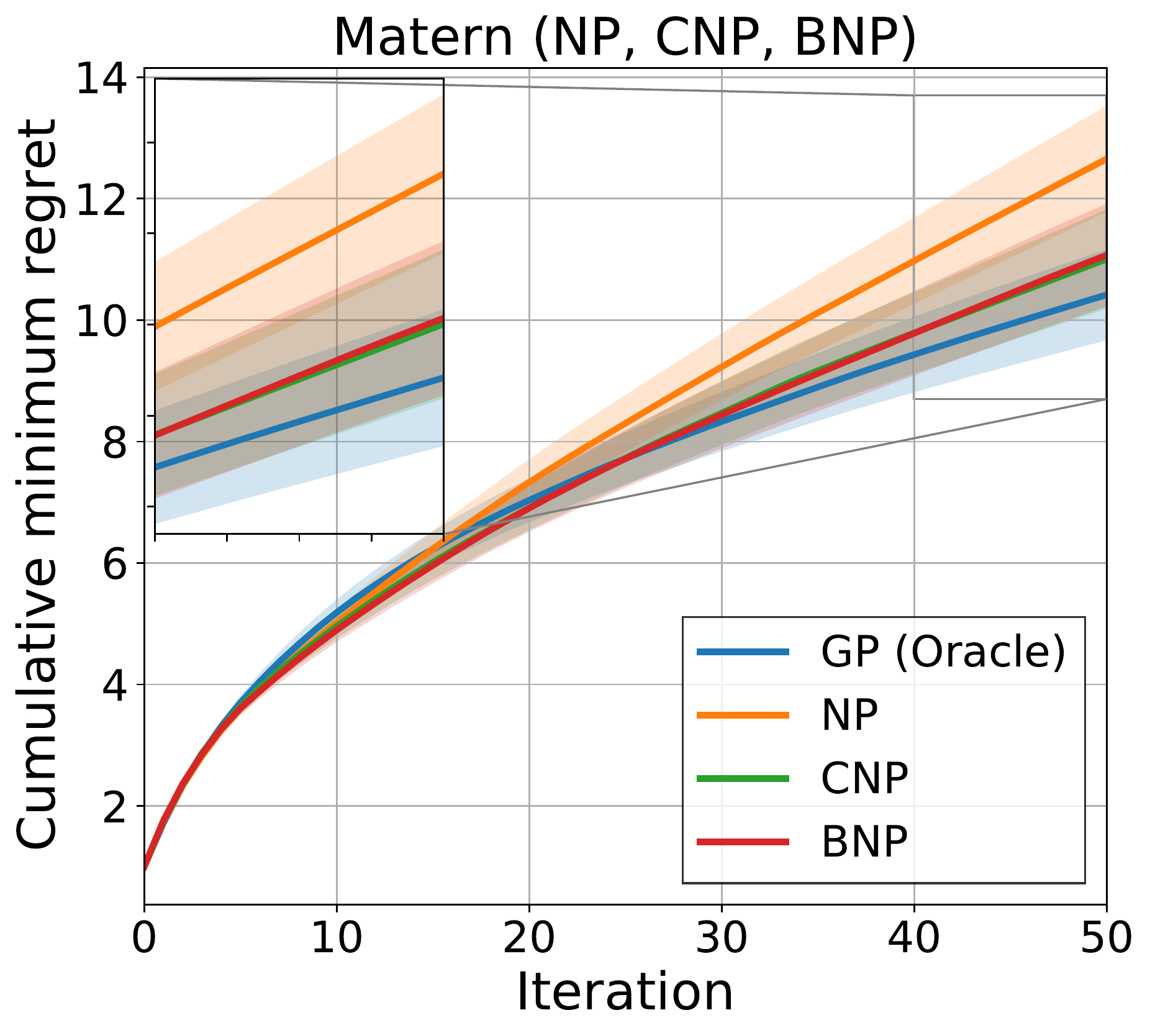}
    \includegraphics[width=0.21\linewidth]{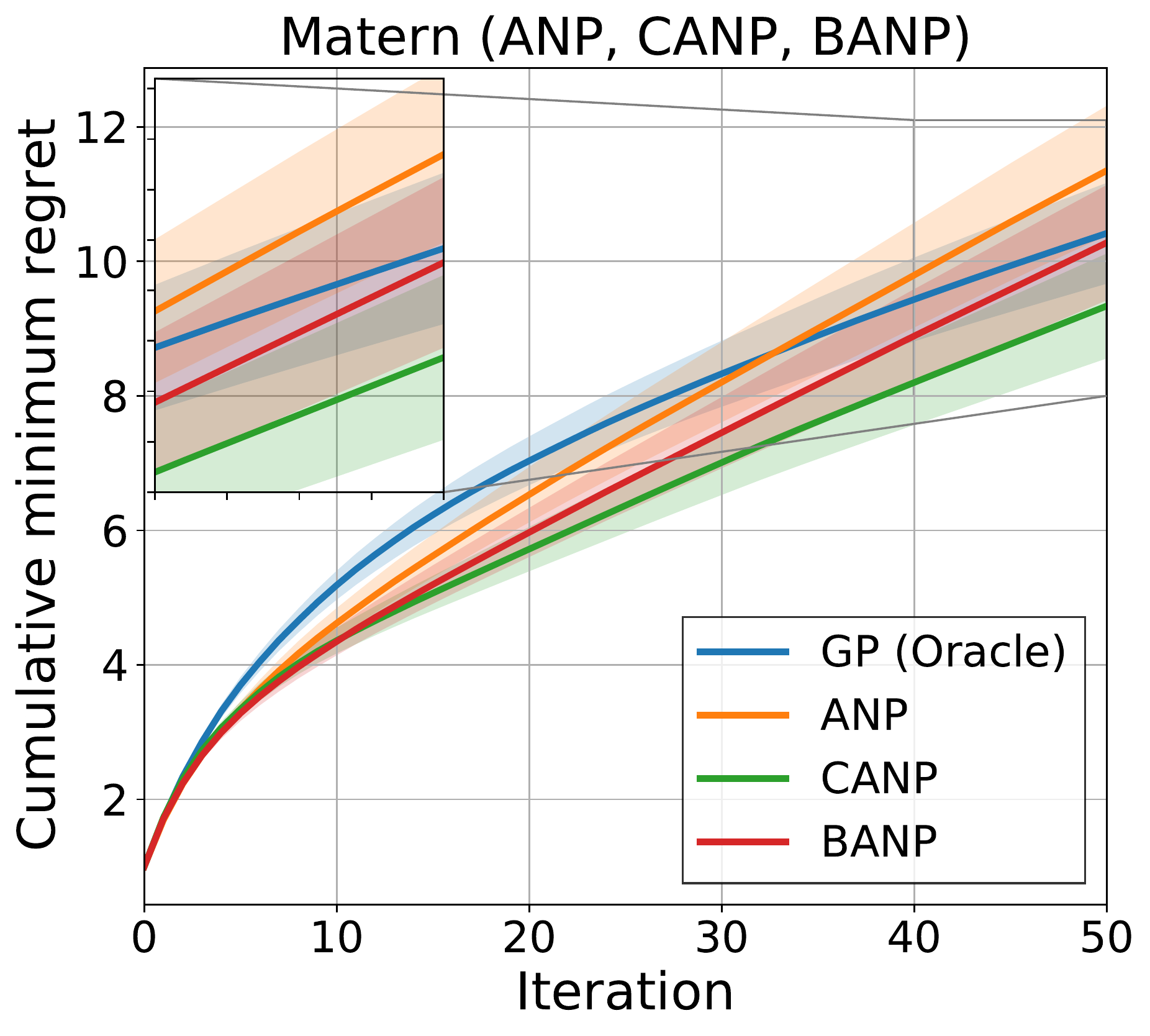}
    \includegraphics[width=0.21\linewidth]{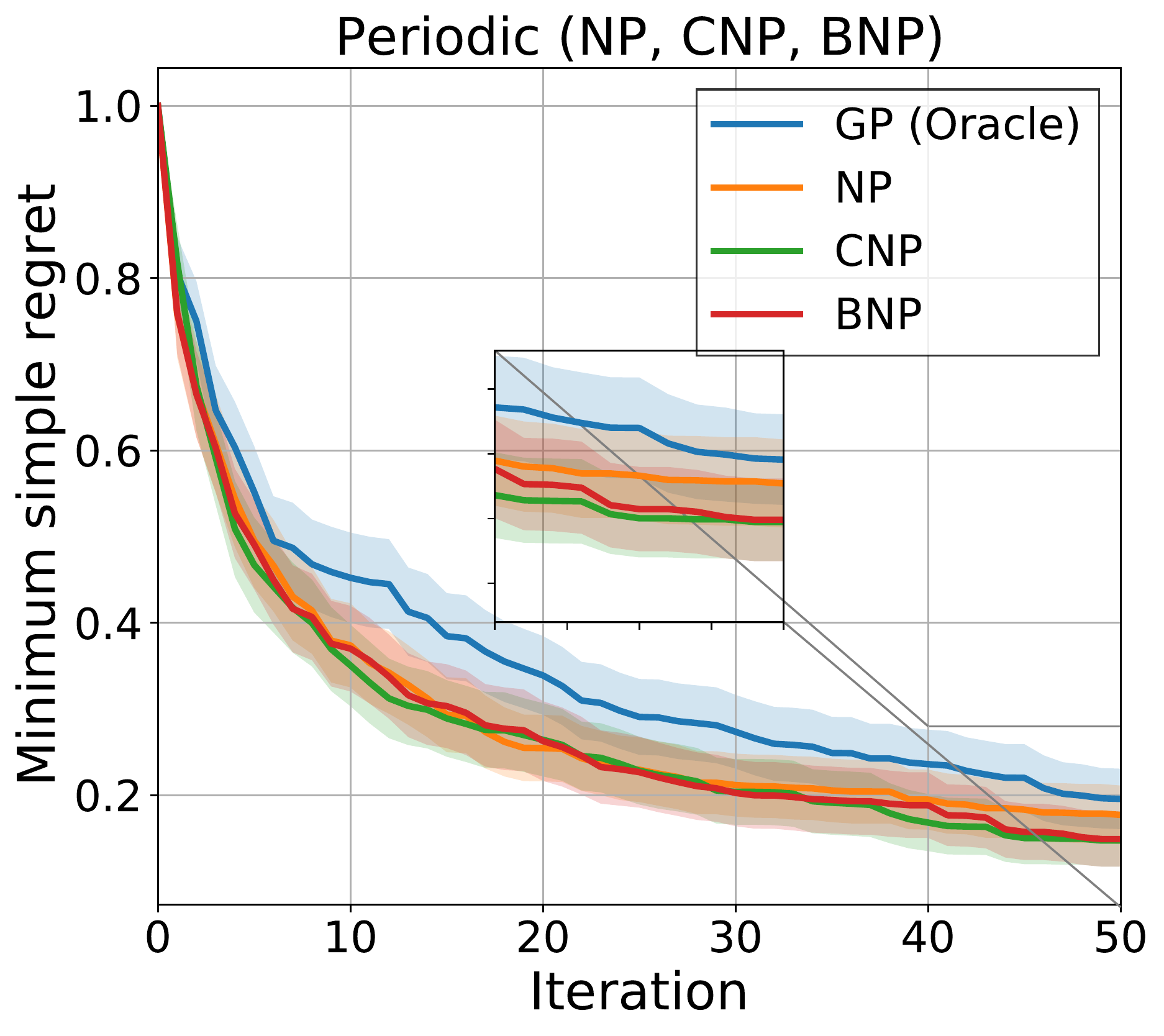}
    \includegraphics[width=0.21\linewidth]{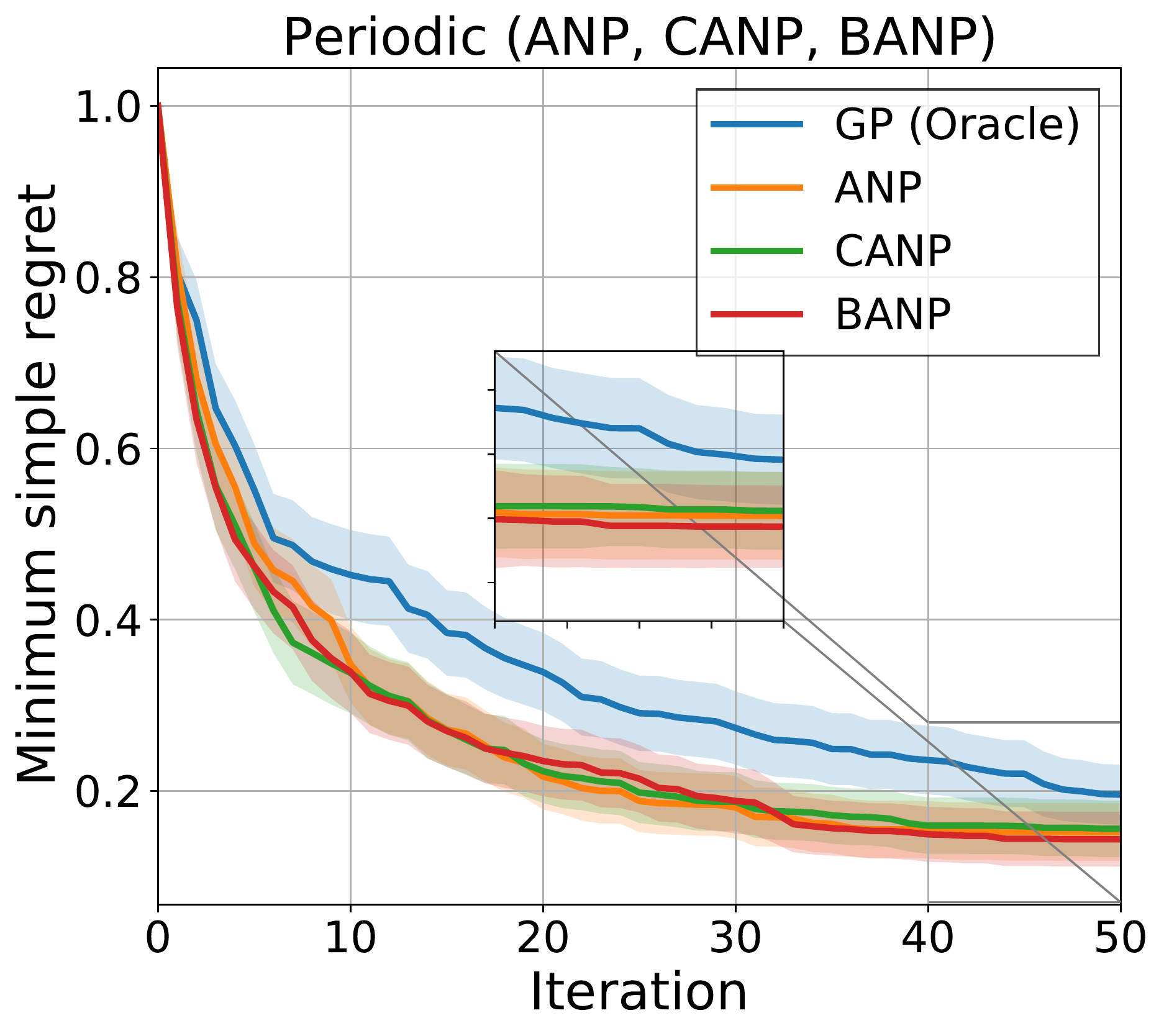}
    \includegraphics[width=0.21\linewidth]{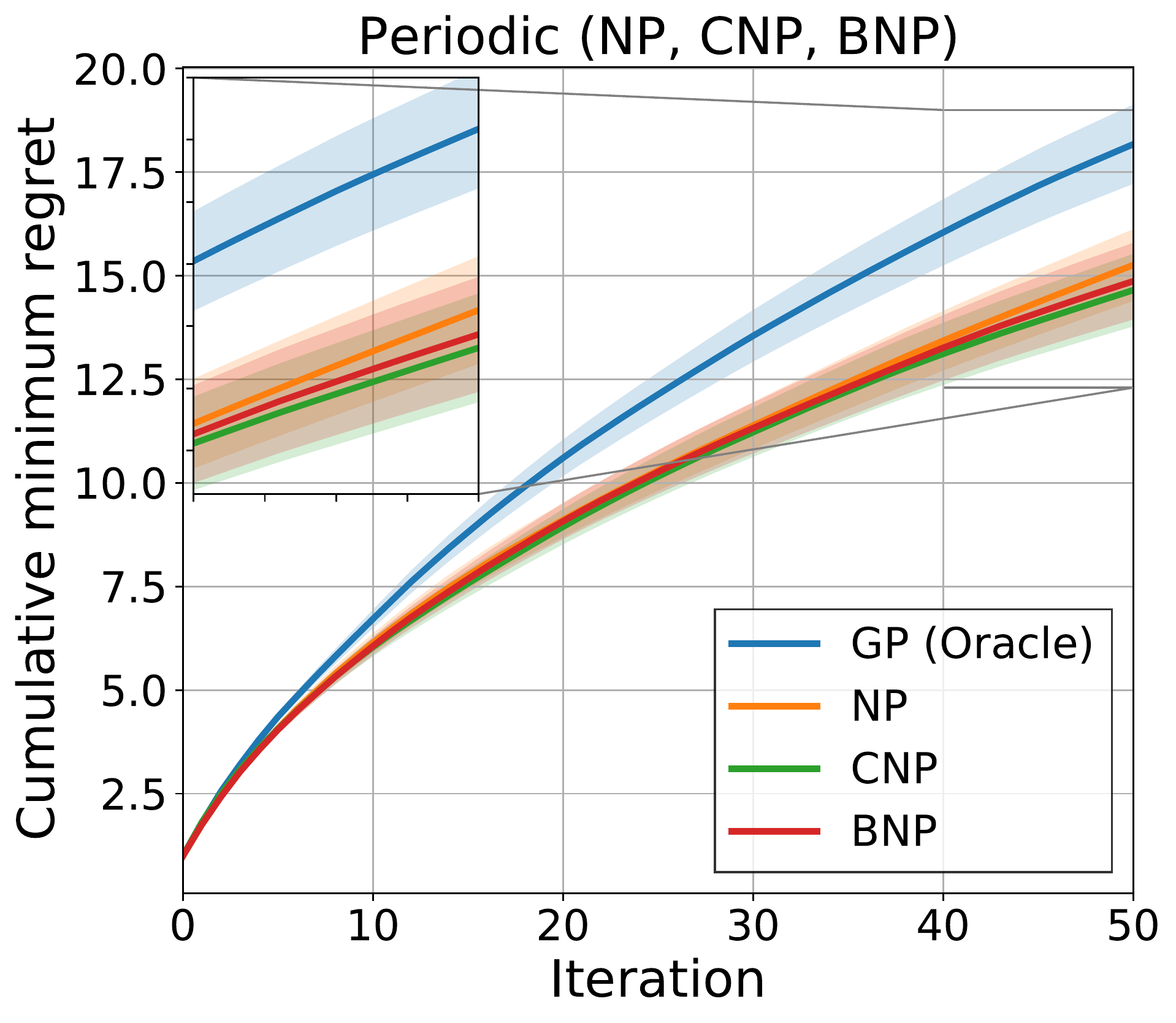}
    \includegraphics[width=0.21\linewidth]{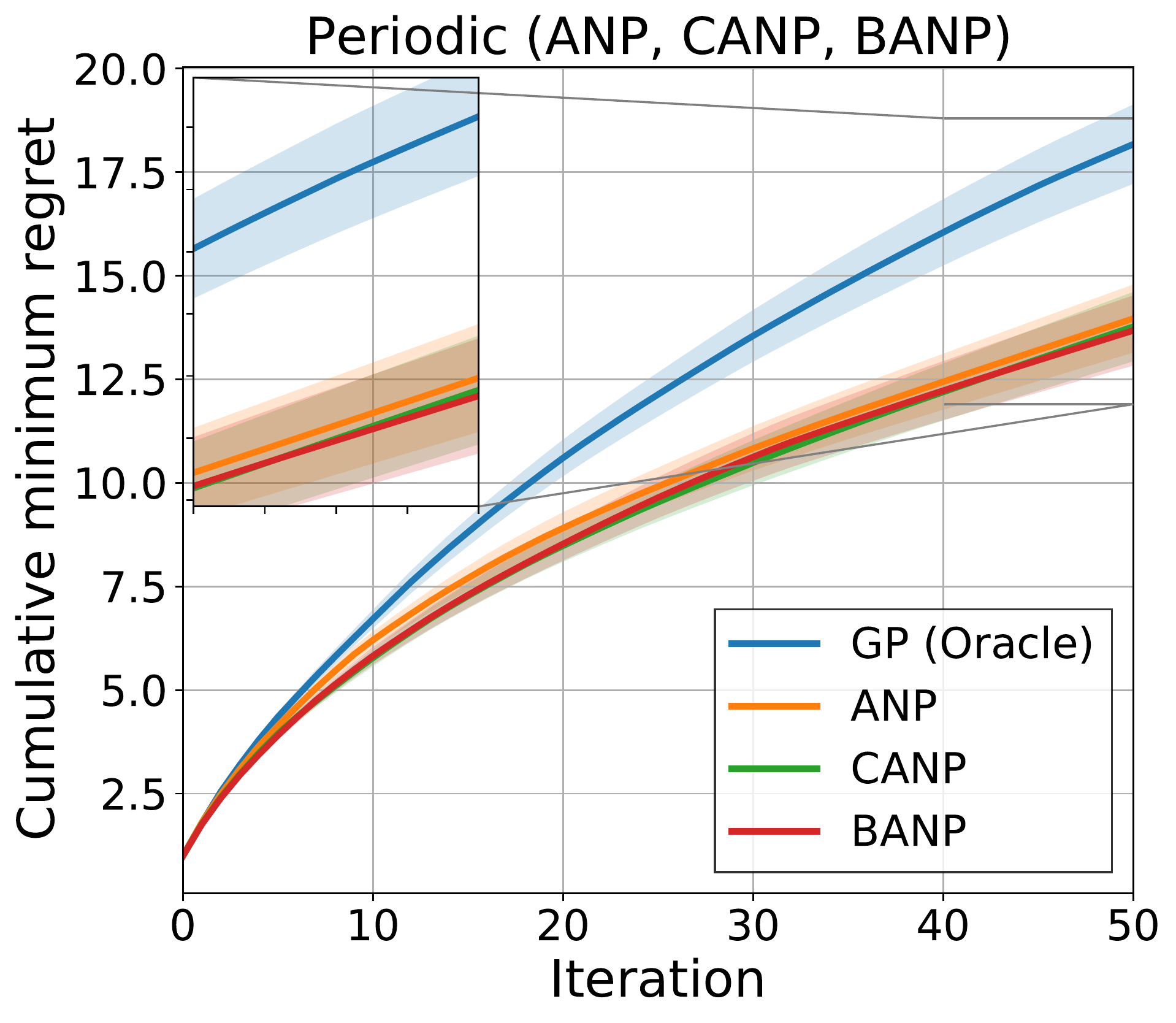}
    \caption{Bayesian optimization results for GP prior functions with (first row) RBF kernel, (second row) RBF kernel + $t$-noise, (third row) Mat\'ern 5/2 kernel, and (fourth row) Periodic kernel.}
    \label{sup:fig:bo_all}
\end{figure}

Bayesian optimization results, showed in~\cref{sup:fig:bo_all} demonstrate 
our methods outperform or are comparable to other methods including GP oracle.
For the RBF case, GP oracle is the best result, 
but our models show the second best results and become comparable to the GP oracle 
at the last of iterations.
On the contrary, in the model-data mismatch setting with $t$-noise (see the second row of \cref{sup:fig:bo_all}), 
our methods outperform other methods, 
which implies that our methods, BNP and BANP are robust to the heavy-tailed noises.
Moreover, while CNP and CANP models show the better results in Mat\'ern 5/2 and Periodic cases, 
our methods are comparable to those methods, as shown in the last two rows of \cref{sup:fig:bo_all}.

\subsection{Image completion} 
\label{subsec:image_completion_additional}
\begin{figure}
    \centering
    \includegraphics[width=0.45\linewidth]{figures/emnist/emnist_24_0-10.pdf} \hspace{0.2cm}  \includegraphics[width=0.45\linewidth]{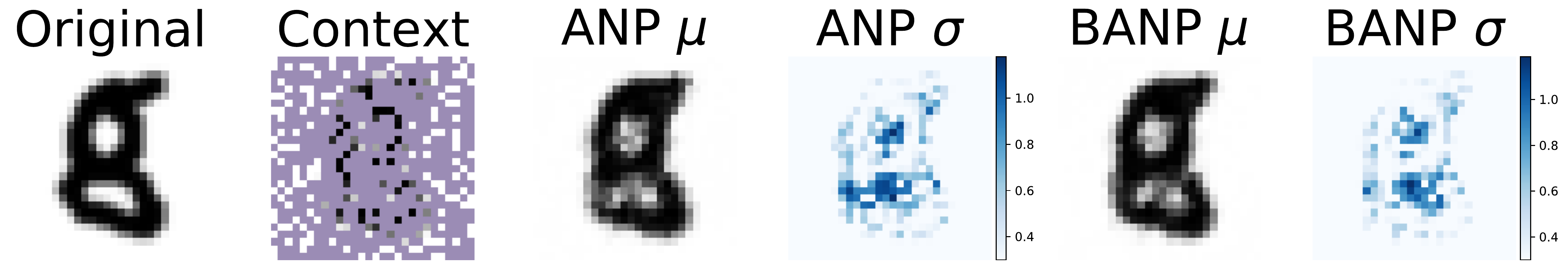} \\ 
    \includegraphics[width=0.45\linewidth]{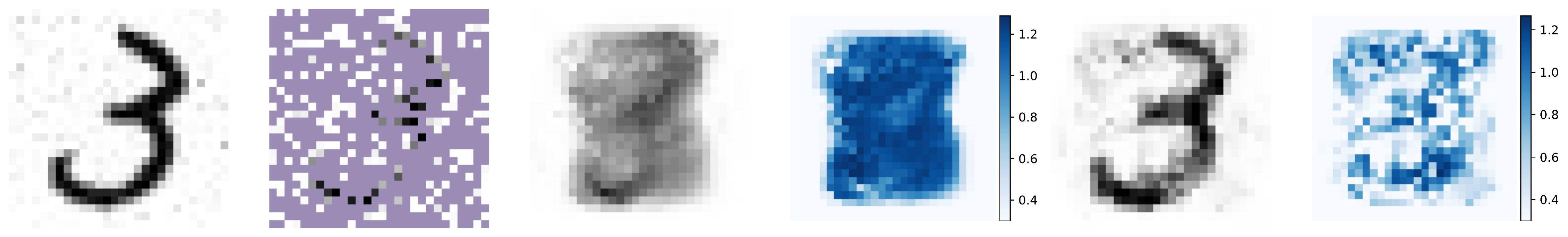} \hspace{0.2cm} \includegraphics[width=0.45\linewidth]{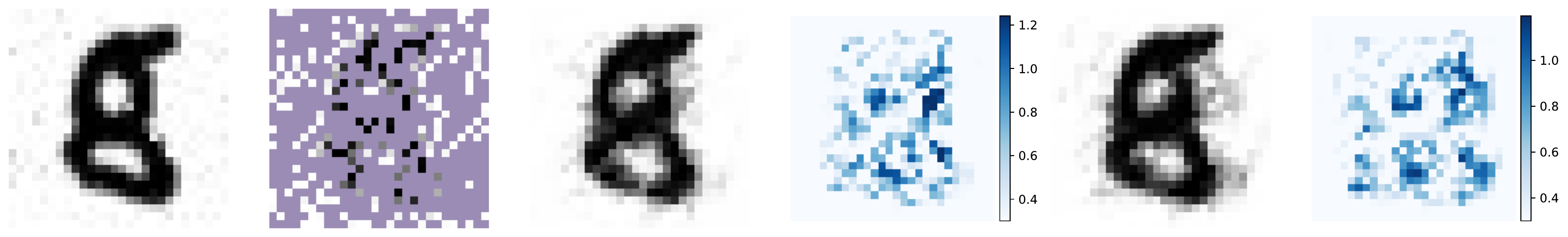} \\ 
    \includegraphics[width=0.45\linewidth]{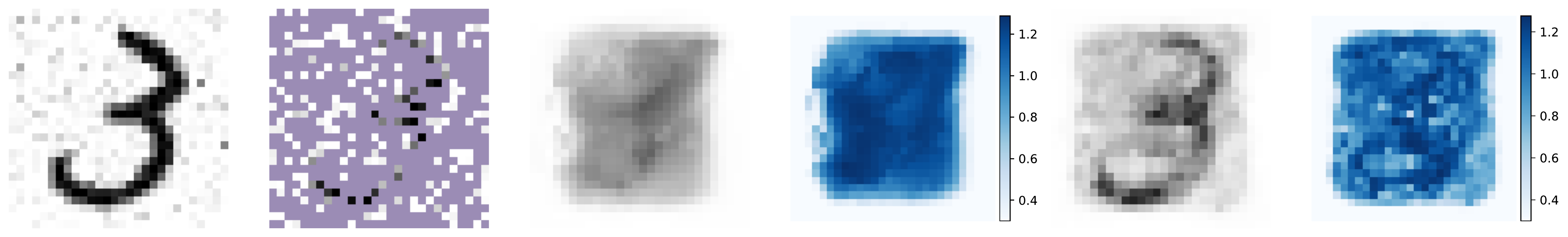} \hspace{0.2cm} \includegraphics[width=0.45\linewidth]{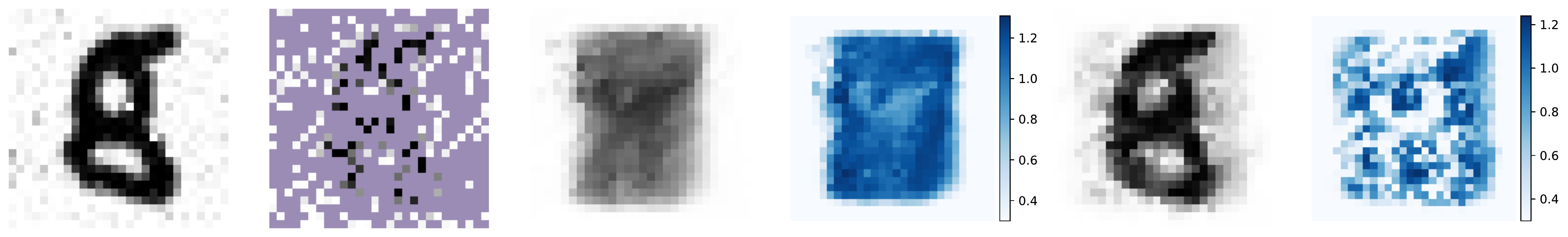} \\ 
    \includegraphics[width=0.45\linewidth]{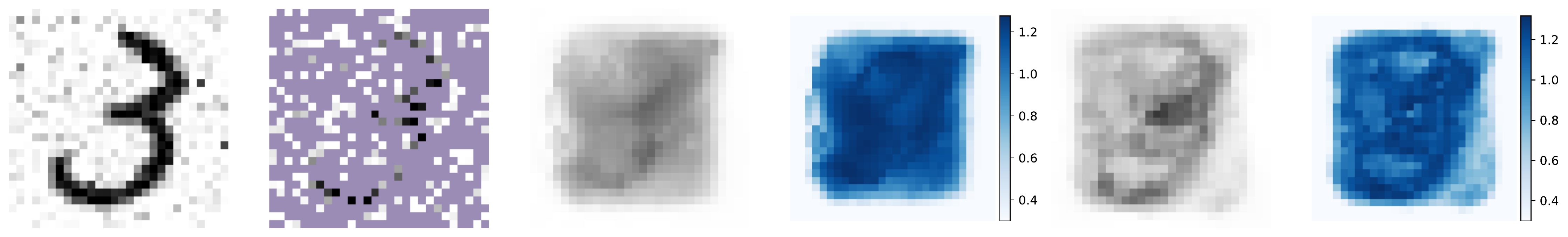} \hspace{0.2cm} \includegraphics[width=0.45\linewidth]{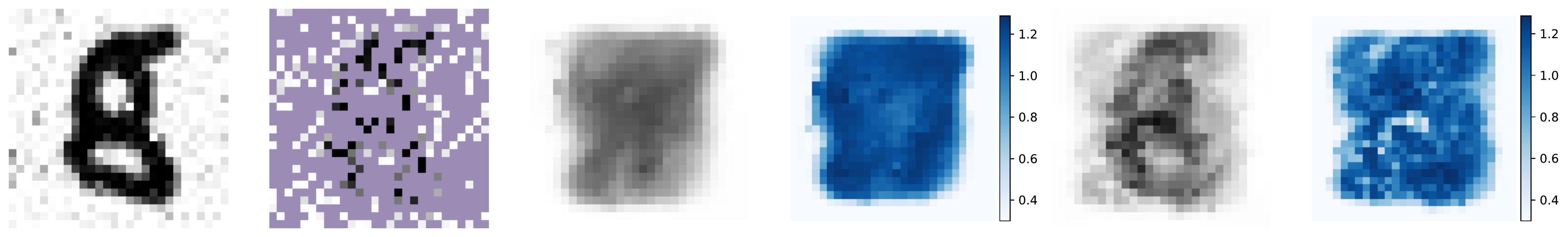} \\ 
    \vspace{0.2cm}
    \includegraphics[width=0.45\linewidth]{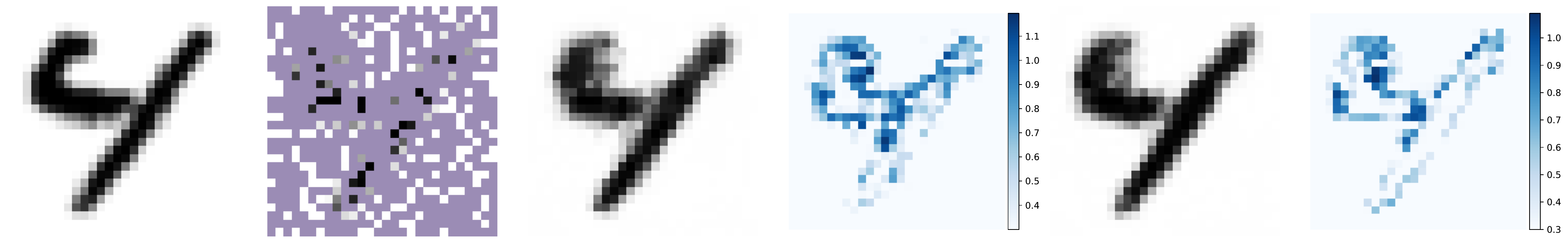} \hspace{0.2cm}  \includegraphics[width=0.45\linewidth]{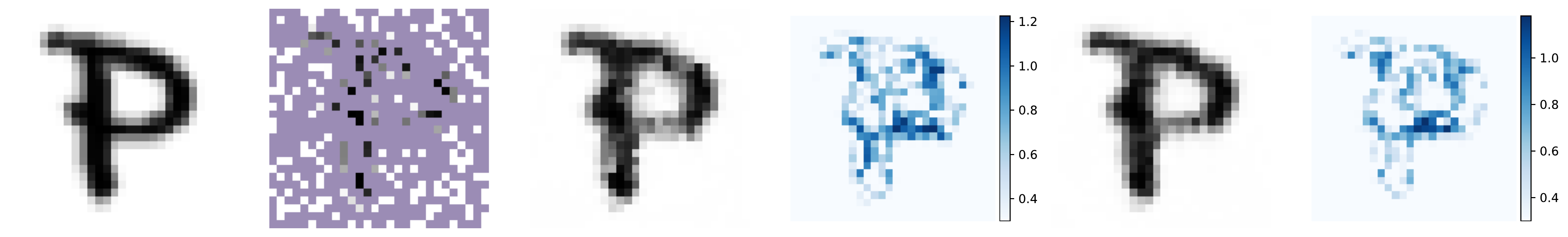} \\ 
    \includegraphics[width=0.45\linewidth]{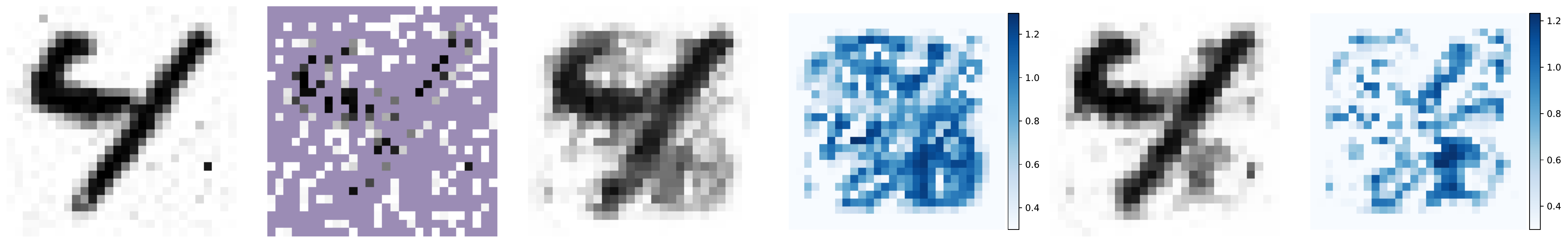} \hspace{0.2cm}  \includegraphics[width=0.45\linewidth]{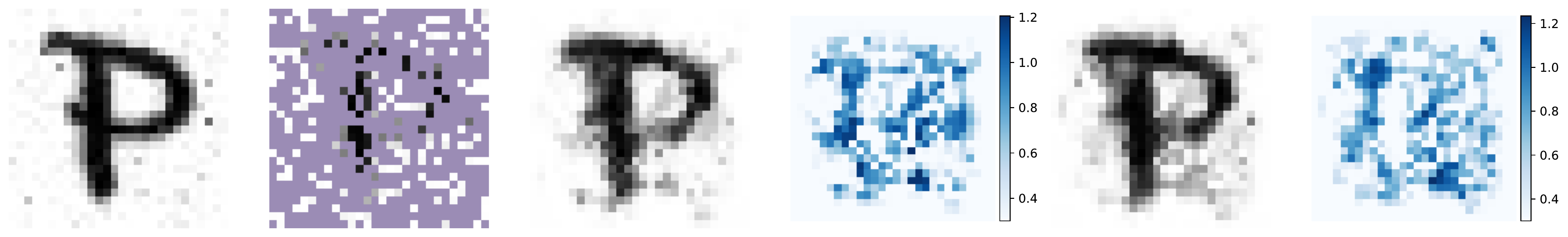} \\ 
    \includegraphics[width=0.45\linewidth]{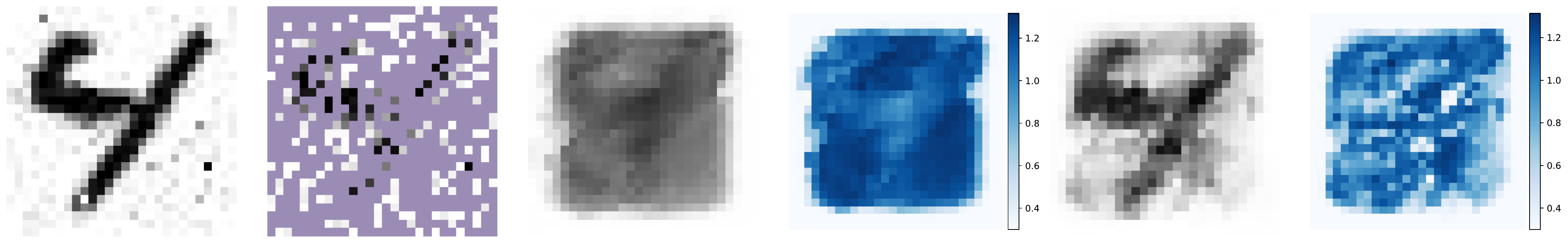} \hspace{0.2cm}  \includegraphics[width=0.45\linewidth]{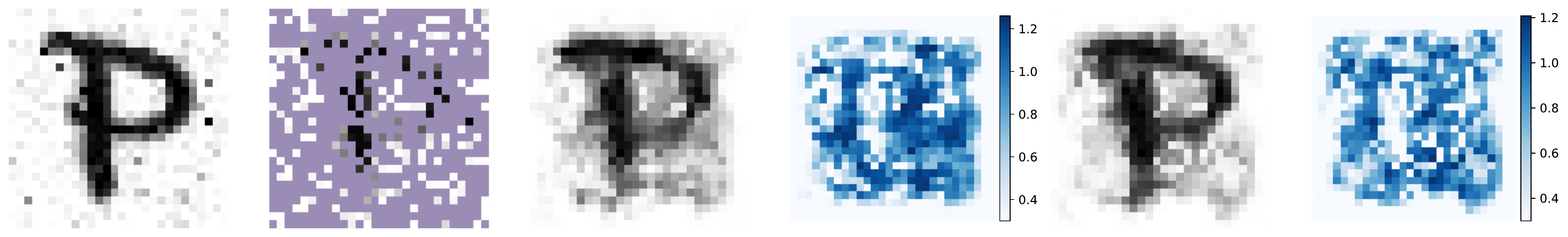} \\ 
    \includegraphics[width=0.45\linewidth]{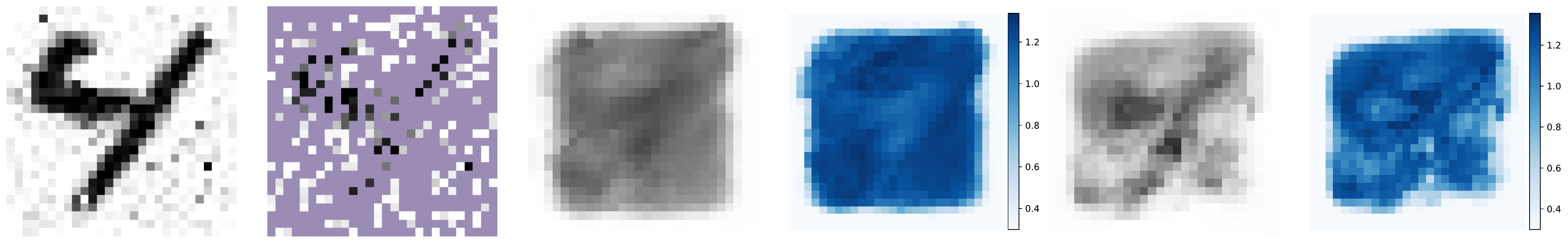} \hspace{0.2cm}  \includegraphics[width=0.45\linewidth]{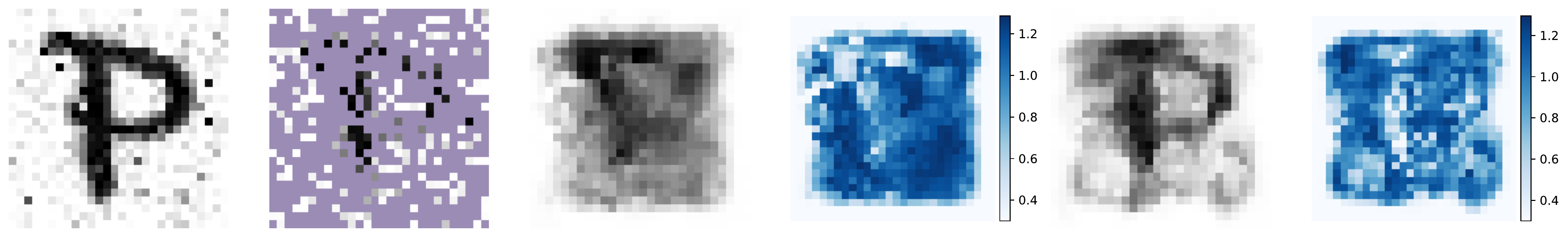} \\ 
    \vspace{0.2cm}
    \includegraphics[width=0.45\linewidth]{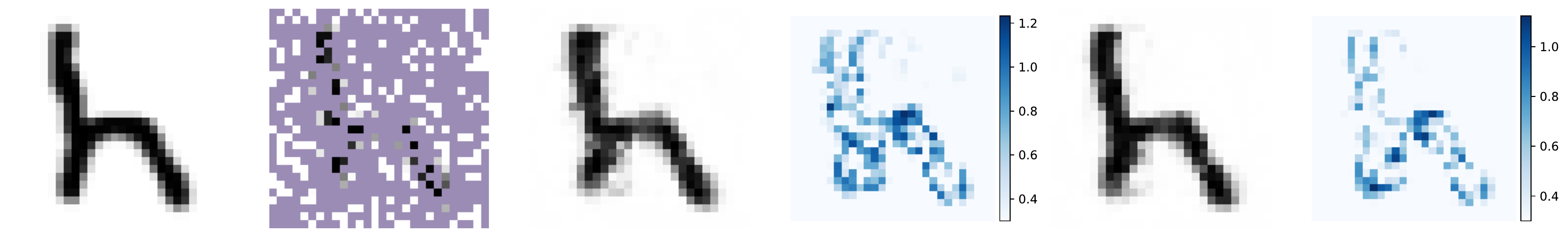} \hspace{0.2cm}  \includegraphics[width=0.45\linewidth]{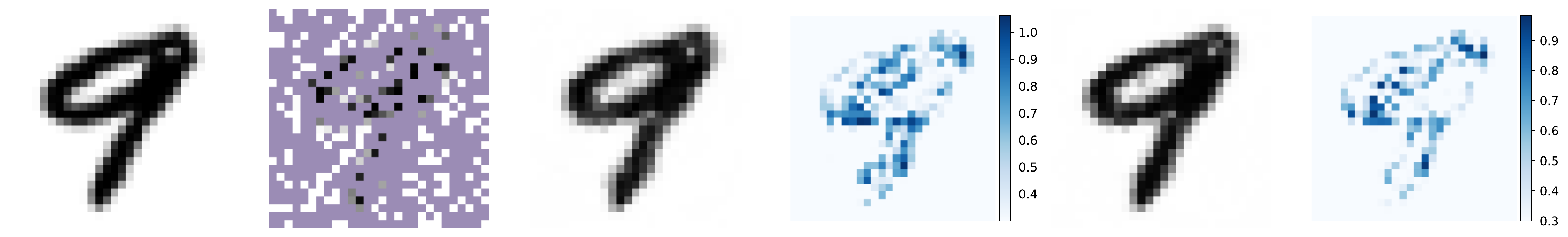} \\ 
    \includegraphics[width=0.45\linewidth]{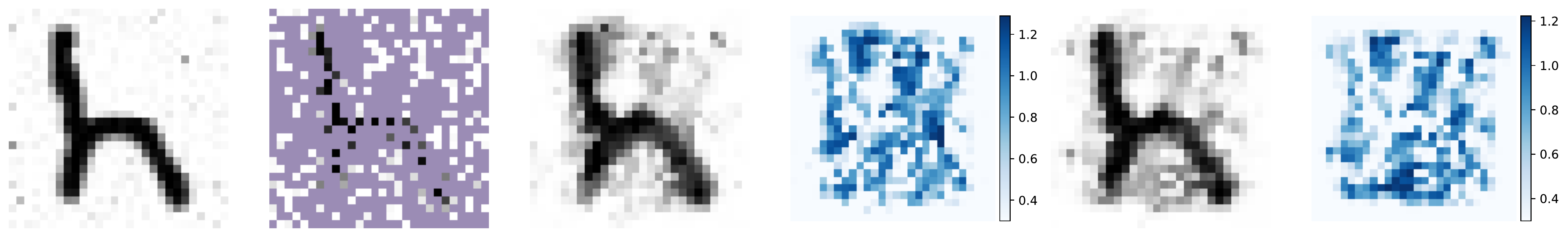} \hspace{0.2cm}  \includegraphics[width=0.45\linewidth]{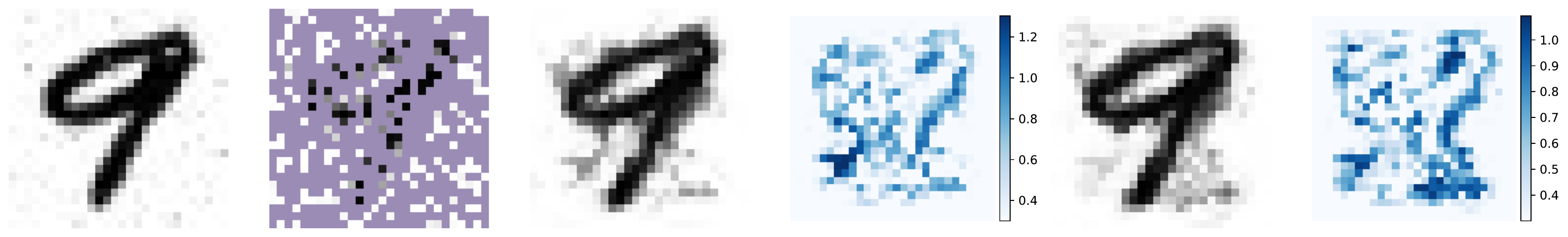}\\ 
    \includegraphics[width=0.45\linewidth]{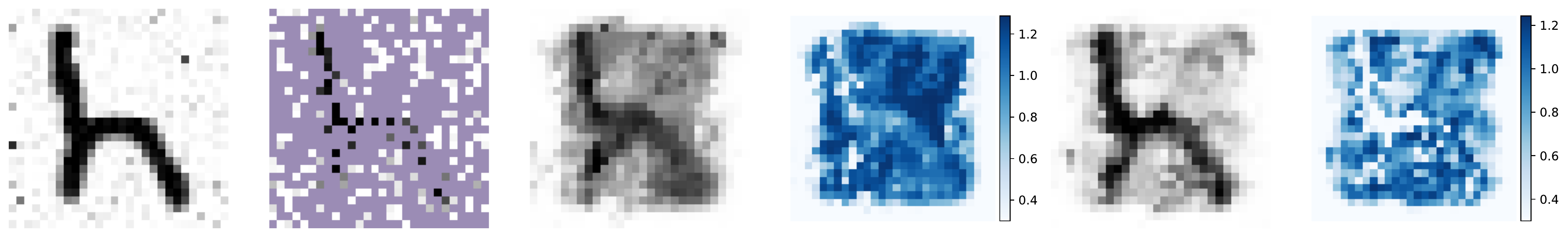} \hspace{0.2cm}  \includegraphics[width=0.45\linewidth]{figures/emnist/emnist_77777_0-47_004.pdf} \\ 
    \includegraphics[width=0.45\linewidth]{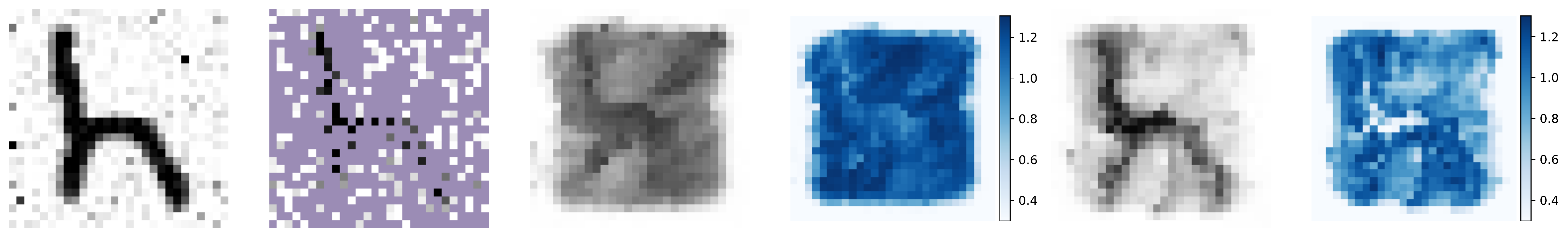} \hspace{0.2cm}  \includegraphics[width=0.45\linewidth]{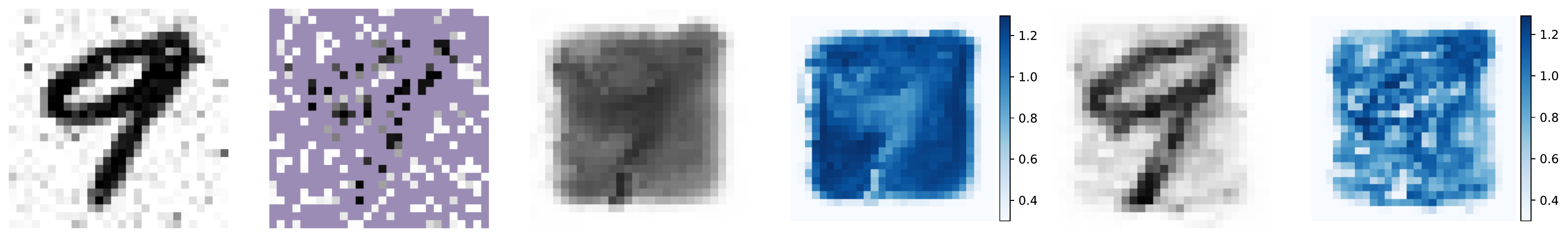} 
    \caption{Image completion results on EMNIST for \gls{anp} and \gls{banp}. The results under increasing noise levels are shown.}
    \label{sup:fig:emnist_more_vis}
\end{figure}

\begin{figure}
    \centering
    \includegraphics[width=0.45\linewidth]{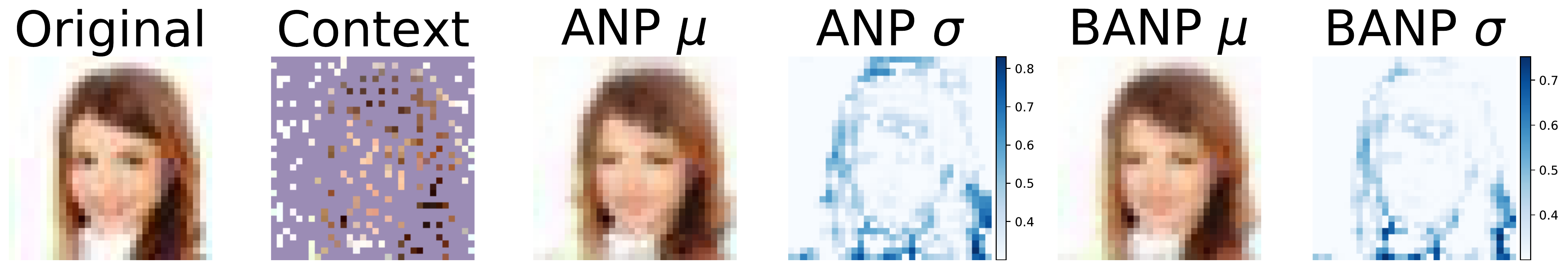} \hspace{0.2cm} \includegraphics[width=0.45\linewidth]{figures/celeba/celeba_423.pdf} \\
    \includegraphics[width=0.45\linewidth]{figures/celeba/celeba_1234_002.pdf} \hspace{0.2cm} \includegraphics[width=0.45\linewidth]{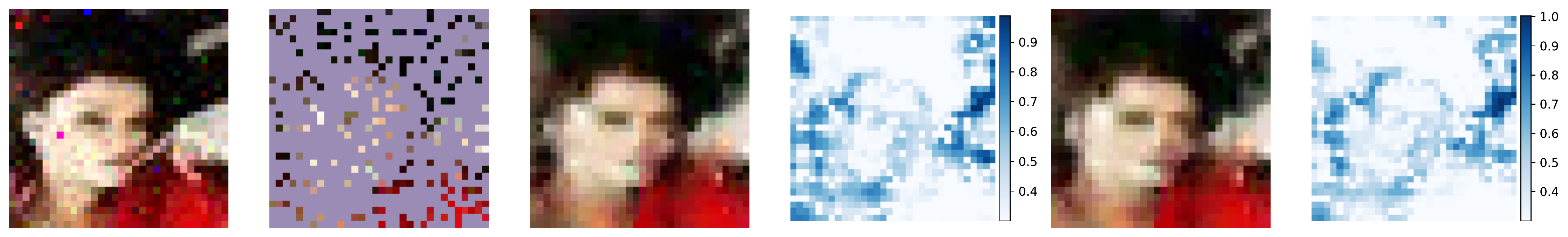} \\
    \includegraphics[width=0.45\linewidth]{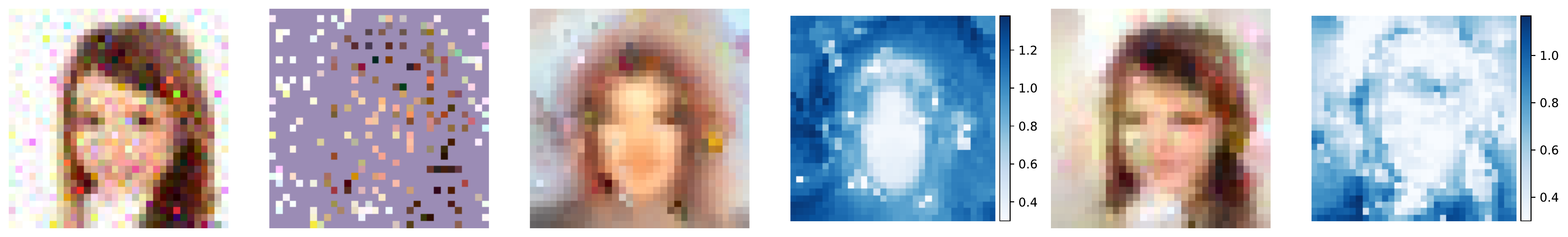} \hspace{0.2cm} \includegraphics[width=0.45\linewidth]{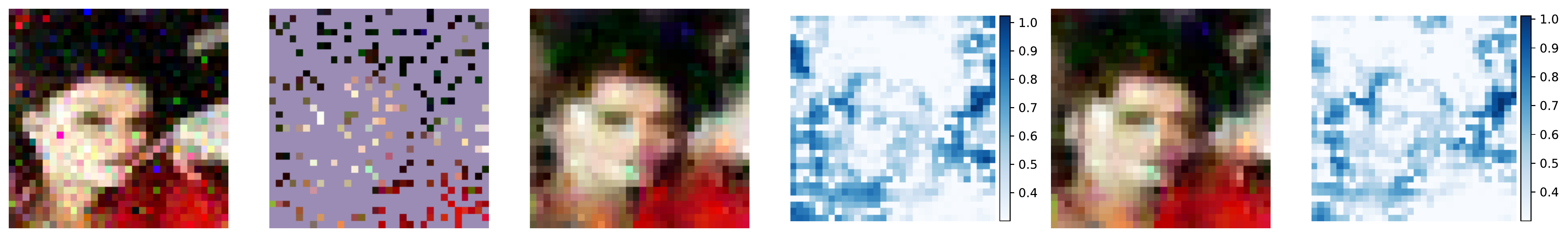} \\
    \includegraphics[width=0.45\linewidth]{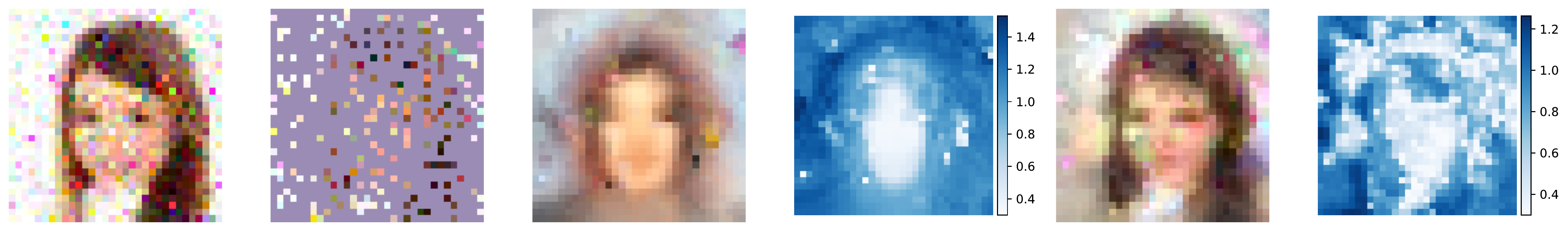} \hspace{0.2cm} \includegraphics[width=0.45\linewidth]{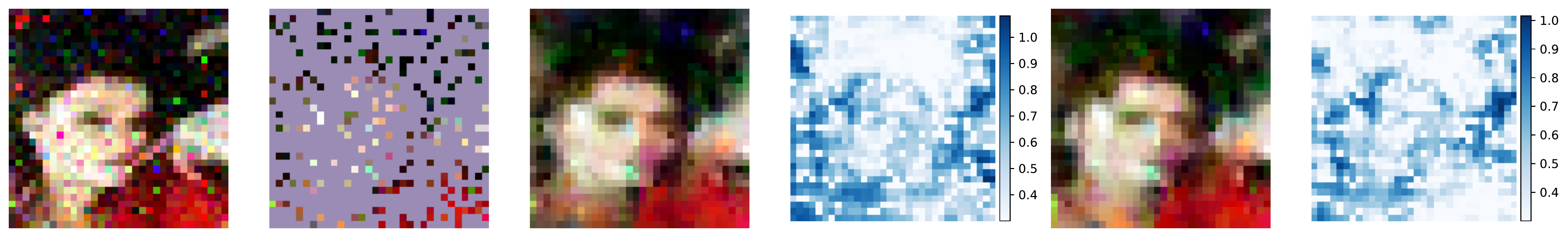} \\
    \vspace{0.2cm}
    \includegraphics[width=0.45\linewidth]{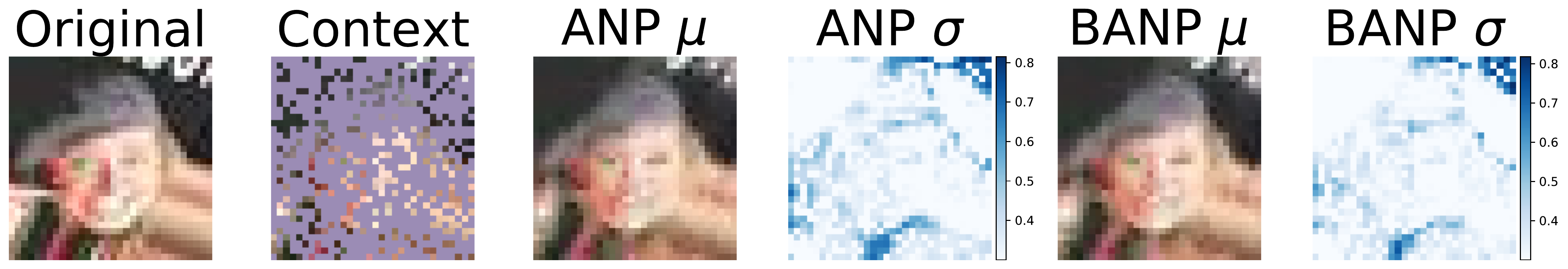} \hspace{0.2cm} \includegraphics[width=0.45\linewidth]{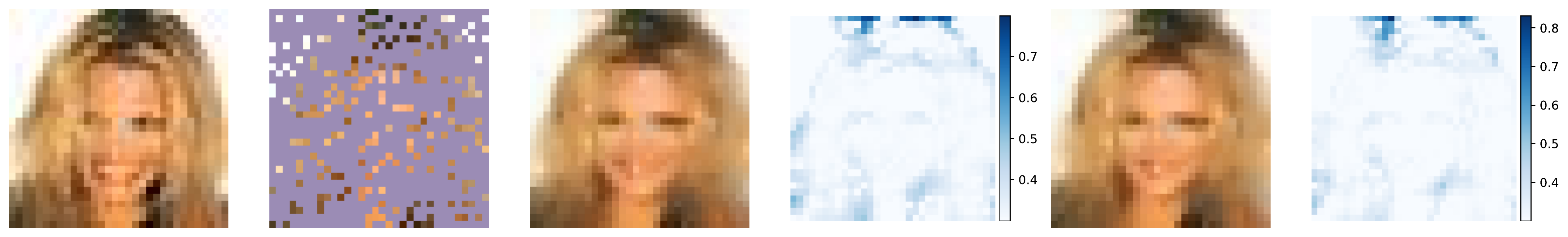} \\
    \includegraphics[width=0.45\linewidth]{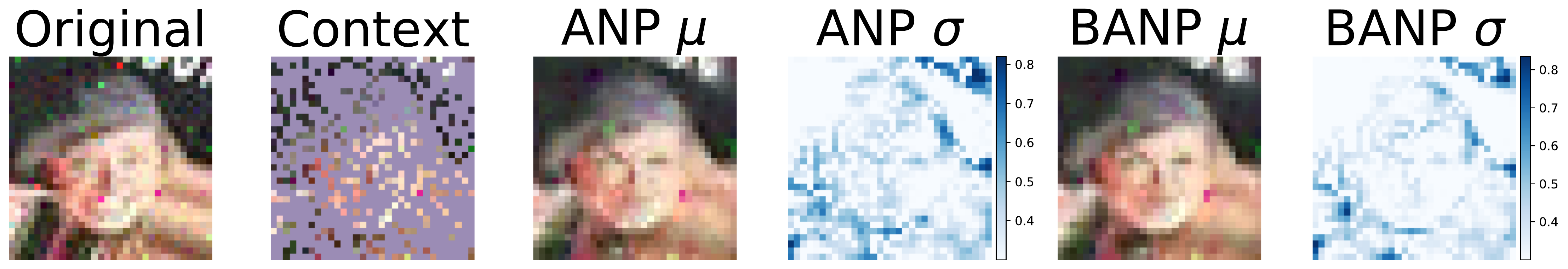} \hspace{0.2cm} \includegraphics[width=0.45\linewidth]{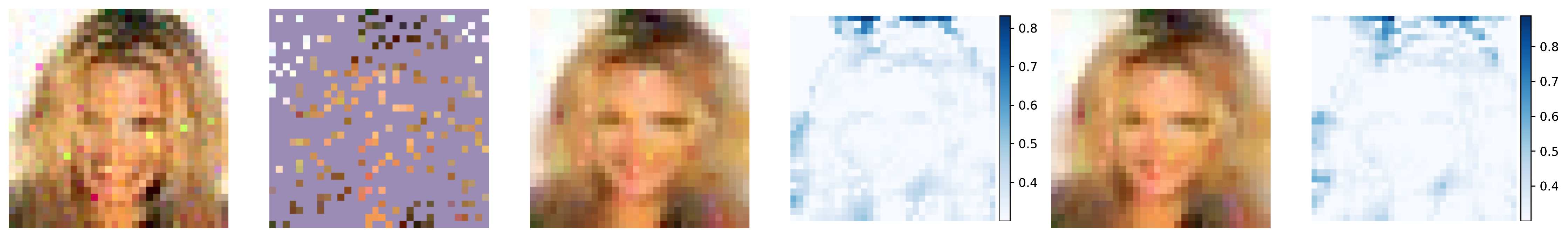} \\
    \includegraphics[width=0.45\linewidth]{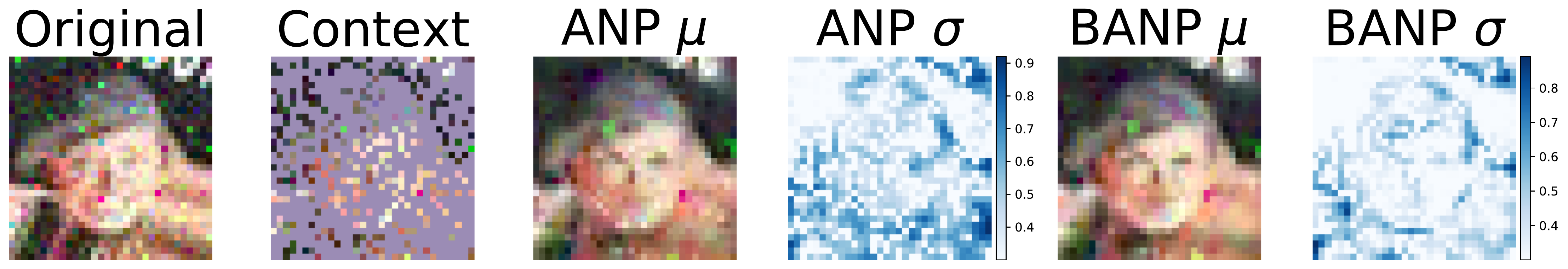} \hspace{0.2cm} \includegraphics[width=0.45\linewidth]{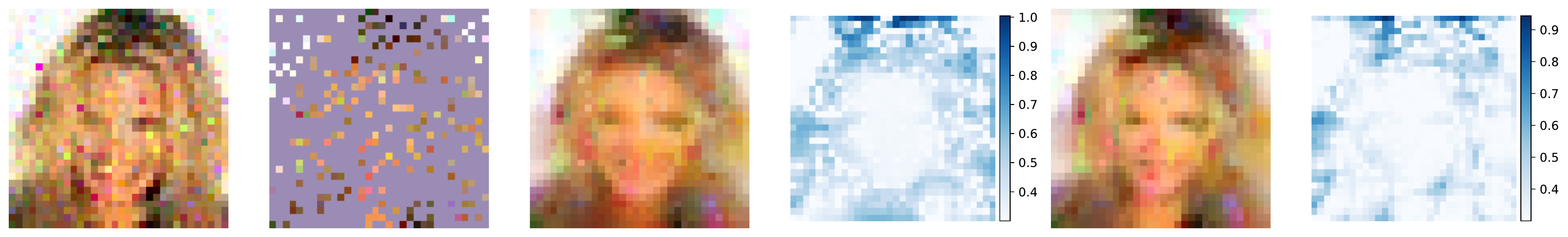} \\
    \includegraphics[width=0.45\linewidth]{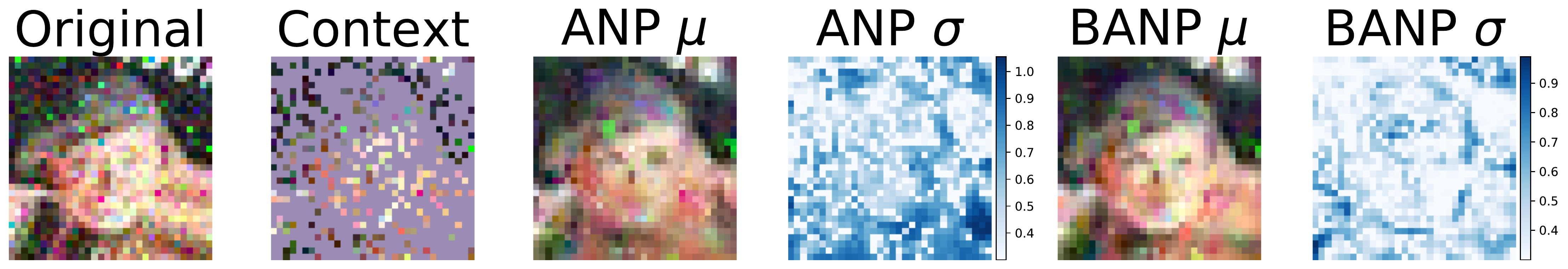} \hspace{0.2cm} \includegraphics[width=0.45\linewidth]{figures/celeba/celeba_450_006.pdf} \\
    \vspace{0.2cm}
    \includegraphics[width=0.45\linewidth]{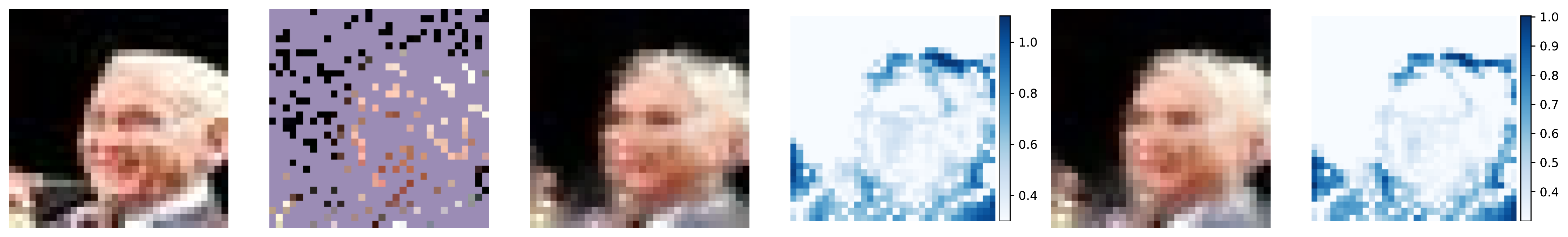} \hspace{0.2cm} \includegraphics[width=0.45\linewidth]{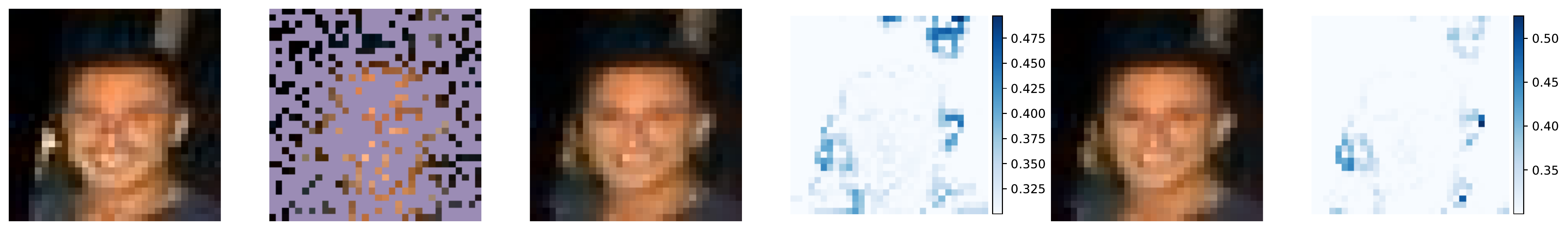} \\
    \includegraphics[width=0.45\linewidth]{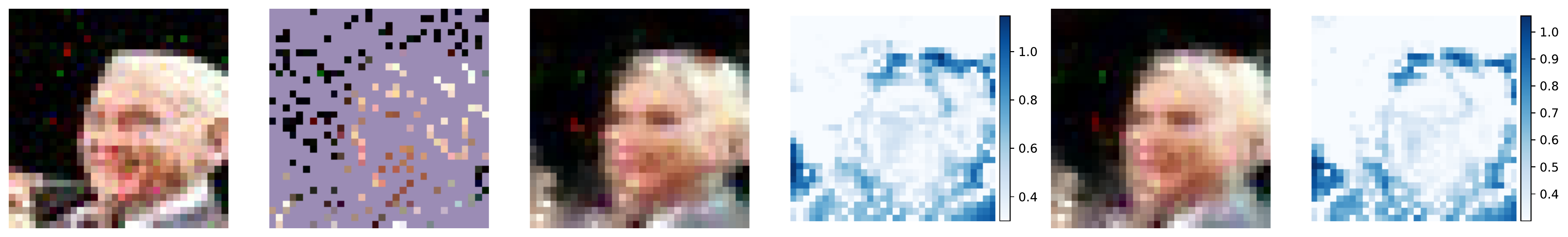} \hspace{0.2cm} \includegraphics[width=0.45\linewidth]{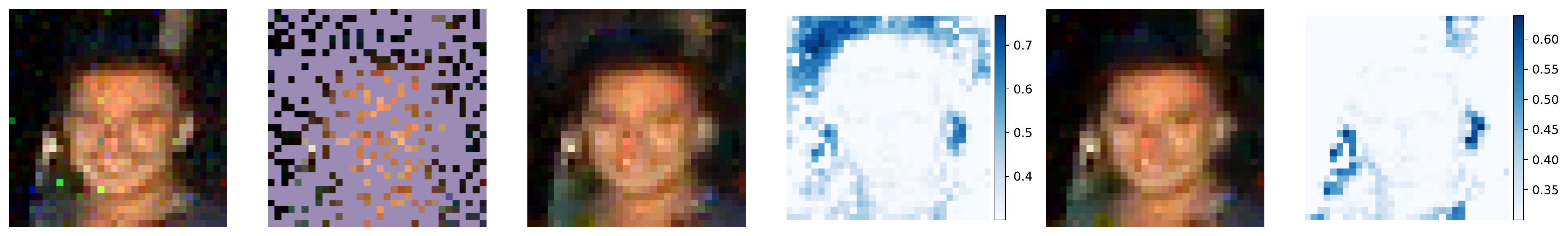} \\
    \includegraphics[width=0.45\linewidth]{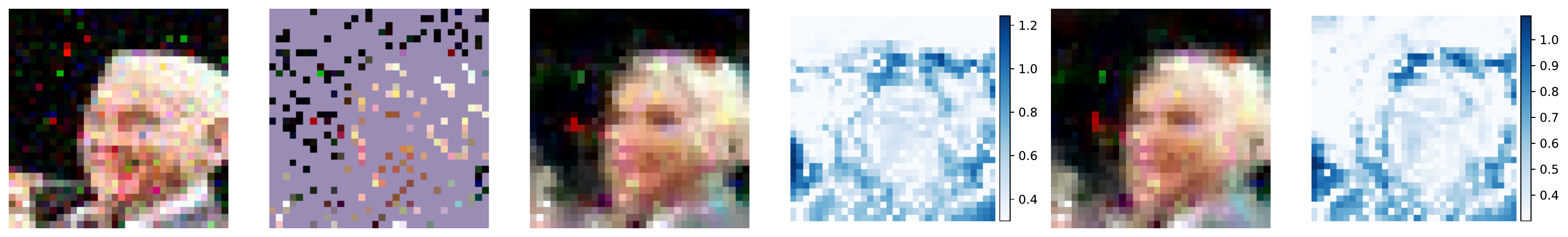} \hspace{0.2cm} \includegraphics[width=0.45\linewidth]{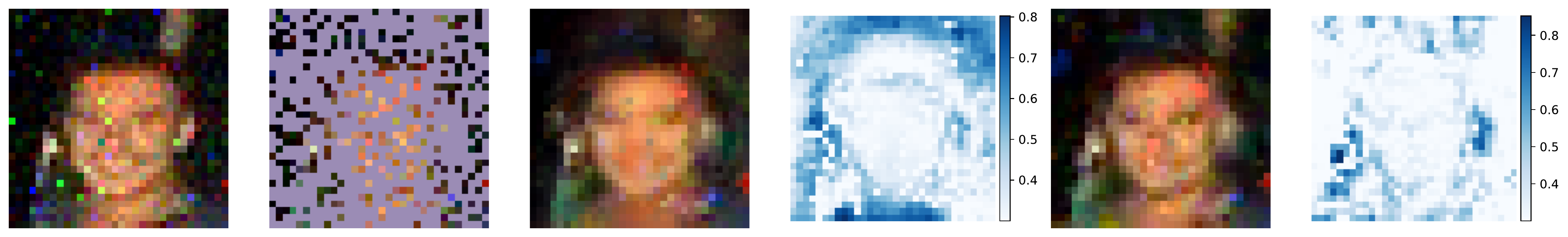} \\
    \includegraphics[width=0.45\linewidth]{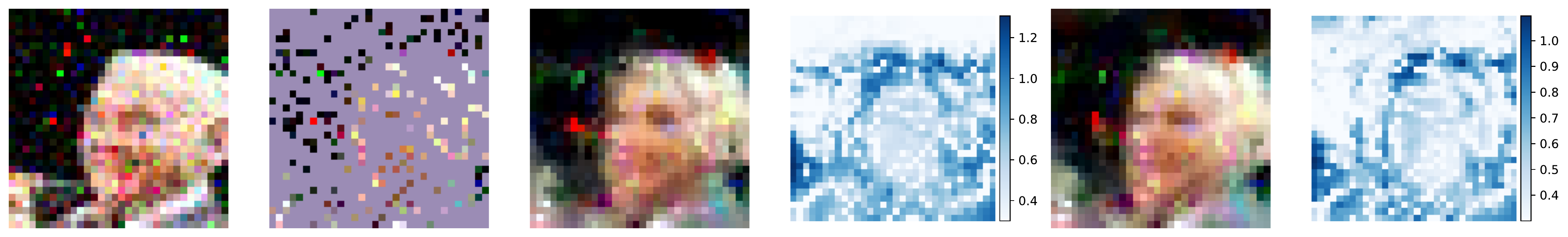} \hspace{0.2cm} \includegraphics[width=0.45\linewidth]{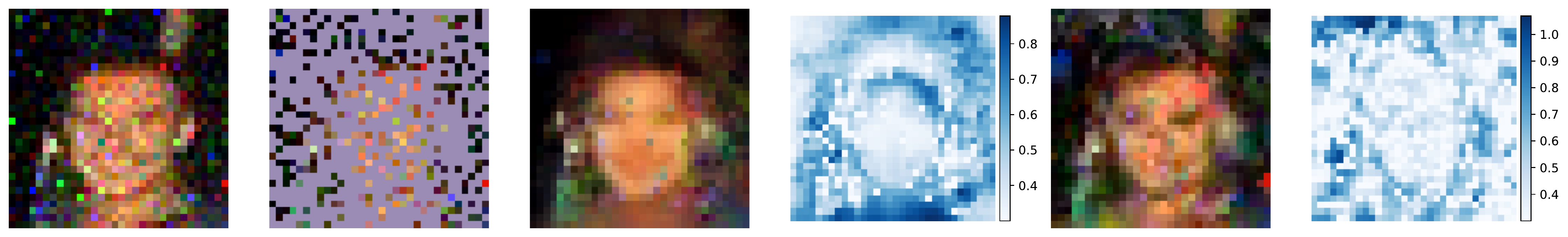} 
      \caption{Image completion results on CelebA32 for \gls{anp} and \gls{banp}. The results under increasing noise levels are shown.}
    \label{sup:fig:celeba_more_vis}
\end{figure}

We present additional visualizations for EMNIST in \cref{sup:fig:emnist_more_vis} and for CelebA in \cref{sup:fig:celeba_more_vis}.

\subsection{Predator-prey model}
\label{sup:subsec:predator_prey_figures}
\begin{figure}
    \centering
    \includegraphics[width=0.9\linewidth]{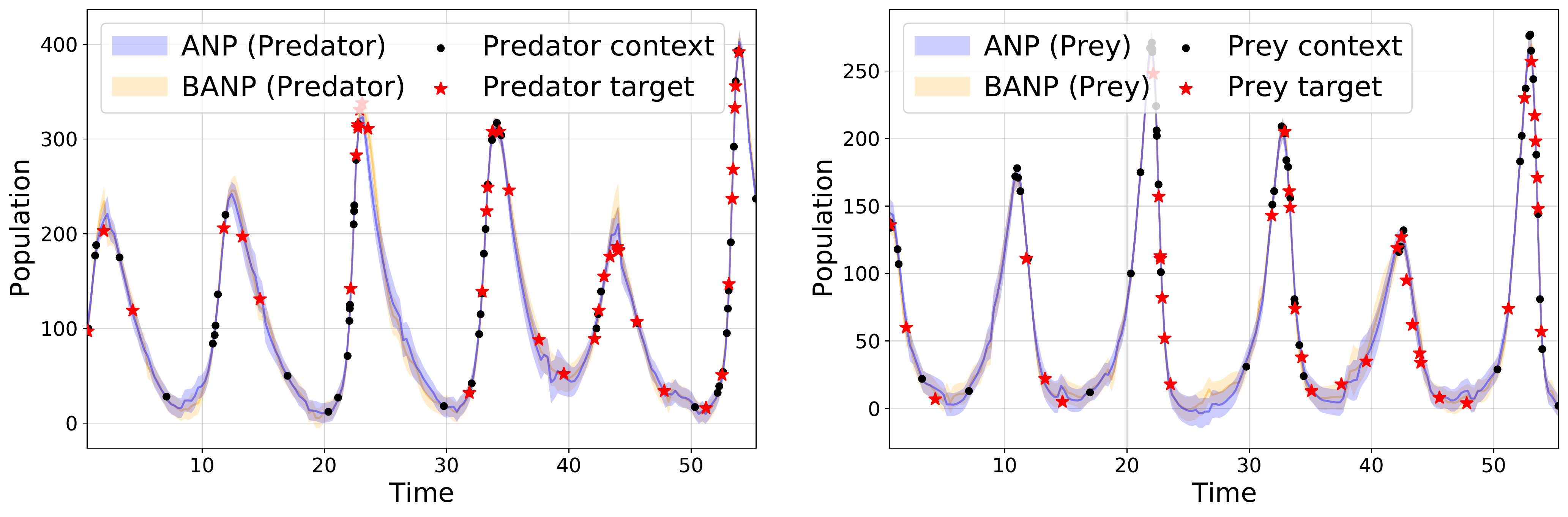}
    \includegraphics[width=0.9\linewidth]{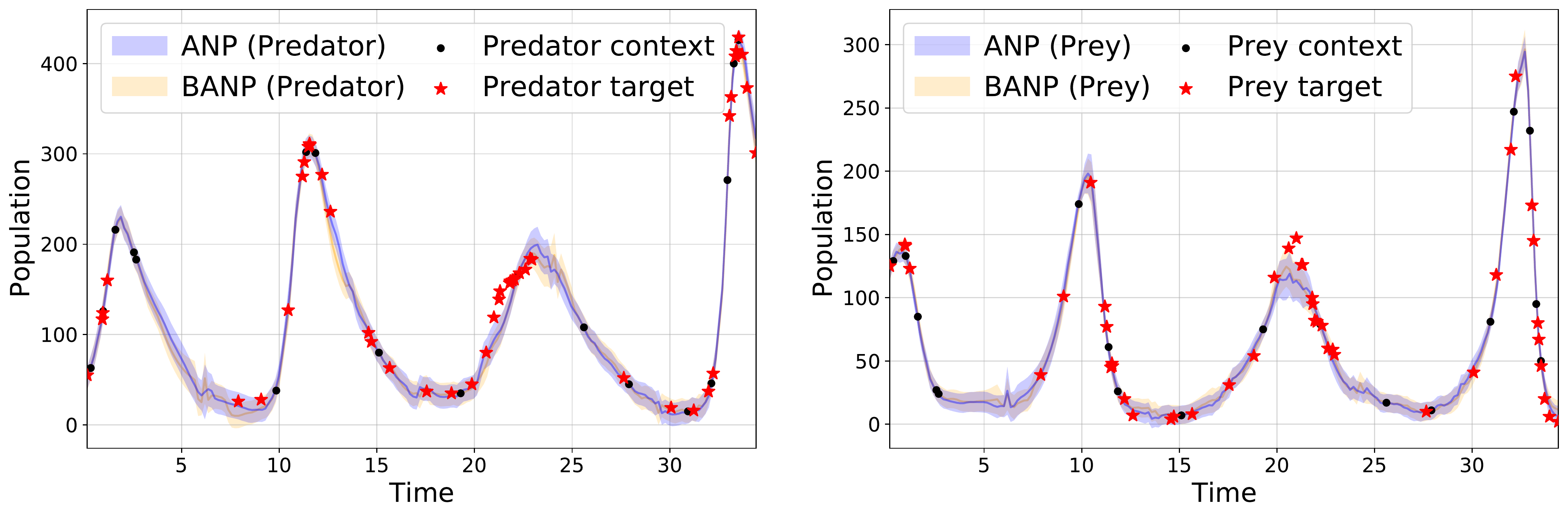}
    \includegraphics[width=0.9\linewidth]{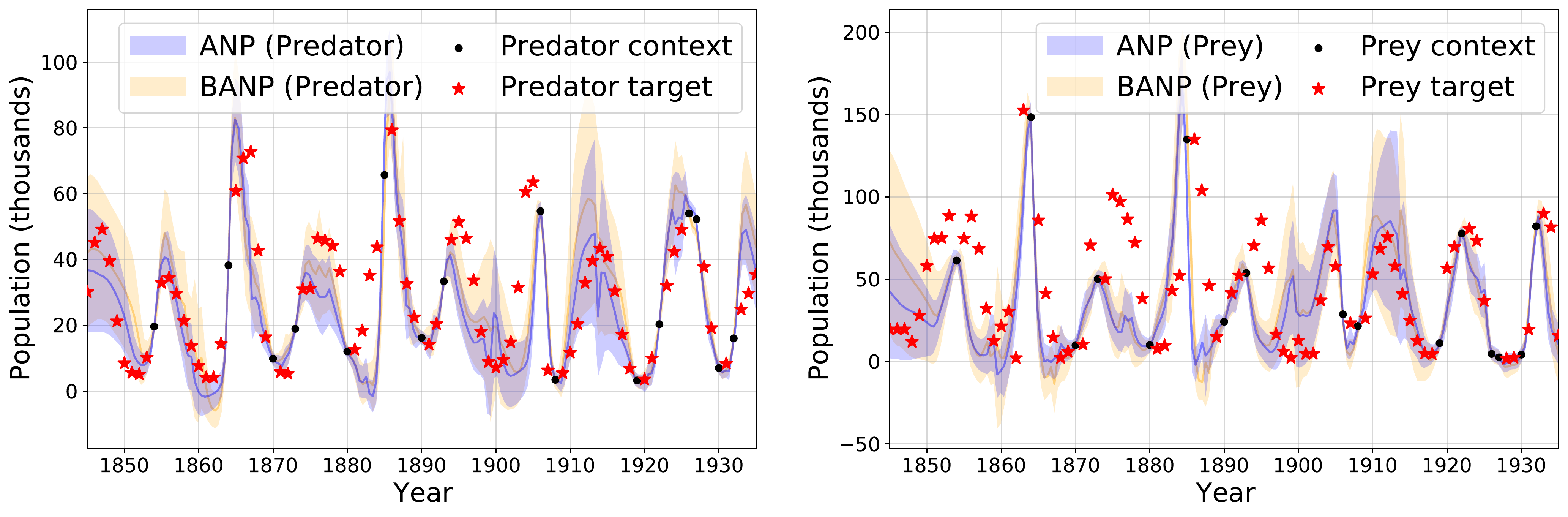}
    \includegraphics[width=0.9\linewidth]{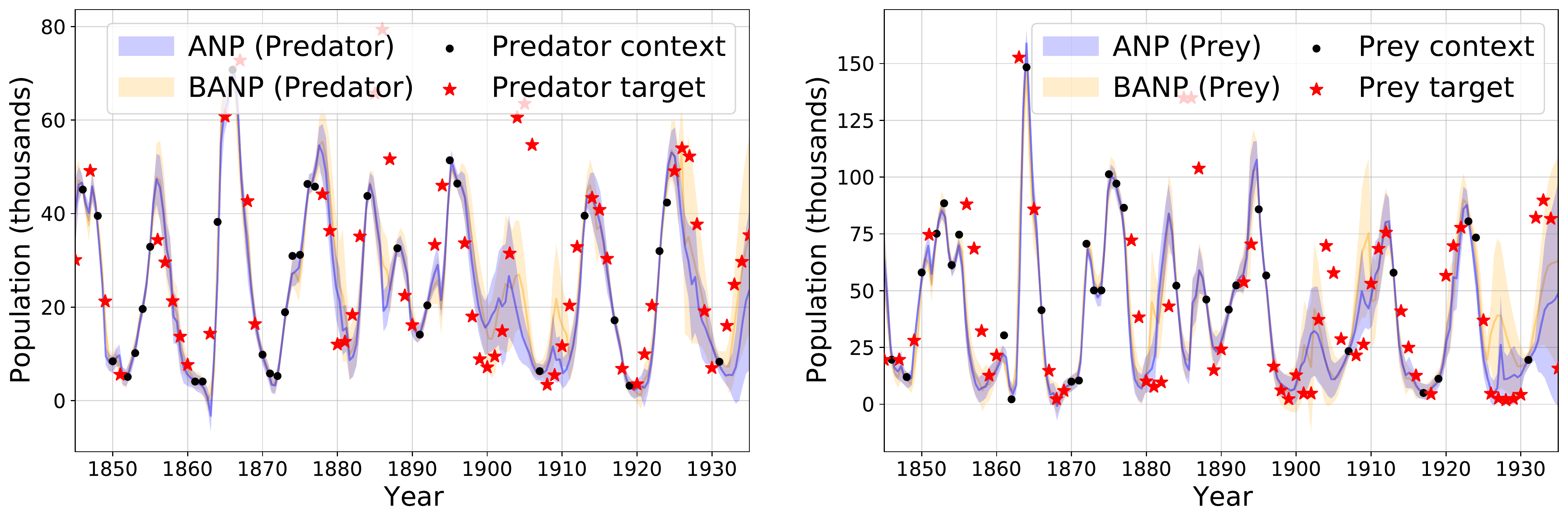}
    \caption{Regression results for predator-prey data. First two rows shows the results for simulated data, and the last two rows shows the results for the real data (Hudson's Bay hare-lynx data).}
    \label{sup:fig:predator_prey_more_vis}
\end{figure}

We present additional visualizations for predator-prey experiment in \cref{sup:fig:predator_prey_more_vis}.

\end{appendices}

\end{document}